\documentclass{article}
\usepackage[margin=1in]{geometry}
\usepackage{authblk}

\usepackage{thmtools, thm-restate}  
\usepackage{hyperref}
\usepackage{url}

\usepackage{microtype}
\usepackage{graphicx}
\usepackage{float}
\usepackage{booktabs} 
\usepackage{bbding}

\usepackage{hyperref}
\usepackage{array}
\newcolumntype{P}[1]{>{\centering\arraybackslash}p{#1}}

\usepackage{amsmath}
\usepackage{amssymb}
\usepackage{mathtools}
\usepackage{amsthm}
\usepackage[authoryear]{natbib}
\usepackage{enumerate}
\usepackage{bbm}

\usepackage{caption}

\usepackage{booktabs} 
\usepackage{tikz} 

\usetikzlibrary{shapes.geometric, arrows}

\tikzstyle{startstop} = [rectangle, rounded corners, 
minimum width=1cm, 
minimum height=0.4cm,
text centered, 
draw=black ]

\tikzstyle{io} = [trapezium, 
trapezium stretches=true, 
trapezium left angle=70, 
trapezium right angle=110, 
minimum width=3.0cm, 
minimum height=1cm, text centered, 
draw=black]

\tikzstyle{process} = [rectangle, 
minimum width=3.0cm, 
minimum height=1cm, 
text centered, 
text width=3.0cm, 
draw=black, 
fill]

\tikzstyle{decision} = [diamond, 
minimum width=3.0cm, 
minimum height=1cm, 
text centered, 
draw=black, 
fill=green!30]
\tikzstyle{arrow} = [thick,->,>=stealth]

\usepackage[american]{babel}
\usepackage{amsmath}
\usepackage{amsthm}
\usepackage{amssymb} 
\usepackage[T1]{fontenc} 
\usepackage[capitalise]{cleveref}
\usepackage{ifthen}

\usepackage[ ]{algorithmic}
\usepackage[vlined,ruled, ]{algorithm2e}

\usetikzlibrary{shapes,decorations,arrows,calc,arrows.meta,fit,positioning}
\tikzset{
    -Latex,auto,node distance =1 cm and 1 cm,semithick,
    state/.style ={ellipse, draw, minimum width = 0.7 cm},
    point/.style = {circle, draw, inner sep=0.04cm,fill,node contents={}},
    bidirected/.style={Latex-Latex,dashed},
    el/.style = {inner sep=2pt, align=left, sloped}
}

\makeatletter
\newcommand*{\centernot}{%
  \mathpalette\@centernot
}
\def\@centernot#1#2{%
  \mathrel{%
    \rlap{%
      \settowidth\dimen@{$\m@th#1{#2}$}%
      \kern.5\dimen@
      \settowidth\dimen@{$\m@th#1=$}%
      \kern-.5\dimen@
      $\m@th#1\not$%
    }%
    {#2}%
  }%
}
\makeatother

\newtheorem{theorem}{Theorem}

\newtheorem{definition}{Definition}

\newcommand{\indep}{\perp \!\!\! \perp}

\usepackage{tikz}

\usepackage[textsize=tiny]{todonotes}

\usepackage[utf8]{inputenc} 
\usepackage[T1]{fontenc}    
\usepackage{hyperref}       
\usepackage{url}            
\usepackage{booktabs}       
\usepackage{amsfonts}       
\usepackage{nicefrac}       
\usepackage{microtype}      
\usepackage{lipsum}
\usepackage{fancyhdr}       
\usepackage{graphicx}       
\graphicspath{{media/}}     

\usepackage{caption}
\usepackage{subcaption}
\usepackage{comment}

\usepackage{amsmath} 
\usepackage{xspace}

\usepackage{amsmath}
\usepackage{amssymb}
\usepackage{mathtools}
\usepackage{amsthm}
\theoremstyle{plain}

\usepackage{lipsum,lineno}
\usepackage{tabularray}

\usepackage[utf8]{inputenc} 
\usepackage[T1]{fontenc}    
\usepackage{hyperref}       
\usepackage{url}            
\usepackage{booktabs}       
\usepackage{amsfonts}       
\usepackage{nicefrac}       
\usepackage{microtype}      

\usepackage{amsmath}
\usepackage{amssymb}
\usepackage{mathtools}
\usepackage{amsthm}

\usepackage{xargs}                      
\usepackage{xcolor}

\usepackage{tikz}
\usetikzlibrary{positioning, calc, shapes.geometric, shapes, shapes.multipart, arrows.meta, arrows, decorations.markings, external, trees}

\usepackage{scalefnt}
\usetikzlibrary{shapes.misc}
\usetikzlibrary{positioning, calc, shapes.geometric, shapes, shapes.multipart, arrows.meta, arrows, decorations.markings, external, trees, fit}
\tikzset{
	-Latex,auto,node distance =1 cm and 1 cm,semithick,
	state/.style ={ellipse, draw, minimum width = 0.7 cm},
	point/.style = {circle, draw, inner sep=0.04cm,fill,node contents={}},
	nnv/.style={
		rectangle, draw,thick,minimum width=0.7cm,minimum height=1.5cm
	},
        nnh/.style={
            rectangle, draw,thick,minimum width=1.5cm,minimum height=1.0cm
          },
	outer/.style={draw=gray,dashed,thick, inner sep=3pt
	},
	XOR/.style={draw,circle,append after command={
			[shorten >=\pgflinewidth, shorten <=\pgflinewidth,]
			(\tikzlastnode.north) edge (\tikzlastnode.south)
			(\tikzlastnode.east) edge (\tikzlastnode.west)
		}
	},
	bidirected/.style={Latex-Latex,dashed},
	el/.style = {inner sep=2pt, align=left, sloped},
	cross/.style={cross out, draw=black, minimum size=2*(#1-\pgflinewidth), inner sep=0pt, outer sep=0pt},
	cross/.default={1pt}
}

\usepackage{natbib}
\usepackage{microtype}

\title{
Multi-Objective Causal Bandits:
Minimal Intervention Space and Policy-Level Learning
}

\author{%
  Muhammad Qasim Elahi$^1$,  Mahsa Ghasemi$^1$, Murat Kocaoglu$^2$ \\
       School of Electrical and Computer Engineering, Purdue University$^1$\\
       School of Computer Science, Johns Hopkins University$^2$\\

  \texttt{elahi0@purdue.edu, mahsa@purdue.edu, mkocaoglu@jhu.edu } \\
}

\usepackage{fancyhdr}
\begin{document}
\maketitle

\begin{abstract}
Many decision-making problems involve multiple objectives, where actions affect several performance metrics, fairness criteria, or safety constraints. We study multi-objective causal bandits with multiple reward nodes, where actions correspond to interventions on subsets of variables in a causal graph and performance is evaluated through Pareto optimality. 
Existing approaches to multi-objective bandits often rely on pairwise Pareto dominance or optimize a fixed scalarization. However, pairwise comparisons alone can fail to identify actions dominated by randomized policies, whose expected reward vectors lie in the convex hull of action-level rewards.
 Thus, an intervention may be suboptimal even if no single intervention dominates it, because it can be dominated by a randomized policy over other interventions. We formalize this policy-level phenomenon and show that it has consequences for both causal action-space reduction and online learning. First, we provide a necessary-and-sufficient graphical characterization of possibly Pareto-optimal minimal intervention sets. Our characterization yields a minimal, sound, and complete candidate intervention family from the graph alone, correcting prior formulations that may include intervention sets that are never Pareto-optimal under any compatible structural causal model. Second, over this reduced intervention space, we design an online UCB-style algorithm that eliminates actions using dominance tests against convex combinations of other actions, and prove logarithmic gap-dependent Pareto regret. Finally, we study constrained multi-objective causal bandits, where feasibility and optimality may be realized only with randomized policies, and develop an online UCB-style constrained bandit algorithm with sublinear regret and constraint-violation guarantees.
\end{abstract}

\section{Introduction}

Many application domains, including healthcare and public policy, require balancing multiple outcomes such as performance, fairness, and safety \cite{keeney1993decisions}. Multi-armed bandits address sequential decision-making under uncertainty, but are traditionally studied with single-objective rewards and independent arms, motivating algorithms such as upper confidence bound (UCB) and Thompson sampling \citep{bubeck2012regret,auer2002finite,thompson1933likelihood}. Structured bandits exploit dependencies between actions, including linear and combinatorial bandits \citep{dani2008stochastic,cesabianchi2012combinatorial}, as well as causal bandits, which treat actions as interventions on a causal graph 
\citep{lattimore2016causal,sen2017identifying,lee2018structural,wei2023approximate,elahi2024partial}. Most existing work on causal bandits remains single-objective, even though many causal decision problems require balancing multiple outcomes such as utility, cost, and risk.

Recent work on multi-objective multi-armed bandits (MO-MAB) extends classical bandits to vector-valued rewards, but typically defines optimality at the action level via pairwise Pareto dominance or scalarization, including Pareto-UCB~\citep{drugan2013designing}, Generalized Gini Index methods~\citep{busa2017multi}, and Pareto-regret formulations for stochastic and adversarial settings without scalarization~\citep{xu2023pareto}. While fixed linear scalarization admits an optimal deterministic action for any specified preference vector, pairwise Pareto comparisons can fail to identify actions dominated by randomized policies. Such policies induce convex combinations of action-level reward vectors, making the achievable set their convex hull. Thus, an action may be suboptimal even when no single action dominates it. Our goal is therefore to identify the complete policy-level Pareto set without specifying preference weights in advance, rather than merely identifying actions that survive pairwise comparisons.

We study the role of randomized policies in multi-objective causal bandits, where actions are interventions on a causal graph and rewards are vector-valued. The closest related work is multi-objective causal Bayesian optimization~\citep{pmlr-v267-bhatija25a}, which extends the structural causal bandit framework of~\citet{lee2018structural} to multiple objectives. We show that directly extending the single-objective graphical construction to multiple reward nodes does not yield a minimal characterization of the intervention space. In particular, Theorem~4.8 of~\citet{pmlr-v267-bhatija25a}, although stated as necessary and sufficient, can introduce intervention sets that are never Pareto-optimal under any model consistent with the graph. Thus, the resulting action space is nonminimal and may require sampling interventions that cannot support the Pareto frontier. In contrast, we provide a necessary-and-sufficient graphical characterization of possibly Pareto-optimal minimal intervention sets (PPOMISs), together with an efficient, though involved, algorithm for computing them. This yields a minimal, sound, and complete candidate family of intervention sets from the causal graph alone.

This structural characterization also shapes the learning problem. Once the candidate intervention family is identified, Pareto optimality must be evaluated over the convex hull of reward vectors, rather than through pairwise comparisons alone. We therefore develop an online UCB-based algorithm that eliminates an action only when it is certified to be dominated by a convex combination of other actions. This leads to mixture-based notions of elimination and regret gaps, which differ from the pairwise action-level gaps used in prior multi-objective bandit work~\citep{drugan2013designing,busa2017multi,xu2023pareto}, and yields logarithmic gap-dependent Pareto regret. Unlike two-phase MO-MAB approaches such as~\citep{davoodi2025stochastic}, which first explore all arms and then commit to a covering subset, our method is fully online and admits an anytime implementation. We further study constrained multi-objective causal bandits, where a linear scalarization is optimized subject to linear constraints. In this setting, feasibility and optimality may be realized only with randomized policies: no individual action may satisfy the desired trade-off, while a randomized policy may. This contrasts with action-level safe bandit formulations such as~\citep{chen2022strategies}. We therefore propose an algorithm that optimizes over distributions of interventions using optimistic estimates and prove sublinear regret and constraint-violation guarantees. Our contributions are summarized below:
\begin{itemize}
    \item We provide a necessary-and-sufficient graphical characterization of \emph{possibly Pareto-optimal minimal intervention sets (PPOMISs)}. This yields a minimal, sound, and complete candidate intervention family from the causal graph and corrects prior formulations that can include intervention sets that are never Pareto-optimal under any compatible structural causal model

    \item We formulate learning at the policy level, where achievable outcomes form the convex hull of the rewards of individual interventions. We design a UCB-based algorithm that eliminates actions via mixture-dominance tests and prove logarithmic gap-dependent Pareto regret.

    \item We study constrained multi-objective causal bandits, where feasible and optimal trade-offs may require randomized policies. We propose an algorithm that optimizes over distributions of interventions and prove sublinear regret and constraint-violation guarantees.
\end{itemize}

\section{Preliminaries and Problem Setup}

We begin by reviewing structural causal models (SCMs) and introducing the notation used throughout the paper. An SCM is a tuple \(\mathcal{M}=\langle \mathbf{V},\mathbf{U},\mathbf{F},P(\mathbf{U})\rangle\), where \(\mathbf{V}=\{X_i\}_{i=1}^{n}\cup \mathbf{Y}\) is the set of observed variables, \(\mathbf{U}\) is the set of independent exogenous variables, \(\mathbf{F}\) is the set of structural equations, and \(P(\mathbf{U})\) is the exogenous distribution~\citep{pearl2009causality}. Each equation \(f_i\in\mathbf{F}\) assigns \(X_i=f_i(\mathsf{pa}(X_i),\mathbf{U}_i)\), where \(\mathbf{U}_i\subseteq\mathbf{U}\). For a set of variables \(\mathbf{X}\), \(\mathcal{D}(\mathbf{X})\) denotes the Cartesian product of their domains. We assume that intervenable variables $X_i$ are discrete and that the reward nodes \(\mathbf{Y}=\{Y_i\}_{i=1}^{d}\) are binary and non-intervenable. Each SCM induces a causal graph \(\mathcal{G}=(\mathbf{V},\mathbf{E})\), with directed edges \(X_i\to X_j\) for parent relations and bidirected edges \(X_i\leftrightarrow X_j\) for shared latent confounders. For \(\mathbf{W}\subseteq\mathbf{V}\), \(\mathcal{G}[\mathbf{W}]\) denotes the induced subgraph. For \(V\in\mathbf{V}\), \(\mathsf{pa}(V)\), \(\mathsf{an}(V)\), and \(\mathsf{de}(V)\) denote its parents, ancestors, and descendants, respectively, excluding \(V\) itself. The c-component \(\mathsf{CC}(V)\) is the maximal set of vertices connected to \(V\) by bidirected paths~\citep{tian2002general,pedemonte2021algorithmic}. All of the notations extends naturally to sets of nodes. Actions consist of hard interventions \(do(\mathbf{X}=\mathbf{x})\), with mean reward
\(\mu_i(\mathbf{X},\mathbf{x}) := \mathbb{E}[Y_i \mid do(\mathbf{X}=\mathbf{x})]\)
for each \(i\in[d]:=\{1,\ldots,d\}\). A randomized policy over intervention values samples \(\mathbf{x}\in\mathcal{D}(\mathbf{X})\) from a fixed distribution \(P(\widetilde{\mathbf{X}})\) and then applies the hard intervention \(do(\mathbf{X}=\mathbf{x})\). Its expected reward is \(\mu_i(\mathbf{X},\widetilde{\mathbf{X}}):=\mathbb{E}[Y_i \mid do(\mathbf{X}=\widetilde{\mathbf{X}})]=\sum_{\mathbf{x}\in\mathcal{D}(\mathbf{X})}\mu_i(\mathbf{X},\mathbf{x})P(\widetilde{\mathbf{X}}=\mathbf{x})\). Thus, \(do(\mathbf{X}=\widetilde{\mathbf{X}})\) is not a new causal intervention operator, but notation for a randomized policy over deterministic hard interventions on the target set \(\mathbf{X}\), which induces a convex combination of their reward vectors. Any hard intervention on \(\mathbf{X}\) induces a post-interventional graph \(\mathcal{G}_{\overline{\mathbf{X}}}\) by removing all incoming edges to \(\mathbf{X}\). We write \(\boldsymbol{\mu}(\mathbf{X},\widetilde{\mathbf{X}}):=[\mu_1(\mathbf{X},\widetilde{\mathbf{X}}),\ldots,\mu_d(\mathbf{X},\widetilde{\mathbf{X}})]\) for the mean reward vector.

\begin{figure}[h]
\begin{subfigure}[b]{5cm}
    \includegraphics[height = 2.5cm,width=5cm]{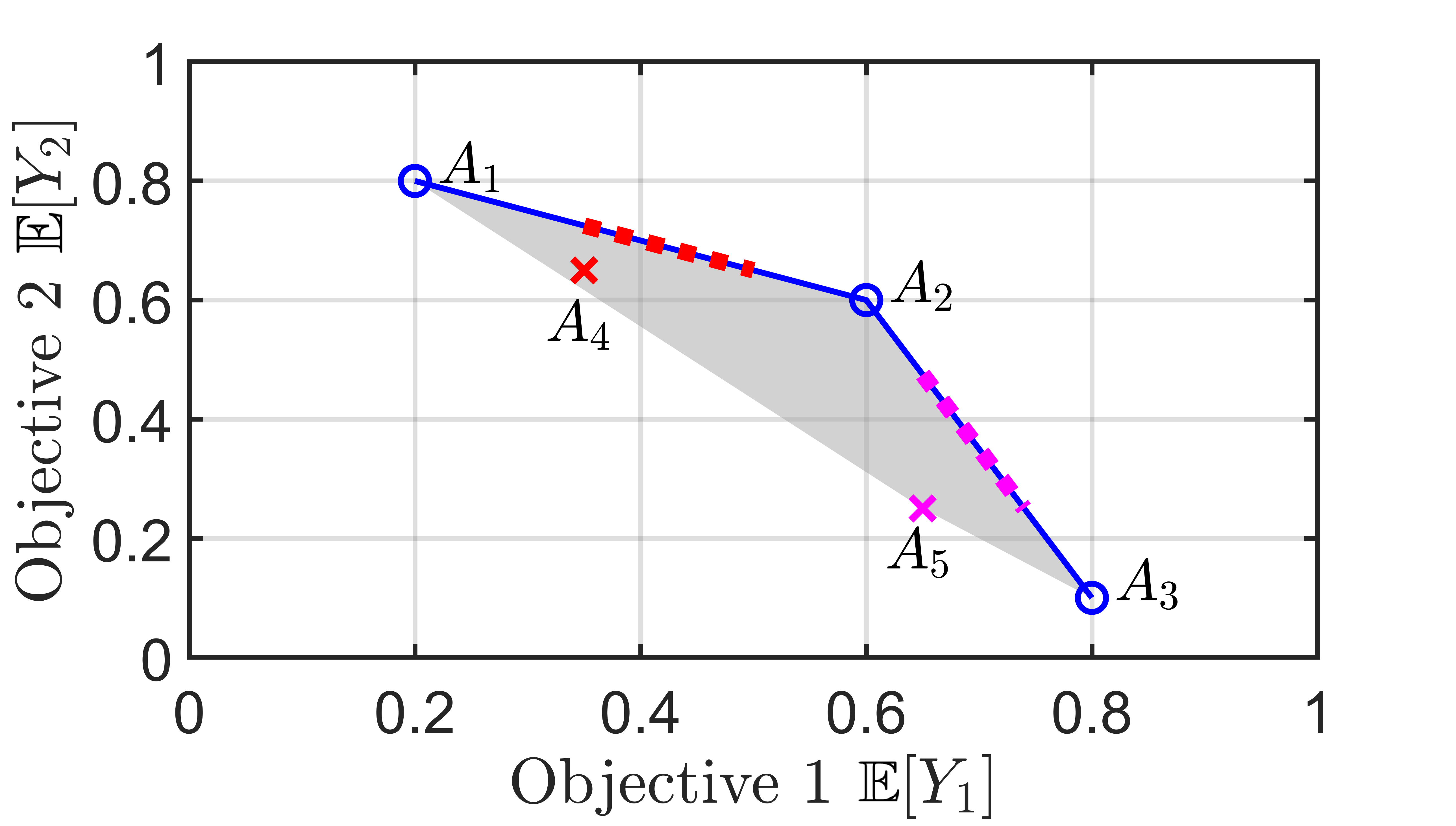}
   \subcaption{Convex hull \& Pareto front}
    \end{subfigure}
\hfill
\begin{subfigure}[b]{3.4cm}
   \centering
    \begin{tikzpicture}
  \node(1) {$X_1$};
      \node[red](2) [right of = 1,xshift = 0.8cm] {$Y_1$};

      \node[red] (5) [below of=1,yshift = -0.2cm] {$Y_2$};

  \draw (1) -> (5); 
    \draw (1) -> (2); 

\draw[dashed, <->, bend left=45] (1) to (2); 


    \end{tikzpicture}
      \subcaption{Causal Graph $\mathcal{G}$}
\end{subfigure} 
\hfill
\begin{subfigure}[b]{5cm}
    \includegraphics[height = 2.5cm,width=5cm]{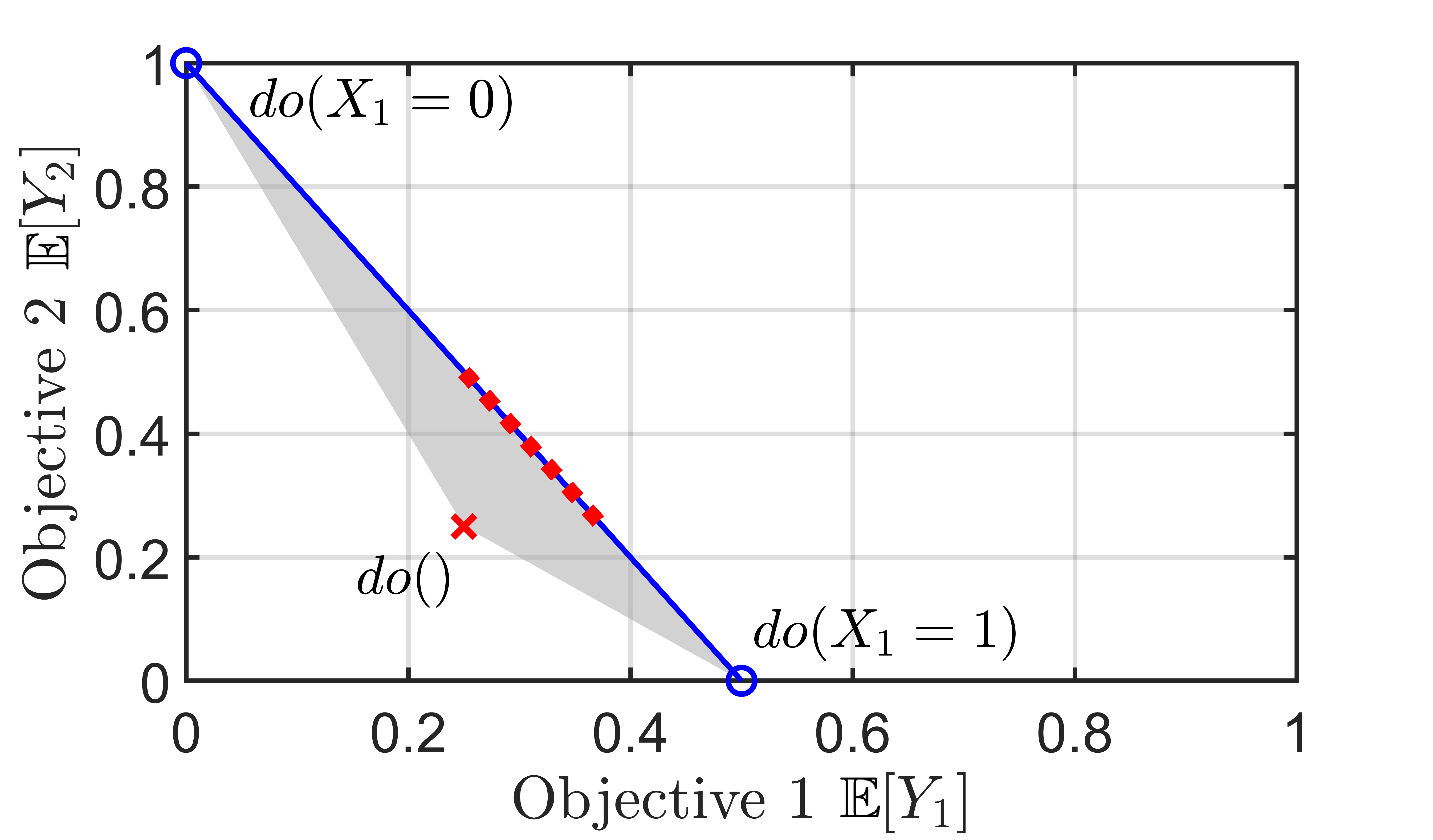}
   \subcaption{Pareto front for $\mathcal{G}$}
    \end{subfigure}
\caption{Figure to highlight role of randomized policies in multi-objective bandits }
\label{exaam}
\end{figure}
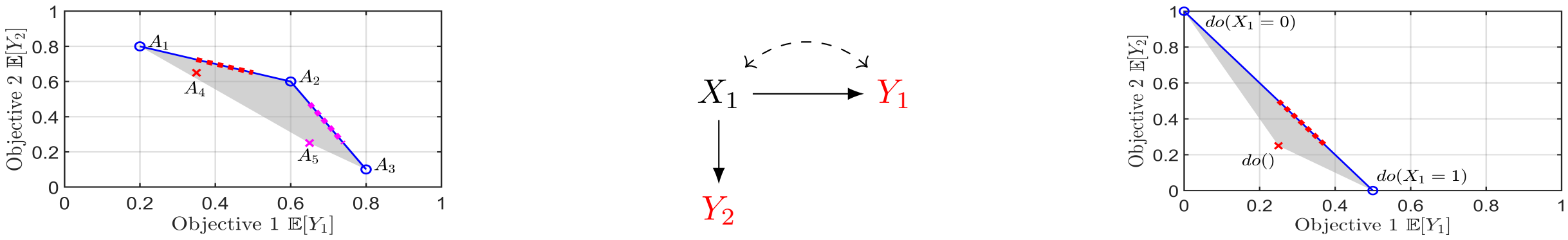

\section{Pareto Optimality under Randomized Policies in Multi-Objective Bandits}

Prior work on multi-objective bandits typically uses pairwise Pareto dominance or fixed linear scalarization~\citep{drugan2013designing,xu2023pareto}. A fixed scalarization admits an optimal deterministic action for any specified weight vector, but pairwise comparisons can miss actions dominated by randomized policies. Such policies induce convex combinations of action-level mean rewards, so the achievable set is their convex hull. Let \(\mathcal{A}\) be the action set and \(\boldsymbol{\mu}(a)\in\mathbb{R}^d\) the mean reward vector of \(a\in\mathcal{A}\). Define \(\mathcal{C}_{\mathcal{A}}:=\operatorname{conv}\{\boldsymbol{\mu}(a):a\in\mathcal{A}\}\). An action can be suboptimal even when no single action dominates it. We seek to explicitly identify the complete policy-level Pareto set without fixing preference weights: action \(a\) is policy-level Pareto-optimal if \(\boldsymbol{\mu}(a)\in\mathcal{P}(\mathcal{C}_{\mathcal{A}})\), where \(\mathcal{P}(\mathcal{C}_{\mathcal{A}})\) is the Pareto frontier of \(\mathcal{C}_{\mathcal{A}}\).

\begin{restatable}{proposition}{propPolicyPareto}
\label{prop:policy_level_pareto}
For any \(a\in\mathcal{A}\), \(\boldsymbol{\mu}(a)\in\mathcal{P}(\mathcal{C}_{\mathcal{A}})\) if and only if there is no \(\mathbf{w}\in\mathcal{S}(\mathcal{A}\setminus\{a\})\) such that \(\sum_{a'\neq a}\mathbf{w}_{a'}\boldsymbol{\mu}(a')\succeq\boldsymbol{\mu}(a)\), with strict inequality in at least one coordinate.
\end{restatable}

Here, \(\mathcal{S}(\mathcal{A})\) denotes the probability simplex over \(\mathcal{A}\). Proposition~\ref{prop:policy_level_pareto} gives a test for whether an action can support the achievable Pareto frontier. Figure~\ref{exaam}(a) illustrates the key distinction. Under pairwise Pareto comparisons, all five actions appear Pareto-optimal, as in prior Pareto-regret formulations based on non-domination among individual arm means~\citep{drugan2013designing,xu2023pareto}. In particular, for each \(j\in\{4,5\}\), there is no \(a'\in\mathcal{A}\) such that \(\boldsymbol{\mu}(a')\succeq \boldsymbol{\mu}(A_j)\) with strict improvement in at least one coordinate. With randomized policies, however, the achievable set becomes the convex hull of the action-level mean reward vectors, shown as the shaded region in Figure~\ref{exaam}(a). The relevant Pareto frontier is then the upper-right boundary highlighted in blue. Thus, only \(A_1,A_2,\) and \(A_3\) are policy-level Pareto-optimal, while \(A_4\) and \(A_5\) are dominated by randomized policies over other actions.

The same issue arises in causal bandits. Consider Figure~\ref{exaam}(b), with binary SCM \(X_1=1-U_1U_2\), \(U_1,U_2\sim\mathrm{Ber}(0.5)\), \(U_1\indep U_2\), \(Y_1=X_1 \times \mathbbm{1}\{U_1=X_1\}\), and \(Y_2=1-X_1\). The interventions \(do(X_1=0)\), \(do(X_1=1)\), and \(do()\) have expected rewards \((0,1)\), \((0.5,0)\), and \((0.25,0.25)\), respectively. As Figure~\ref{exaam}(c) shows, \(do()\) is dominated only by a randomized policy over the other two deterministic interventions, and is therefore not policy-level Pareto-optimal. This motivates Section~\ref{sec:pareto_ucb}: after the causal structure yields a candidate intervention family, the learner must also eliminate interventions dominated by mixtures. The example also motivates policy-level constrained learning: maximizing \(\mathbb{E}[Y_2\mid\pi]\) subject to \(\mathbb{E}[Y_1\mid\pi]\ge 0.25\) requires randomization, so action-level constrained formulations such as~\citet{chen2022strategies} are not directly applicable. We study this problem in Section~\ref{sec:constraints}. Finally, since actions in causal bandits are interventions, this motivates the PPOMIS characterization in Section~\ref{sec:ppomis}: before learning, one must identify the intervention sets that can support the achievable Pareto frontier under some SCM compatible with causal graph.

\section{Characterizing Possibly Pareto-Optimal Minimal Intervention Sets}
\label{sec:ppomis}

The previous section shows that, in multi-objective bandits, Pareto optimality must be evaluated at the policy level, i.e., over the convex hull of reward vectors. In causal bandits, this raises a structural question: from the graph alone, which intervention targets can support the frontier? Given a causal graph \(\mathcal{G}\) with reward nodes \(\mathbf{Y}=\{Y_i\}_{i=1}^{d}\), we seek to reduce the intervention space while retaining all intervention sets that can be Pareto-optimal for some SCM compatible with \(\mathcal{G}\). In the single-reward setting, \citet{lee2018structural} address the analogous problem using the \emph{Minimal Unobserved Confounder Territory (MUCT)}: starting from \(Y\), the MUCT expands through \(c\)-components and descendants until convergence, and its external parents form a \emph{possibly optimal minimal intervention set (POMIS)}, or interventional border \(\mathrm{IB}(\mathcal{G},Y)\). POMISs are then enumerated over post-interventional graphs \(\mathcal{G}_{\overline{\mathbf{W}}}\) for \(\mathbf{W}\subseteq \mathbf{V}\setminus\{Y\}\). \citet{pmlr-v267-bhatija25a} extend the single-reward MUCT/interventional-border construction of~\citet{lee2018structural} to multiple reward nodes, defining \emph{possibly Pareto-optimal minimal intervention sets} (PPOMISs; Definition~\ref{defppomis}). More precisely, their Definition~4.5 considers $\mathcal{H}=\mathcal{G}[\mathbf{Y}\cup\mathsf{an}(\mathbf{Y})]$ and defines a UC-territory as a set $\mathbf{T}\supseteq\mathbf{Y}$ closed under descendants and c-components, i.e., $\mathbf{T}\cup\mathsf{de}(\mathbf{T})_{\mathcal{H}}=\mathbf{T}$ and $\mathsf{CC}(\mathbf{T})_{\mathcal{H}}=\mathbf{T}$. The smallest such territory is denoted $\mathrm{MUCT}(\mathcal{G},\mathbf{Y})$. Their Definition~4.6 defines the interventional border as $\mathrm{IB}(\mathcal{G},\mathbf{Y})=\mathsf{pa}(\mathbf{T})_{\mathcal{G}}\setminus\mathbf{T}$, where $\mathbf{T}=\mathrm{MUCT}(\mathcal{G},\mathbf{Y})$. Their Theorem~4.8 states that an intervention set $\mathbf{W}$ is a PPOMIS if and only if $\mathrm{IB}(\mathcal{G}_{\overline{\mathbf{W}}},\mathbf{Y})=\mathbf{W}$. We show through counterexamples that the sufficient (``if'') direction of this multi-objective characterization does not hold: the graphical equality can hold for intervention sets that are not PPOMISs under any SCM compatible with $\mathcal{G}$. Thus, the issue is not merely a nonminimal action space, but false PPOMISs certified by the claimed characterization. Here, $\mathrm{MUCT}$ refers to the construction of~\citet{pmlr-v267-bhatija25a}, which need not coincide with the reward-specific MUCT returned by our Algorithm~\ref{alg:pomis_muct}.

\begin{definition}
A set \(\mathbf{X}\) is a \textbf{minimal intervention set (MIS)} if there does not exist \(\mathbf{X}'\subsetneq \mathbf{X}\) such that, for every SCM conforming to \(\mathcal{G}\) and every \(\mathbf{x}\in\mathcal{D}(\mathbf{X})\), there exists \(\mathbf{x}'\in\mathcal{D}(\mathbf{X}')\) satisfying \(\mu_i(\mathbf{X}',\mathbf{x}') \ge \mu_i(\mathbf{X},\mathbf{x})\) for all \(i\in[d]\).
\label{defmis}
\end{definition}

\begin{restatable}{proposition}{propPPOMIS}
\label{prop1}
A set \(\mathbf{X}\) is a minimal intervention set if and only if \(\mathbf{X}\subseteq \mathsf{an}_{\mathcal{G}_{\overline{\mathbf{X}}}}(\mathbf{Y})\).
\end{restatable}

\begin{figure}[t]
\begin{subfigure}[b]{3.4cm}
\centering
        \begin{tikzpicture}
      \node(1) {$X_3$};
      \node[red](2) [right of = 1,xshift = 0.05cm] {$Y_2$};
      \node(3) [left of = 1,xshift = -0.05cm] {$X_2$};
      \node(4) [above of = 3,yshift = -0.3em] {$X_1$};
      \node[red] (5) [below of=3,yshift = +0.3em] {$Y_1$};
 
        \draw (4) -> (3); 
  \draw (3) -> (5); 
  \draw (3) -> (1); 
  \draw (1) -> (2); 

   \draw[dashed, <->, bend right=45] (4) to (5); 

    \end{tikzpicture}
    \subcaption{$\mathcal{G}_1$}
\end{subfigure}
\hfill
\begin{subfigure}[b]{3.4cm}
    \centering
    \begin{tikzpicture}
  \node(1) {$X_3$};
      \node[red](2) [right of = 1,xshift = 0.05cm] {$Y_2$};
      \node(3) [left of = 1,xshift = -0.05cm] {$X_2$};
      \node(4) [above of = 3,yshift = -0.3em] {$X_1$};
      \node[red] (5) [below of=3,yshift = 0.3em] {$Y_1$};
 
        \draw (4) -> (3); 
  \draw (3) -> (5); 
  \draw (1) -> (3); 
  \draw (1) -> (2); 
\draw[dashed, <->, bend left=45] (3) to (2); 


    \end{tikzpicture}
    \subcaption{$\mathcal{G}_2$}
\end{subfigure} 
\hfill
\begin{subfigure}[b]{3.4cm}
\centering
   \begin{tikzpicture}
      \node(1) {$X_3$};
      \node[red](2) [right of = 1,xshift = 0.05cm] {$Y_2$};
      \node(3) [left of = 1,xshift = -0.08cm] {$X_2$};
      \node(4) [above of = 3,yshift = -0.10em] {$X_1$};
        \node(6) [right of = 4,xshift = 0.05cm] {$X_4$};
        \node[red](7) [right of = 6,xshift = 0.05em] {$Y_3$};
      \node[red] (5) [below of=3,yshift = +0.10em] {$Y_1$};
 
        \draw (4) -> (3); 
  \draw (3) -> (5); 
  \draw (3) -> (1); 
  \draw (1) -> (2); 
 \draw (1) -> (2);
  \draw (6) -> (7);
    \draw (3) -> (6);
   \draw[dashed, <->, bend right=70] (3) to (5); 

    \end{tikzpicture}
    \subcaption{$\mathcal{G}_3$}
\end{subfigure}
\hfill
\begin{subfigure}[b]{3.4cm}
\centering
    \begin{tikzpicture}
      \node(1) {$X_3$};
      \node[red](2) [right of = 1,xshift = 0.15cm] {$Y_2$};
      \node(3) [left of = 1,xshift = -0.15cm] {$X_2$};
      \node(4) [above of = 3,yshift = -0.10em] {$X_1$};
        \node(6) [right of = 4,xshift = 0.15cm] {$X_4$};
        \node[red](7) [right of = 6,xshift = 0.15cm] {$Y_3$};
      \node[red] (5) [below of=3,yshift =  +0.10em] {$Y_1$};
 
        \draw (4) -> (3); 
  \draw (3) -> (5); 
  \draw (3) -> (1); 
  \draw (1) -> (2); 
 \draw (1) -> (2);
  \draw (6) -> (7);
\draw (4) -> (6);
   \draw[dashed, <->, bend right=40] (4) to (5); 

    \end{tikzpicture}
    \subcaption{$\mathcal{G}_4$}
\end{subfigure}
\caption{Examples of causal graphs with multiple reward nodes}

\label{exp_graphs_m}
\end{figure}
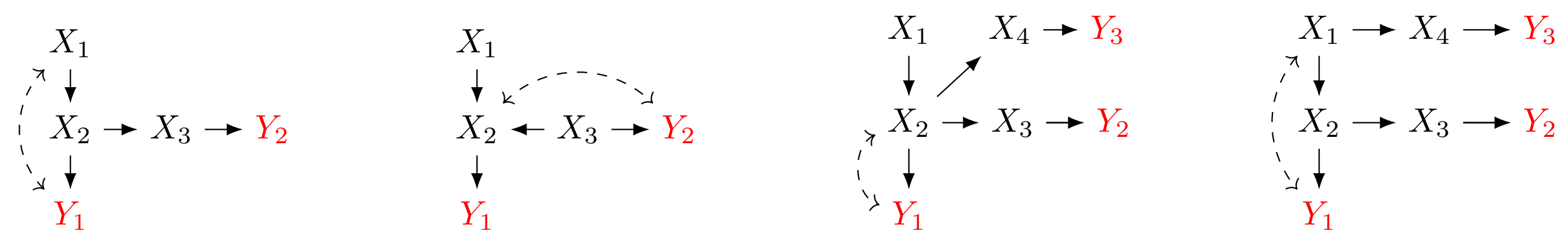

Here, \(\mathsf{an}_{\mathcal{G}_{\overline{\mathbf{X}}}}(\mathbf{Y})\) denotes the ancestors of \(\mathbf{Y}\) in the post-interventional graph. PPOMISs must be minimal intervention sets (MISs), meaning that no proper subset uniformly weakly dominates them across all reward components (Definition~\ref{defmis}). Proposition~\ref{prop1} graphically characterizes MISs, as in \cite{pmlr-v267-bhatija25a}, while our proof also covers discrete SCMs without arbitrary real-valued intervention flexibility. For example, in \(\mathcal{G}_1\) of Figure~\ref{exp_graphs_m}, \(\{X_2,X_3\}\) is an MIS. Let \(\mathbb{M}_{\mathcal{G},\mathbf{Y}}\) denote the collection of all MISs.

\begin{definition} A set $\mathbf{X}\in\mathbb{M}_{\mathcal{G},\mathbf{Y}}$ is a \textbf{Possibly Pareto-Optimal Minimal Intervention Set (PPOMIS)} if there exists at least one SCM conforming to $\mathcal{G}$ and intervention $do(\mathbf{X}=\widetilde{\mathbf{X}})$ such that no other $\mathbf{X}'\in\mathbb{M}_{\mathcal{G},\mathbf{Y}}$, has an intervention $do(\mathbf{X}'=\widetilde{\mathbf{X}}')$ satisfying $\mu_i(\mathbf{X}',\widetilde{\mathbf{X}}')\ge \mu_i(\mathbf{X},\widetilde{\mathbf{X}})$ for all $i\in[d]$. \label{defppomis} \end{definition}

The limitation of the characterization in~\citet{pmlr-v267-bhatija25a} is illustrated by the graphs in Figure~\ref{exp_graphs_m}(a)--(b), with reward nodes $\mathbf{Y}=\{Y_1,Y_2\}$. For $\mathcal{G}_1$, their construction gives $\mathrm{MUCT}(\mathcal{G}_1,\mathbf{Y})=\{Y_1,Y_2,X_1,X_2,X_3\}$: starting from $\{Y_1,Y_2\}$, it adds $X_1$ due to its confounding with $Y_1$, followed by $X_2$ and $X_3$ through descendant closure. Consequently, $\mathrm{IB}(\mathcal{G}_{1,\overline{\emptyset}},\mathbf{Y})=\emptyset$, so Theorem~4.8 classifies $\emptyset$ as a PPOMIS. Applying the same criterion to post-interventional graphs yields the candidate collection $\{\emptyset,\{X_3\},\{X_2,X_3\}\}$. However, $\emptyset$ is not a valid PPOMIS: intervening on $X_3$ leaves $Y_1$ unchanged, while some value of $X_3$ achieves an expected reward on $Y_2$ at least as large as the observational mean. For $\mathcal{G}_2$, their construction gives $\mathrm{MUCT}(\mathcal{G}_2,\mathbf{Y})=\{Y_1,Y_2,X_2\}$, since $X_2$ is confounded with $Y_2$. In particular, $\mathrm{IB}(\mathcal{G}_{2,\overline{\{X_1,X_3\}}},\mathbf{Y})=\{X_1,X_3\}$, so Theorem~4.8 classifies $\{X_1,X_3\}$ as a PPOMIS, yielding the candidate collection $\{\{X_1,X_3\},\{X_2,X_3\}\}$. However, after fixing $X_3$, intervening directly on $X_2$ can match or improve the expected reward on $Y_1$ obtained by intervening on its parent $X_1$, without changing the expected reward on $Y_2$. Thus, $\{X_1,X_3\}$ is not a valid PPOMIS. These examples directly disprove the sufficient (``if'') direction of Theorem~4.8: its graphical condition can hold for intervention sets that are not PPOMISs, rather than merely producing a nonminimal collection of valid PPOMISs. We begin with a base case: if no reward node is confounded with any of its ancestors and no reward node is a parent of another reward node, then the parent set of the rewards is the only PPOMIS.

\begin{restatable}{proposition}{propbase}
\label{prop2}
If no reward node \(Y_i\in\mathbf{Y}\) is confounded with any of its ancestors, and \(\mathsf{pa}(\mathbf{Y})\cap\mathbf{Y}=\emptyset\), then \(\mathsf{pa}(\mathbf{Y})\) is the only possibly Pareto-optimal minimal intervention set.
\end{restatable}

When the base case of Proposition \ref{prop2} does not hold  we employ Algorithm \ref{alg:pomis_muct} to compute two sets: $ \text{MUCT}(\mathcal{G},\mathbf{Y})$ and $ \text{PPOMIS}(\mathcal{G},\mathbf{Y})$. The intuition behind $ \text{MUCT}(\mathcal{G},\mathbf{Y})$ is that intervening on any node in this set can lead to suboptimal outcomes for at least one reward node in some SCM compatible with $\mathcal{G}$. Algorithm \ref{alg:pomis_muct} assumes that no reward node is an ancestor of another reward node in graph~$\mathcal{G}$; this assumption is made solely to simplify the exposition. A fully general algorithm that handles arbitrary reward--reward dependencies via additional projection steps is provided in the supplementary material (Algorithm~\ref{alg:pomis_muct_full}), and all theoretical proofs and guarantees are established for the full version. In each iteration, Algorithm \ref{alg:pomis_muct} identifies a set $\mathbf{T}$ consisting of all nodes in $\mathsf{an}(Y_i)$ that are confounded with a corresponding $Y_i$. As part of the proof of Proposition~\ref{prop3} in the supplementary material, we show that there exists a structural causal model compatible with $\mathcal{G}$ such that intervening on any node in $\mathbf{T}$ is suboptimal for at least one reward variable. For illustration, consider the graph $\mathcal{G}_1$ in Figure~\ref{exp_graphs_m}, where $X_1$ is confounded with $Y_1$ and $X_1 \in \mathsf{an}(Y_1)$. One can construct a compatible model by letting $U_1 \sim \mathrm{Ber}(0.5)$, $X_1 = U_1$, $X_2 = X_1$, and $Y_1 = \mathbbm{1}\{X_2=U_1\}$, under which intervening on $X_1$ necessarily results in sub-optimality  for $Y_1$. Since the goal is to construct one PPOMIS set, we can simply treat nodes in $\mathbf{T}$ as non-manipulable and project $\mathcal{G}$ onto $\mathbf{V}_{\mathcal{G}}\setminus \mathbf{T}$ using a procedure similar to \cite{lee2019structural}. The process end when base case is satisfied, and the union of all sets  $\mathbf{T}$ across iterations yields the final $ \text{MUCT}(\mathcal{G},\mathbf{Y})$. In the resulting projected graph, Proposition \ref{prop2} holds, and therefore $\mathsf{pa}(\mathbf{Y})$ in this final graph yields a PPOMIS. Full details are provided in the proof of Proposition \ref{prop3}.

\begin{algorithm}[h]
\small
\DontPrintSemicolon
\SetKwFunction{FMain}{ComputeMUCTandPPOMIS}
\SetKwProg{Fn}{Function}{:}{}
\Fn{\FMain{$\mathcal{G},\mathbf{Y}$}}{
    $\mathrm{MUCT} \gets \mathbf{Y}$\;
    \While{some $Y_i \in \mathbf{Y}$ is confounded with some $V_i \in \mathsf{an}(Y_i)_{\mathcal{G}}$}{
        For each $Y_i \in \mathbf{Y}$, let $\mathbf{T}_i$ be the set of nodes in $\mathsf{an}(Y_i)_{\mathcal{G}}$ that are confounded with $Y_i$\;
        $\mathbf{T} \gets \bigcup_{i:Y_i \in \mathbf{Y}} \mathbf{T}_i$\;
        Obtain the projection graph $\mathcal{H}$ on $\mathbf{V}_{\mathcal{G}} \setminus \mathbf{T}$; see Section~\ref{proj}\;
        Update: $\mathcal{G} \gets \mathcal{H}$, \quad $\mathrm{MUCT} \gets \mathrm{MUCT} \cup \mathbf{T}$\;
        Retain the remaining reward nodes: $\mathbf{Y} \gets \mathbf{Y} \cap \mathbf{V}_{\mathcal{G}}$\;
    }
    $\mathrm{PPOMIS} \gets \mathsf{pa}(\mathbf{Y})_{\mathcal{G}}$\;
    \Return $\mathrm{MUCT}, \mathrm{PPOMIS}$\;
}
\caption{Compute MUCT and PPOMIS from Graph \(\mathcal{G}\)}
\label{alg:pomis_muct}

\end{algorithm}

\begin{restatable}{proposition}{propobs}
\label{prop3}
$\text{PPOMIS}(\mathcal{G},\mathbf{Y})$ returned by Algorithm~\ref{alg:pomis_muct_full} is a possibly Pareto-optimal minimal intervention set.
\end{restatable}

\begin{restatable}{lemma}{lem}
\label{lem1}
Let $\mathbf{W} \subseteq \mathbf{V} \setminus \mathbf{Y}$ and $\mathbf{T} = \text{MUCT}(\mathcal{G}, \mathbf{Y})$. Then 
$\text{MUCT}(\mathcal{G}_{\overline{\mathbf{W}}}, \mathbf{Y}) = \text{MUCT}(\mathcal{G}_{\overline{\mathbf{W} \cap \mathbf{T}}}, \mathbf{Y})$ and 
$\text{PPOMIS}(\mathcal{G}_{\overline{\mathbf{W}}}, \mathbf{Y}) = \text{PPOMIS}(\mathcal{G}_{\overline{\mathbf{W} \cap \mathbf{T}}}, \mathbf{Y})$.
\end{restatable}

Proposition~\ref{prop3} characterizes a single PPOMIS, but our goal is to identify all PPOMISs. A naive approach would repeat Algorithm~\ref{alg:pomis_muct} or Algorithm~\ref{alg:pomis_muct_full} on every post-interventional graph \(\mathcal{G}_{\overline{\mathbf{W}}}\) for \(\mathbf{W}\subseteq\mathbf{V}\setminus\mathbf{Y}\), requiring an exponential search over node subsets. Lemma~\ref{lem1} shows that it suffices to consider only subsets of \(\mathrm{MUCT}(\mathcal{G},\mathbf{Y})\setminus\mathbf{Y}\), which is typically much smaller than \(\mathbf{V}\).

\begin{restatable}{proposition}{propint}
\label{prop4}
$\text{PPOMIS}(\mathcal{G}_{\overline{\mathbf{W}}},\mathbf{Y})$ returned by Algorithm~\ref{alg:pomis_muct_full} is a possibly Pareto-optimal minimal intervention set for any $\mathbf{W} \subseteq \mathrm{MUCT}(\mathcal{G}, \mathbf{Y}) \setminus \mathbf{Y}$.
\end{restatable}

Proposition~\ref{prop4} shows that Algorithm~\ref{alg:pomis_muct} or Algorithm~\ref{alg:pomis_muct_full} generates a collection of intervention sets, each of which is necessary: for some SCM compatible with \(\mathcal{G}\), the returned set is Pareto-optimal and is not dominated by any other intervention set across all objectives. It remains to prove sufficiency.

\begin{restatable}{theorem}{thm}
\label{thm1}
A set $\mathbf{W} \subseteq \mathbf{V} \setminus \mathbf{Y}$ is a possibly Pareto-optimal minimal intervention set if and only if
$\text{PPOMIS}(\mathcal{G}_{\overline{\mathbf{W}}}, \mathbf{Y})$ returned by Algorithm~\ref{alg:pomis_muct_full} equals $\mathbf{W}$.
\end{restatable}

Theorem~\ref{thm1} shows that applying Algorithm~\ref{alg:pomis_muct}/ Algorithm~\ref{alg:pomis_muct_full}  to  graphs \(\mathcal{G}_{\overline{\mathbf{W}}}\) with \(\mathbf{W}\subseteq \mathrm{MUCT}(\mathcal{G},\mathbf{Y})\setminus\mathbf{Y}\) recovers all PPOMISs. The ``if'' direction follows from Lemma~\ref{lem1} and Proposition~\ref{prop4}. For the ``only if'' direction, we show that any \(\mathbf{W}\) satisfying \(\mathrm{PPOMIS}(\mathcal{G}_{\overline{\mathbf{W}}},\mathbf{Y})=\mathbf{X}\neq\mathbf{W}\) cannot be a PPOMIS, since every intervention on \(\mathbf{W}\) is weakly dominated, for every SCM consistent with \(\mathcal{G}\), by some intervention on \(\mathbf{X}\). Full proof details are in the supplementary material. For \(\mathcal{G}_1\) in Figure~\ref{exp_graphs_m}, Algorithm~\ref{alg:pomis_muct} gives \(\mathrm{MUCT}(\mathcal{G}_1,\mathbf{Y})=\{Y_1,Y_2,X_1,X_2\}\) and \(\mathrm{PPOMIS}(\mathcal{G}_1,\mathbf{Y})=\{X_3\}\). Iterating over all \(\mathcal{G}_{1,\overline{\mathbf{W}}}\) with \(\mathbf{W}\subseteq \mathrm{MUCT}(\mathcal{G}_1,\mathbf{Y})\setminus\mathbf{Y}\) yields \(\{\{X_3\},\{X_2,X_3\}\}\); notably, the empty intervention does not appear, unlike in \cite{pmlr-v267-bhatija25a}. For \(\mathcal{G}_2\), the same procedure yields single PPOMIS \(\{X_2,X_3\}\), again avoiding redundancy. The contrast is stronger in larger examples: for \(\mathcal{G}_3\), the method of \cite{pmlr-v267-bhatija25a} returns \(\{\{X_1\},\{X_1,X_4\},\{X_1,X_3\},\{X_1,X_3,X_4\},\{X_2,X_3,X_4\}\}\), while ours returns the exact non-redundant collection \(\{\{X_1,X_3,X_4\},\{X_2,X_3,X_4\}\}\). Similarly, for \(\mathcal{G}_4\), their procedure returns \(\{\emptyset,\{X_4\},\{X_3\},\{X_3,X_4\},\{X_2,X_3,X_4\}\}\), whereas ours returns only \(\{\{X_3,X_4\},\{X_2,X_3,X_4\}\}\). Section~\ref{runnexm} provides step-by-step executions of Algorithm~\ref{alg:pomis_muct} on these graphs. Lemma~\ref{lem2} shows that, for the sake of efficiency, one does not need to consider all graphs $\mathcal{G}_{\overline{\mathbf{W}}}$ for every $\mathbf{W} \subseteq \text{MUCT}(\mathcal{G}, \mathbf{Y}) \setminus \mathbf{Y}$. More specifically, once we compute $\text{MUCT}(\mathcal{G}_{\overline{\mathbf{W}}},\mathbf{Y})$, there is no need to consider other graphs $\mathcal{G}_{\overline{\mathbf{W} \cup \mathbf{W}'}}$ for which $\mathbf{W}' \cap \mathbf{T} = \emptyset$, where $\mathbf{T} = \text{MUCT}(\mathcal{G}_{\overline{\mathbf{W}}},\mathbf{Y})$. Lemma~\ref{lem3} characterizes the recursive nature of the two sets $\text{MUCT}(\mathcal{G},\mathbf{Y})$ and $\text{PPOMIS}(\mathcal{G},\mathbf{Y})$.

\begin{algorithm}[t]
\small
\DontPrintSemicolon
\SetKwFunction{FMain}{\text{PPOMISs}}
\SetKwFunction{FDummy}{\text{subPPOMISs}} 
\SetKwProg{Fn}{Function}{:}{}
\Fn{\FMain{$\mathcal{G}, \mathbf{Y}$}}{
    $\mathbf{T}, \mathbf{I} = \text{ComputeMUCTandPPOMIS}(\mathcal{G}, \mathbf{Y})$
 
    \textbf{return} $\mathbf{I} \cup \text{subPPOMISs}(\mathcal{G}, \mathbf{T}, \mathbf{Y})$ 
}
\textbf{End Function}

\Fn{\FDummy{$\mathcal{G}, \mathbf{T}, \mathbf{Y}$}}{
 $\mathbf{P} = \emptyset$, $\mathbf{T}$ has a fixed arbitrary ordering.\\
  \For{$Z \in \mathbf{T} \setminus \mathbf{Y}$}{
        $\mathbf{T}_Z, \mathbf{I}_Z = \text{ComputeMUCTandPPOMIS}(\mathcal{G}_{\overline{Z}}, \mathbf{Y})$\\ 
        $Z_{\text{index}} = \mathbf{T}.\text{index}(Z)$, 
        $\;\mathbf{W} = \mathbf{T}_Z \cap  \mathbf{T}\big[\;Z_{\text{index}}+1:|\mathbf{T}|\;\big]$\\ 
        $\mathbf{P} = \mathbf{P} \cup \mathbf{I}_Z \cup \text{subPPOMISs}(\mathcal{G}_{\overline{\mathbf{I}_Z}}[\mathbf{T}_Z \cup \mathbf{I}_Z], \mathbf{W}, \mathbf{Y})$
    }
    \textbf{return} $\mathbf{P}$
}
\textbf{End Function}

\caption{Enumerate all PPOMISs from Graph $\mathcal{G}$}
\label{alg:final_ppomis}

\end{algorithm}

\begin{restatable}{lemma}{lemmm}
\label{lem2}
Consider two sets $\mathbf{W} \subseteq \text{MUCT}(\mathcal{G},\mathbf{Y}) \setminus \mathbf{Y}$ and $\mathbf{W}' \subseteq (\text{MUCT}(\mathcal{G},\mathbf{Y}) \setminus \mathbf{Y}) \setminus \mathbf{W}$, and let $\mathbf{T} = \text{MUCT}(\mathcal{G}_{\overline{\mathbf{W}}},\mathbf{Y})$. Then we have $\text{MUCT}(\mathcal{G}_{\overline{\mathbf{W} \cup \mathbf{W}'}},\mathbf{Y}) = \text{MUCT}(\mathcal{G}_{\overline{\mathbf{W} \cup (\mathbf{W}' \cap \mathbf{T})}},\mathbf{Y})$ and $\text{PPOMIS}(\mathcal{G}_{\overline{\mathbf{W} \cup \mathbf{W}'}},\mathbf{Y}) = \text{PPOMIS}(\mathcal{G}_{\overline{\mathbf{W} \cup (\mathbf{W}' \cap \mathbf{T})}},\mathbf{Y})$.
\end{restatable}

\begin{restatable}{lemma}{lemmmmm}
\label{lem3}
Consider a set $\mathbf{W} \subseteq \text{MUCT}(\mathcal{G},\mathbf{Y}) \setminus \mathbf{Y}$, and let $\mathbf{T} = \text{MUCT}(\mathcal{G}_{\overline{\mathbf{W}}},\mathbf{Y})$ and $\mathbf{I} = \text{PPOMIS}(\mathcal{G}_{\overline{\mathbf{W}}},\mathbf{Y})$. For any $\mathbf{W}' \subseteq \mathbf{T} \setminus \mathbf{Y}$, we have $\text{MUCT}(\mathcal{G}_{\overline{\mathbf{W} \cup \mathbf{W}'}},\mathbf{Y}) = \text{MUCT}(\mathcal{G}_{\overline{\mathbf{I} \cup \mathbf{W}'}}[\mathbf{T} \cup \mathbf{I}],\mathbf{Y})$ and $\text{PPOMIS}(\mathcal{G}_{\overline{\mathbf{W} \cup \mathbf{W}'}},\mathbf{Y}) = \text{PPOMIS}(\mathcal{G}_{\overline{\mathbf{I} \cup \mathbf{W}'}}[\mathbf{T} \cup \mathbf{I}],\mathbf{Y})$.
\end{restatable}

Instead of running Algorithm~\ref{alg:pomis_muct} or Algorithm~\ref{alg:pomis_muct_full} on all subgraphs \(\mathcal{G}_{\overline{\mathbf{W}}}\) for \(\mathbf{W}\subseteq \mathrm{MUCT}(\mathcal{G},\mathbf{Y})\setminus\mathbf{Y}\), Algorithm~\ref{alg:final_ppomis} uses a recursive strategy based on Lemmas~\ref{lem2} and~\ref{lem3} to explore only the necessary subsets and enumerate all and only the PPOMISs. It first computes \(\mathrm{MUCT}\) \((\mathbf{T})\) and \(\mathrm{PPOMIS}\) \((\mathbf{I})\) via \(\mathrm{ComputeMUCTandPPOMIS}(\mathcal{G},\mathbf{Y})\), and then calls \(\mathrm{subPPOMISs}\). This subroutine iteratively selects a node \(Z\in\mathbf{T}\setminus\mathbf{Y}\), computes \((\mathbf{T}_Z,\mathbf{I}_Z)=\mathrm{ComputeMUCTandPPOMIS}(\mathcal{G}_{\overline{Z}},\mathbf{Y})\), and recursively searches only within the relevant subgraph \(\mathcal{G}_{\overline{\mathbf{I}_Z}}[\mathbf{T}_Z\cup\mathbf{I}_Z]\). Lemma~\ref{lem2} justifies ignoring nodes outside \(\mathrm{MUCT}(\mathcal{G}_{\overline{Z}},\mathbf{Y})\), while Lemma~\ref{lem3} justifies restricting the recursion to \(\mathcal{G}_{\overline{\mathbf{I}_Z}}[\mathbf{T}_Z\cup\mathbf{I}_Z]\).

\begin{restatable}{theorem}{thmmm}
\label{thm2}
(Soundness and Completeness) Algorithm~\ref{alg:final_ppomis}, i.e., $\text{PPOMISs}(\mathcal{G}, \mathbf{Y})$, returns all and only Possibly Pareto-Optimal Minimal Intervention Sets.
\end{restatable}

Algorithm~\ref{alg:final_ppomis} also handles non-manipulable variables \(\mathbf{N}\): project \(\mathcal{G}\) onto \(\mathbf{V}\setminus\mathbf{N}\) using Section~\ref{proj}, and then apply the algorithm to the projected graph. Once the PPOMIS action space is identified, optimality remains policy-level: an intervention may be non-dominated by every individual intervention yet dominated by a randomized policy, equivalently by a convex combination of other interventions. Thus, learning must certify dominance against mixtures, not just individual actions. 
Existing multi-objective bandit algorithms typically rely on scalarization or pairwise Pareto comparisons~\citep{drugan2013designing,busa2017multi,xu2023pareto}. While fixed linear scalarization admits an optimal deterministic action for any specified preference vector, pairwise comparisons alone can fail to identify interventions dominated by randomized mixtures. Our goal is to identify the complete policy-level Pareto set without fixing preference weights. In the next section, we develop an online algorithm that starts from the PPOMIS action space and eliminates interventions certified not to lie on the policy-level Pareto frontier.

\section{Pareto-UCB for Multi-Objective Causal Bandits}
\label{sec:pareto_ucb}
The previous section identifies, from the causal graph alone, the intervention targets that are possibly Pareto-optimal. We now study the online learning problem under the unknown SCM: among the interventions induced by these PPOMISs, which are truly Pareto-optimal at the policy level? Let \(\mathcal{P}_{\mathcal{G},\mathbf{Y}}\) be the collection of PPOMISs returned by Algorithm~\ref{alg:final_ppomis}. This induces the finite deterministic action set \(\mathcal{A}:=\{\mathrm{do}(\mathbf{W}=\mathbf{w})\mid \mathbf{W}\in\mathcal{P}_{\mathcal{G},\mathbf{Y}},\ \mathbf{w}\in\mathcal{D}(\mathbf{W})\}\). We relabel it as \(\mathcal{A}=\{a_1,\ldots,a_k\}\), where \(k=|\mathcal{A}|\), and write \(\mu_a^i:=\mathbb{E}[Y_i\mid a]\) for the expected reward of action \(a\) on objective \(i\in[d]\). Unlike prior multi-objective bandit algorithms based on scalarization or pairwise Pareto comparisons, we evaluate optimality over stochastic policies: an intervention may be non-dominated by every individual intervention yet dominated by a convex combination of others. Accordingly, an action \(a\in\mathcal{A}\) belongs to the set of \textbf{true Pareto-optimal actions}, denoted by \(\mathcal{F}^*\), if and only if there is no \(\mathbf{w}\in\mathcal{S}(\mathcal{A}\setminus\{a\})\) such that \(\sum_{a'\in\mathcal{A}\setminus\{a\}}\mathbf{w}_{a'}\mu^i_{a'}>\mu_a^i\) for all \(i\in[d]\).

\begin{algorithm}[h]
\caption{Pareto-UCB with Convex-Combination Elimination}
\label{alg:pareto_ucb_cc}

\KwIn{Intervention action set $\mathcal{A}$ induced by PPOMISs, horizon $T$, confidence level $\delta\in(0,1)$.}
\textbf{Initialize:} $N_a\gets 0$ and $\widehat{\mu}_a^i\gets 0$ for all $a\in\mathcal{A}$ and $i\in[d]$.\\
Play each $a\in\mathcal{A}$ once and update $N_a$ and $\widehat{\mu}_a^i$ for all $i\in[d]$; set $\widehat{\mathcal{F}}_{|\mathcal{A}|+1}\gets\mathcal{A}$.\\

\For{$t=|\mathcal{A}|+1$ \KwTo $T$}{
    \For{$a\in\widehat{\mathcal{F}}_t$ and $i\in[d]$}{
        $\mu_a^{+i}(t)\gets \widehat{\mu}_a^i(t)+\sqrt{\frac{\log(2kdT/\delta)}{2N_a(t)}}$, \quad
        $\mu_a^{-i}(t)\gets \widehat{\mu}_a^i(t)-\sqrt{\frac{\log(2kdT/\delta)}{2N_a(t)}}$
    }

Eliminate $a\in\widehat{\mathcal{F}}_t$ if $\exists\,\mathbf{w}\in\mathcal{S}(\widehat{\mathcal{F}}_t\setminus\{a\})$ s.t.
$\sum_{\substack{a'\in\widehat{\mathcal{F}}_t\\ a'\neq a}} w_{a'}\mu_{a'}^{-i}(t)>\mu_a^{+i}(t)$ for all $i\in[d]$.\\

    Let $\widehat{\mathcal{F}}_{t+1}$ be the set of remaining active arms\\
    Play $A_t\gets \arg\min_{a\in\widehat{\mathcal{F}}_{t+1}}N_a(t)$ and update $N_{A_t}$ and $\widehat{\mu}_{A_t}^i$ for all $i\in[d].$
}

\end{algorithm}

Algorithm~\ref{alg:pareto_ucb_cc} maintains confidence intervals for each arm and forms an online estimate of the Pareto-optimal set $\widehat{\mathcal{F}}_t$ by conservatively eliminating an arm via an LP feasibility test whenever the confidence bounds certify that it is dominated on all objectives by a convex combination of other arms. Consequently, under the event that confidence intervals hold, no true Pareto-optimal arm is ever eliminated, i.e., $\widehat{\mathcal{F}}_t \supseteq \mathcal{F}^*$ for all $t$. The action $A_t$ is the least-sampled active arm at time $t$, with ties broken uniformly at random. To bound the number of times an arm $a \notin \mathcal{F}^*$ is played, i.e., $N_a(T)$, we define its elimination gap as $\Delta_a^{\mathrm{elim}} := \max_{\mathbf{w} \in \mathcal{S}(\mathcal{F}^*)} \min_{i \in [d]} \big(\sum_{a' \in \mathcal{F}^*} \mathbf{w}_{a'} \mu_{a'}^i - \mu_a^i\big)$. For all $a \notin \mathcal{F}^*$, $\Delta_a^{\mathrm{elim}} > 0$, quantifying the difficulty of eliminating that arm. We define the instantaneous Pareto regret incurred by playing an arm
\(a \notin \mathcal{F}^*\) as
\(\Delta_a^{\mathrm{reg}} := \min_{\mathbf{w} \in \mathbb{D}_a}
\max_{i \in [d]}
\big( \sum_{a' \in \mathcal{F}^* \setminus \{a\}} \mathbf{w}_{a'} \mu_{a'}^i
- \mu_a^i \big)\),
where
\(\mathbb{D}_a := \{\mathbf{w} \in \mathcal{S}(\mathcal{F}^* \setminus \{a\}) \mid
\sum_{a' \in \mathcal{F}^* \setminus \{a\}} \mathbf{w}_{a'} \mu_{a'}^i
- \mu_a^i \ge 0,\ \forall i \in [d]\}\).
For any \(a \notin \mathcal{F}^*\), we have \(\mathbb{D}_a \neq \emptyset\) and thus
\(\Delta_a^{\mathrm{reg}} \ge 0\), while for arms \(a^* \in \mathcal{F}^*\),
\(\mathbb{D}_{a^*} = \emptyset\) and we set
\(\Delta_{a^*}^{\mathrm{reg}} = 0\) by convention.
Alternatively, one may define \(\Delta_a^{\mathrm{reg}}\) using an
\(\ell_2\)-type aggregation, namely
\(\min_{\mathbf{w} \in \mathbb{D}_a}
\sum_{i \in [d]}
\big( \sum_{a' \in \mathcal{F}^* \setminus \{a\}} \mathbf{w}_{a'} \mu_{a'}^i
- \mu_a^i \big)^2\).
The cumulative Pareto regret is then
\(R_T = \sum_{a \notin \mathcal{F}^*} \Delta_a^{\mathrm{reg}} \, N_a(T)\).
For full details, we refer the reader to the proof of Theorem~\ref{thm3}.

\begin{restatable}{theorem}{thmmmm}
\label{thm3}
For Algorithm~\ref{alg:pareto_ucb_cc} , with probability at least $1-\delta$, we have
$\mathcal{F}^* \subseteq \widehat{\mathcal{F}}_t\;$ $\forall t \in [T]$,
$\;N_a(T) = \mathcal{O}  \left(  \frac{\log (2kdT/ \delta)}{(\Delta_a^{\mathrm{elim}})^2}  \right) $  $\forall a \notin \mathcal{F}^*$,
and the regret satisfies
$R_T = \mathcal{O}  \left( \sum_{a \notin \mathcal{F}^*} \frac{\Delta_a^{\mathrm{reg}}}{(\Delta_a^{\mathrm{elim}})^2} \log  \left(\frac{2 k d T}{\delta}\right) \right)$.
\end{restatable}

Theorem~\ref{thm3} shows that Algorithm~\ref{alg:pareto_ucb_cc} achieves logarithmic cumulative Pareto regret. Corollary~\ref{crr1} gives a pure-exploration stopping rule ensuring that, with high probability, all active arms in \(\widehat{\mathcal{F}}_t\) at termination are truly Pareto-optimal; the unknown \(\Delta\) can be replaced by a conservative lower bound. An anytime, horizon-independent version is given in Algorithm~\ref{alg:pareto_ucb_cc_any} and Theorem~\hyperlink{keypoint}{A.1} in Section~\ref{pff_ap1}.

\begin{restatable}{corollary}{crr}
\label{crr1}
For Algorithm~\ref{alg:pareto_ucb_cc}, let \(T_0(\delta):=\tau+k\), where \(\tau\) is the smallest time step satisfying \(\min\limits_{a\in\widehat{\mathcal{F}}_\tau} N_a(\tau) > \frac{8\log(2kdT/\delta)}{\Delta^2}\), with \(\Delta:=\min\limits_{a\notin\mathcal{F}^*}\Delta_a^{\mathrm{elim}}\). Then \(\mathbb{P}\!\left(\widehat{\mathcal{F}}_t=\mathcal{F}^*,\ \forall t>T_0(\delta)\right)\ge 1-\delta\).
\end{restatable}

\section{Multi-Objective Causal Bandits with Linear Constraints}
\label{sec:constraints}

We next consider constrained optimization under linear scalarization. This setting captures problems in which a scalarized objective is optimized subject to constraints on safety, fairness, performance, or resources. In the unconstrained case, an optimal policy is deterministic, so the problem reduces to a single objective bandit over the PPOMIS-induced action set, with UCB and Thompson Sampling algorithms given in Section~\ref{secalg}. The constrained case is fundamentally different: deterministic interventions may be infeasible or overly conservative, while randomized policies can be feasible and optimal. For example, with reward vectors \((1,0)\) and \((0,1)\), and objective \(\max \mathbb{E}[Y_1]\) subject to \(\mathbb{E}[Y_2]\ge 0.5\), neither intervention alone achieves the best feasible trade-off, whereas their equal mixture yields \((0.5,0.5)\). Thus, action-level constrained bandit formulations such as~\citep{chen2022strategies} are insufficient here, because feasibility must be imposed on the reward vector induced by a distribution over interventions. This yields a policy-level learning problem that requires controlling both objective regret and cumulative constraint violation induced by randomized policies over interventions.

\begin{algorithm}[t!]
\caption{UCB-Style Constrained Bandit Algorithm}
\label{alg:constraint}

\KwIn{Intervention action set $\mathcal{A}$ induced by PPOMISs, horizon $T$, confidence level $\delta\in(0,1)$.}
\textbf{Initialize:} $N_a\gets 0$ and $\widehat{\mu}_a^i\gets 0$ for all $a\in\mathcal{A}$ and $i\in[d]$\;
Play each $a\in\mathcal{A}$ once and update $N_a$ and $\widehat{\mu}_a^i$ for all $i\in[d]$.\\

\For{$t=|\mathcal{A}|+1$ \KwTo $T$}{
    \For{$a\in\mathcal{A}$ and $i\in[d]$}{
        $\mu_a^{+i}(t)\gets \min\left\{\widehat{\mu}_a^i(t)+\sqrt{\frac{\log(2kdT/\delta)}{2N_a(t)}},\,1\right\}$.\\
    }

    Set $\mathbf{U}_a(t)\gets[\mu_a^{+i}(t)]_{i=1}^{d}$ for all $a\in\mathcal{A}$\\
  
  \[
    \mathbf{w}^t\in
    \operatorname*{arg\,max}_{\mathbf{w}\in\mathcal{S}(\mathcal{A})}
    \mathbf{c}^{\top}\sum_{a\in\mathcal{A}}\mathbf{w}_a\mathbf{U}_a(t)
    \quad
    \text{s.t.}\quad
    \mathbf{A}\sum_{a\in\mathcal{A}}\mathbf{w}_a\mathbf{U}_a(t)\ge\mathbf{b}.
    \]
 
    Play $A_t\sim\mathbf{w}^t$ and update $N_{A_t}$ and $\widehat{\mu}_{A_t}^i$ for all $i\in[d]$.
}
\end{algorithm}

We formalize the constrained policy-level setting as follows. Let \(\mathbf{w}\in\mathcal{S}(\mathcal{A})\) be a stochastic policy over the PPOMIS-induced action set, with expected reward vector \(\boldsymbol{\mu}(\mathbf{w}) := \sum_{a\in\mathcal{A}}\mathbf{w}_a\boldsymbol{\mu}_a\), where \(\boldsymbol{\mu}_a := [\mu_a^i]_{i=1}^{d}\in\mathbb{R}^d\). The goal is to find \(\mathbf{w}^\star \in \operatorname*{arg\,max}_{\mathbf{w}\in\mathcal{S}(\mathcal{A})}\mathbf{c}^{\top}\boldsymbol{\mu}(\mathbf{w})\) subject to \(\mathbf{A}\boldsymbol{\mu}(\mathbf{w})\ge\mathbf{b}\), where \(\mathbf{A}\in\mathbb{R}_+^{m\times d}\), \(\mathbf{c}\in\mathbb{R}_+^d\), and \(\mathbf{b}\in\mathbb{R}_+^m\). Assuming feasibility, the learner must estimate the unknown reward vectors online. We measure performance using the one-step loss \(r_t := \max\{(\mathbf{c}^{\top}\boldsymbol{\mu}(\mathbf{w}^{\star})-\mathbf{c}^{\top}\boldsymbol{\mu}(\mathbf{w}^{t}))_+,\mathbf{1}_m^{\top}[\mathbf{b}-\mathbf{A}\boldsymbol{\mu}(\mathbf{w}^{t})]_+\}\), where \((x)_+ := \max\{0,x\}\), \([\mathbf{x}]_+\) is applied componentwise, \(\mathbf{1}_m\in\mathbb{R}^m\) is the all-ones vector, and \(\mathbf{w}^t\) is the policy selected at round \(t\). Thus, \(r_t=0\) only when the selected policy is both feasible and optimal in scalarized value, and the expected cumulative regret is \(\mathbb{E}[R_T] := \mathbb{E}[\sum_{t=1}^T r_t]=\sum_{t=1}^T\mathbb{E}[r_t]\). Algorithm~\ref{alg:constraint} optimizes over stochastic policies using optimistic reward estimates. At round \(t\), it forms upper confidence bounds \(\mu_a^{+i}(t)\) for each action \(a\in\mathcal{A}\) and objective \(i\in[d]\), and defines \(\mathbf{U}_a(t):=[\mu_a^{+i}(t)]_{i=1}^d\). It then computes \(\mathbf{w}^t\in\operatorname*{arg\,max}_{\mathbf{w}\in\mathcal{S}(\mathcal{A})}\mathbf{c}^{\top}\sum_{a\in\mathcal{A}}\mathbf{w}_a\mathbf{U}_a(t)\) subject to \(\mathbf{A}\sum_{a\in\mathcal{A}}\mathbf{w}_a\mathbf{U}_a(t)\ge\mathbf{b}\), samples \(A_t\sim\mathbf{w}^t\), and updates the estimates. On the high-probability confidence event, the optimistic estimates make the true optimal policy \(\mathbf{w}^\star\) feasible for the optimistic problem at every round. Thus, optimizing over the optimistic feasible region encourages exploration while selecting policies with high estimated scalarized reward and small estimated constraint violation.

\begin{restatable}{theorem}{thmmmmmm}
\label{thm5}
For Algorithm~\ref{alg:constraint}, if we set $\delta=1/T$, then the expected cumulative regret satisfies $\mathbb{E}[R_T]\le 4\left(\|\mathbf{c}\|_1+\mathbf{1}_m^\top\mathbf{A}\mathbf{1}_d\right)\sqrt{kT\log\!\left(2kdT^2\right)}+\max\{\|\mathbf{c}\|_1,\|\mathbf{b}\|_1\}$.
\end{restatable}

In the proof of Theorem~\ref{thm5}, we show that the two components of \(r_t\), objective suboptimality and constraint violation, each have sublinear cumulative growth in the number of rounds; hence, the cumulative regret is sublinear in \(T\). Unlike the safe-bandit formulation of~\citep{chen2022strategies}, where feasibility is arm-wise and the benchmark is the best individual safe arm, our setting is defined over distributions of interventions. Feasibility and optimality may hold only for a stochastic policy over the PPOMIS-induced action set, even when no single intervention is feasible or optimal. Thus, the analysis must learn vector-valued rewards and optimize over \(\mathcal{S}(\mathcal{A})\), rather than filter arms by individual safety indices. See the proof of Theorem~\ref{thm5} for details. Algorithm~\ref{alg:constraint} can be run on all interventions over the PPOMISs returned by Algorithm~\ref{alg:final_ppomis}; when \(|\mathcal{A}|=k\) is large, the action set can first be pruned using Algorithm~\ref{alg:pareto_ucb_cc}, incurring additional regret during pruning.

\section{Experiments}
\vspace{-0.35em}

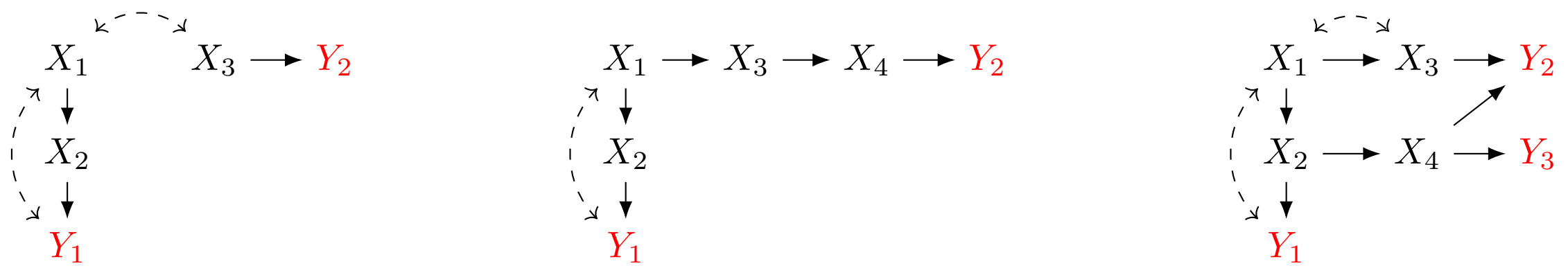
\begin{figure*}[h]
\centering

\begin{subfigure}[t]{0.30\textwidth}
\centering
\begin{tikzpicture}
  \node(1) {$X_3$};
  \node[red] (2) [right of = 1,xshift = 0.15cm] {$Y_2$};
  \node(3) [left of = 1,xshift = -0.40cm] {$X_1$};
  \node(4) [below of = 3,yshift = 0.3em] {$X_2$};
  \node[red] (5) [below of=4,yshift = 0.3em] {$Y_1$};

  \draw (3) -> (4); 
  \draw (4) -> (5); 
  \draw[dashed, <->, bend left=45] (3) to (1);
  \draw (1) -> (2); 

  \draw[dashed, <->, bend right=45] (3) to (5); 
\end{tikzpicture}
\subcaption{Graph $\mathcal{G}_5$}
\end{subfigure}
\hfill
\begin{subfigure}[t]{0.30\textwidth}
\centering
\begin{tikzpicture}
  \node(1) {$X_3$};

  \node(3) [left of = 1,xshift = -0.15cm] {$X_1$};
  \node(6) [right of = 1,xshift = 0.15cm] {$X_4$};
  \node[red] (2) [right of = 6,xshift = 0.15cm] {$Y_2$};
  \node(4) [below of = 3,yshift = 0.3em] {$X_2$};
  \node[red] (5) [below of=4,yshift = 0.3em] {$Y_1$};

  \draw (3) -> (4); 
  \draw (4) -> (5); 

  \draw (6) -> (2); 
  \draw (1) -> (6); 
  \draw (3) -> (1); 

  \draw[dashed, <->, bend right=45] (3) to (5); 
\end{tikzpicture}
\subcaption{Graph $\mathcal{G}_6$}
\end{subfigure}
\hfill
\begin{subfigure}[t]{0.30\textwidth}
\centering
\begin{tikzpicture}
  \node(1) {$X_3$};

  \node(3) [left of = 1,xshift = -0.25cm] {$X_1$};

  \node[red] (2) [right of = 1,xshift = 0.15cm] {$Y_2$};
  \node(4) [below of = 3,yshift = 0.3em] {$X_2$};
  \node(6) [right of = 4,xshift = 0.25cm] {$X_4$};
  \node[red] (7) [right of = 6,xshift = 0.15cm] {$Y_3$};
  \node[red] (5) [below of=4,yshift = 0.3em] {$Y_1$};

  \draw (3) -> (4); 
  \draw (4) -> (5); 

  \draw (6) -> (2); 
  \draw (1) -> (2); 
  \draw (3) -> (1); 
  \draw (6) -> (7); 
  \draw (4) -> (6); 
  \draw[dashed, <->, bend left=45] (3) to (1); 
  \draw[dashed, <->, bend right=45] (3) to (5); 
\end{tikzpicture}
\subcaption{Graph $\mathcal{G}_7$}
\end{subfigure}
\caption{Causal graphs with multiple reward nodes used for the experiments}

\label{grphs_exp}
\end{figure*}

\begin{figure*}[h]
\centering
\begin{subfigure}[b]{4.4cm}
  \includegraphics[height = 3.0cm,width=4.4cm]{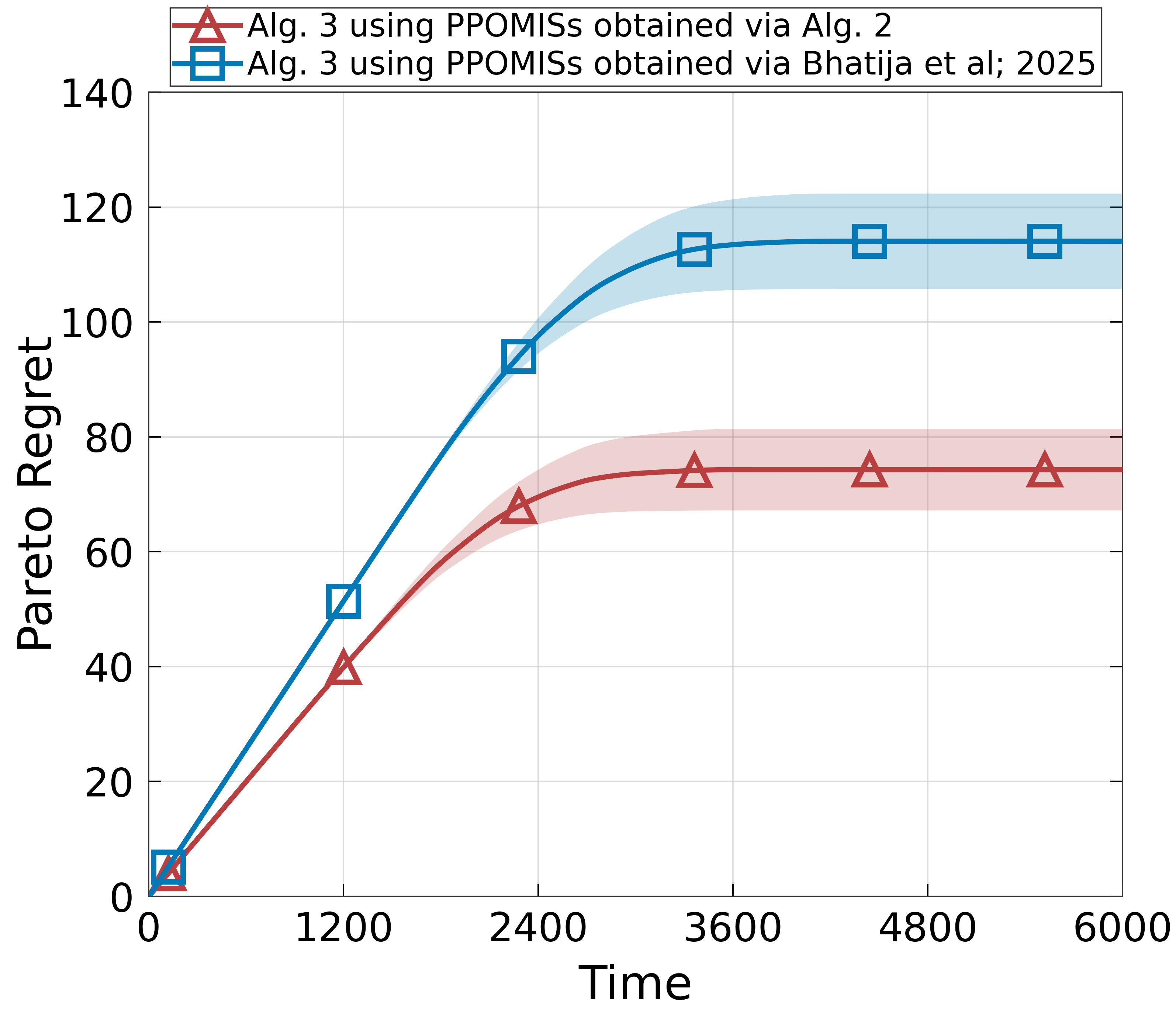}
  \subcaption{Plot for Graph $\mathcal{G}_5$}
\end{subfigure}
\enspace
\begin{subfigure}[b]{4.4cm}
  \includegraphics[height = 3.0cm,width=4.4cm]{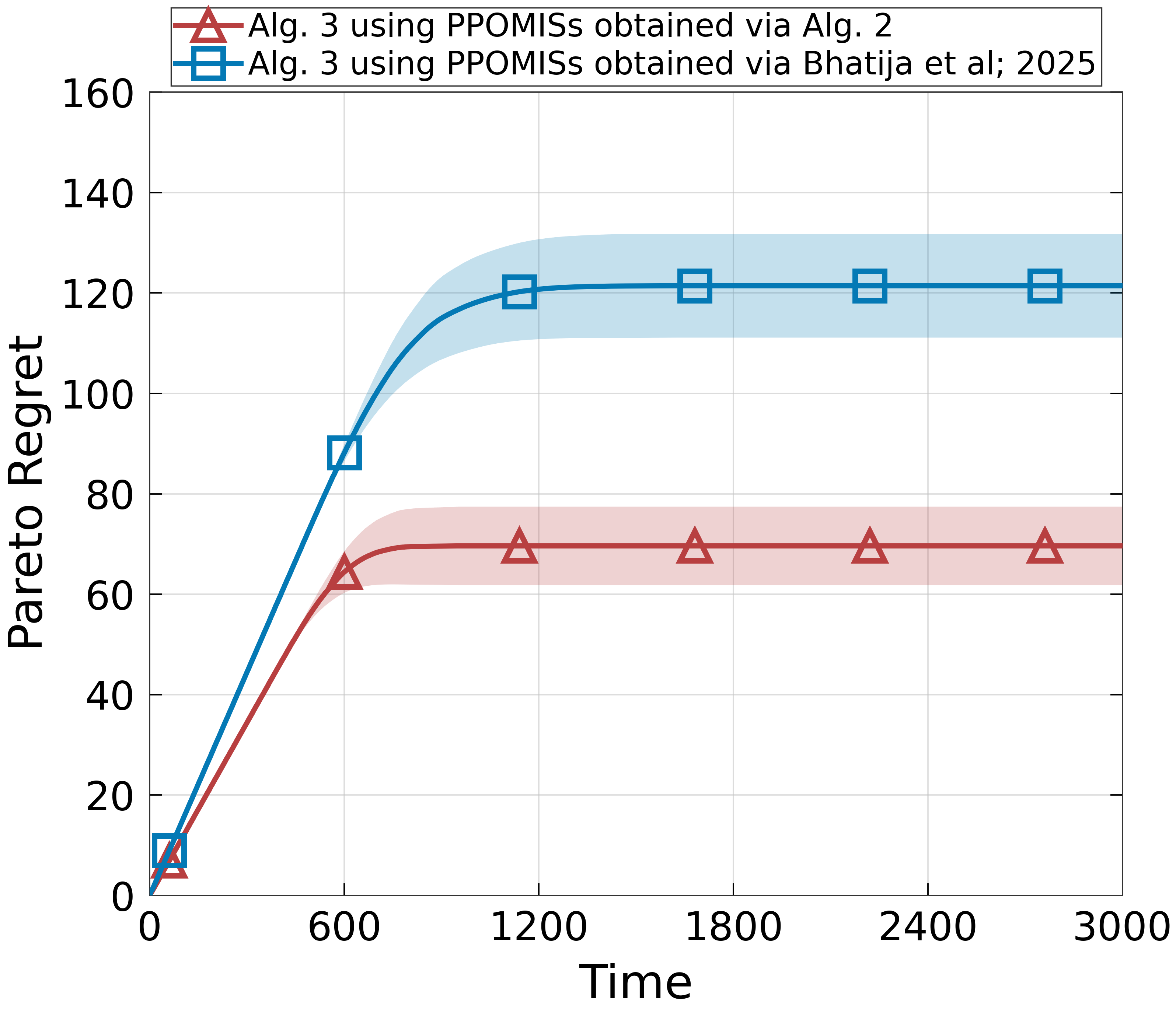}
  \subcaption{Plot for Graph $\mathcal{G}_6$}
\end{subfigure}
\enspace
\begin{subfigure}[b]{4.4cm}
  \includegraphics[height = 3.0cm,width=4.4cm]{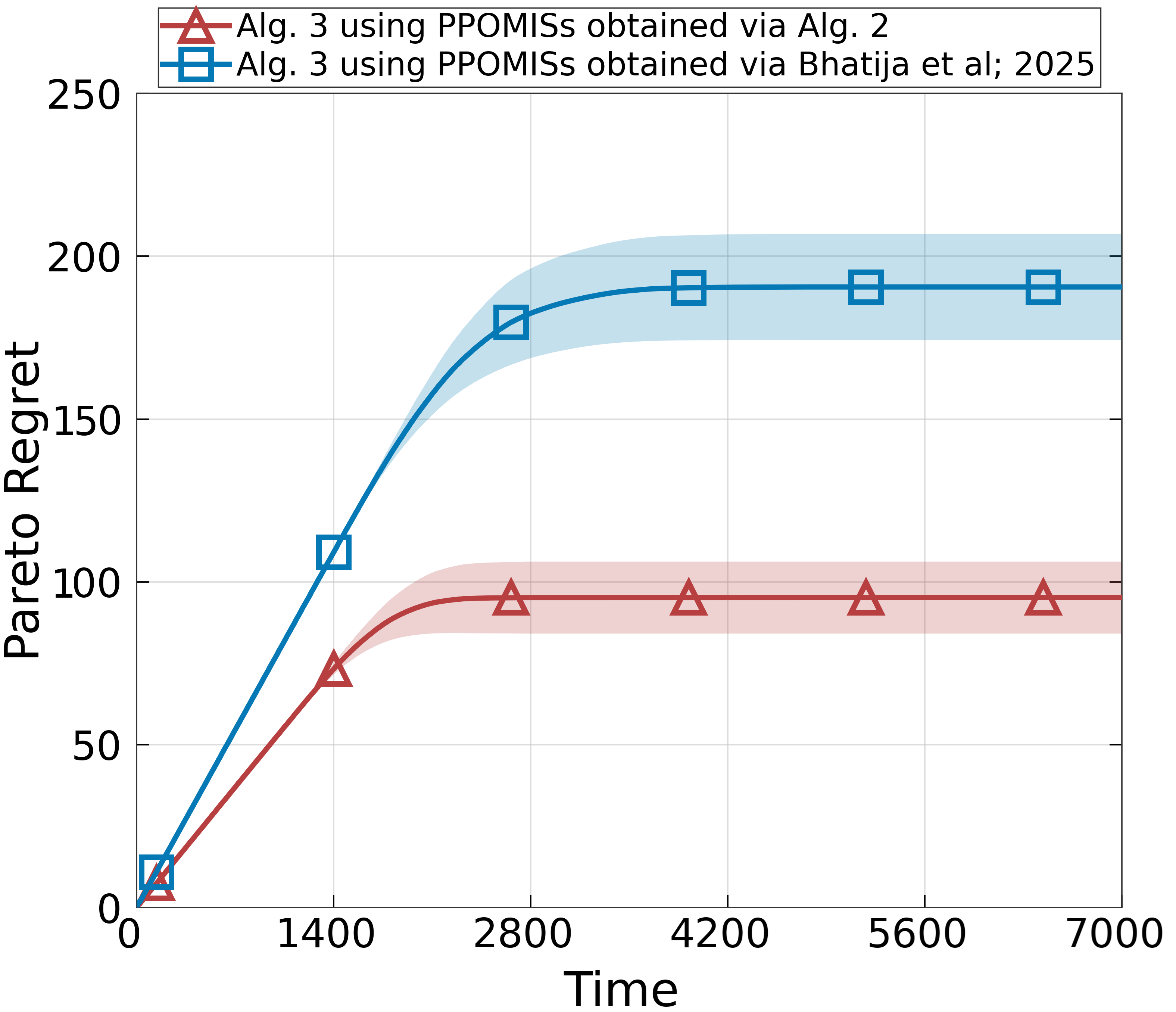}
  \subcaption{Plot for Graph $\mathcal{G}_7$}
\end{subfigure}
\caption{Cumulative Pareto regret obtained using Algorithm~\ref{alg:pareto_ucb_cc} for causal graphs shown in Figure~\ref{grphs_exp}.}
\label{syn_res1}
\end{figure*}

\begin{figure*}[t]
\centering
\begin{subfigure}[t]{4.4cm}
  \includegraphics[height=3.0cm,width=4.4cm]{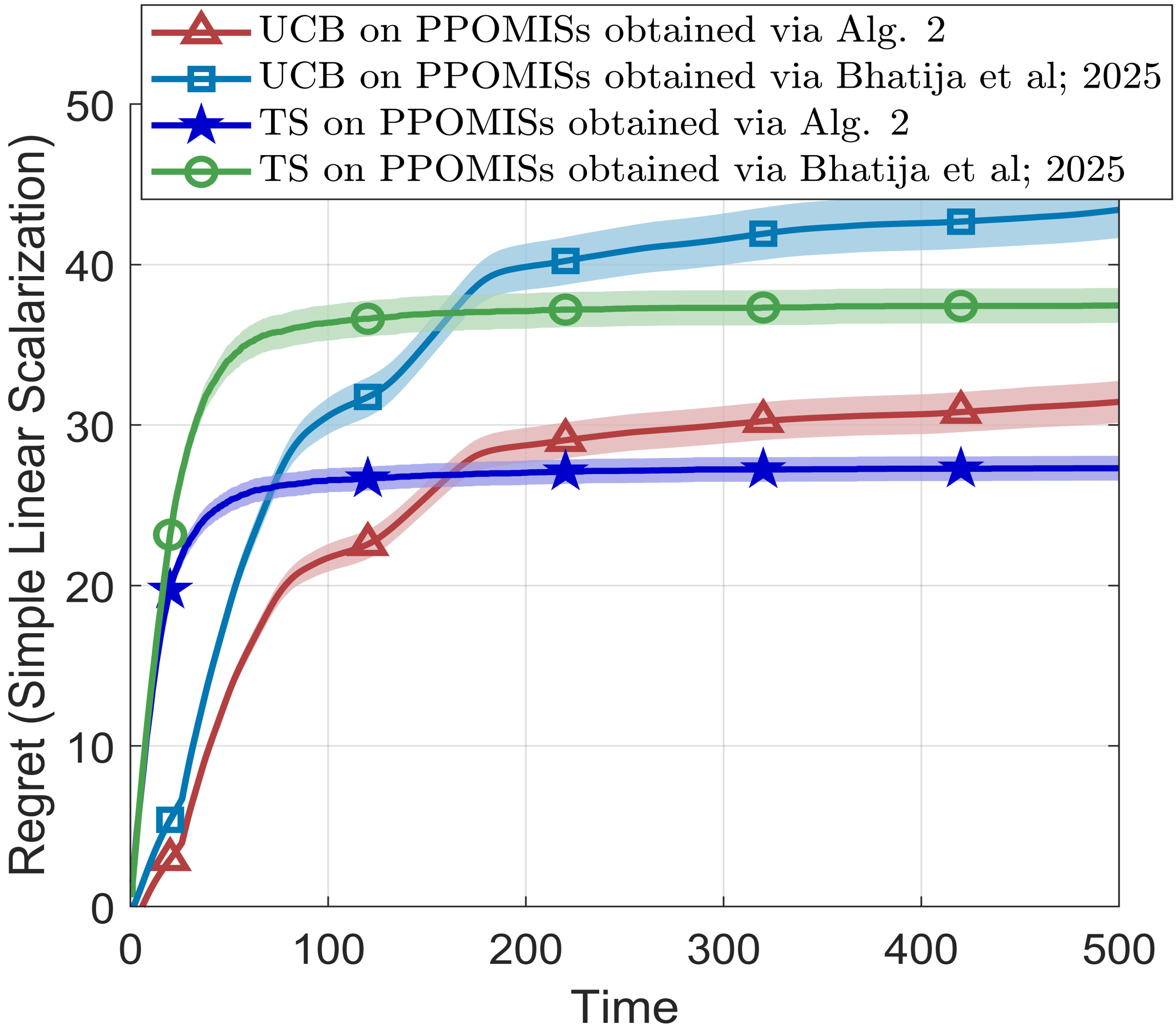}
  \subcaption{Simple Scalarization on $\mathcal{G}_5$}
\end{subfigure}
\enspace
\begin{subfigure}[t]{4.4cm}
  \includegraphics[height=3.0cm,width=4.4cm]{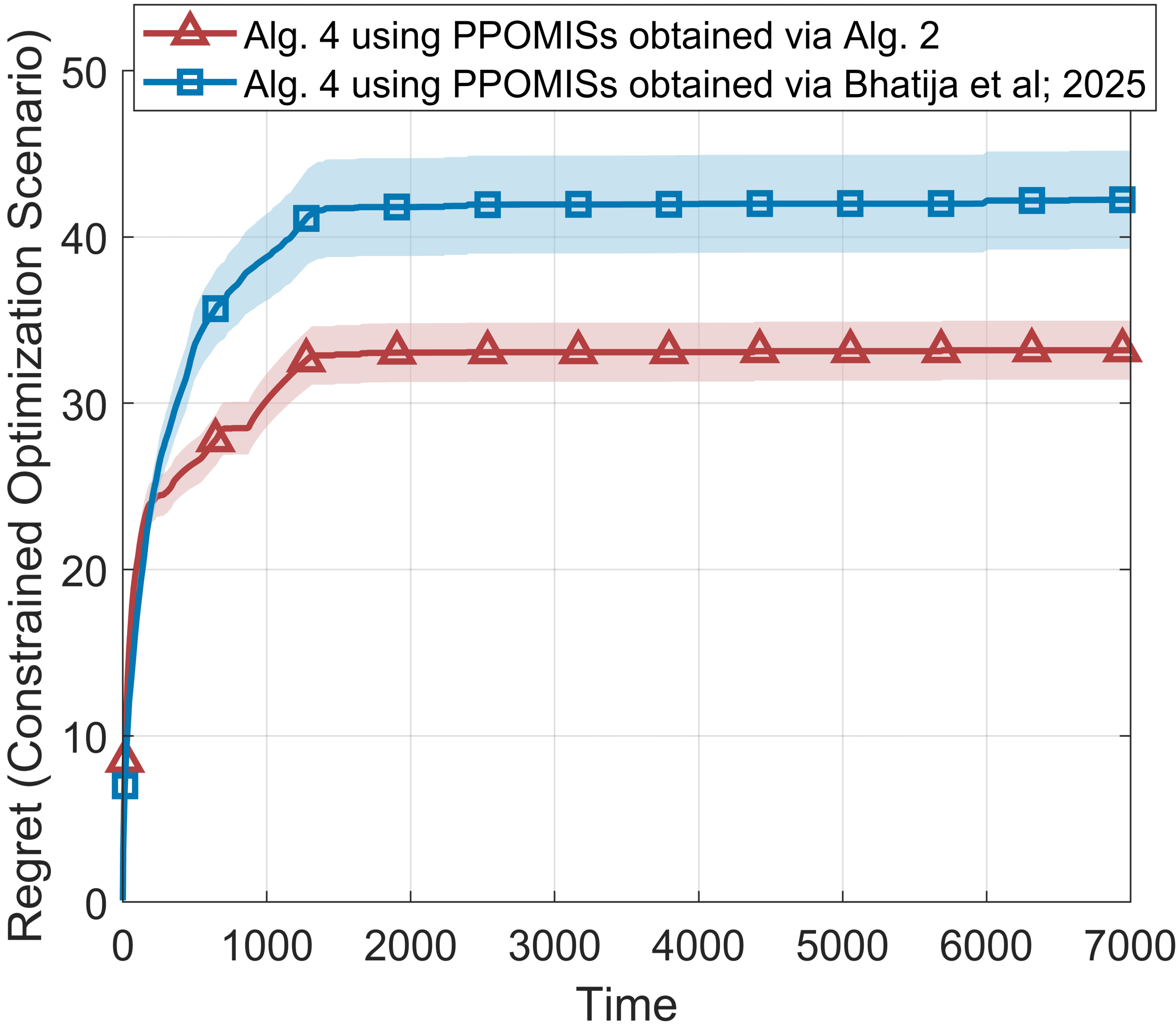}
  \subcaption{Constrained Optimization on $\mathcal{G}_6$}
\end{subfigure}
\enspace
\begin{subfigure}[t]{4.4cm}
  \includegraphics[height=3.0cm,width=4.4cm]{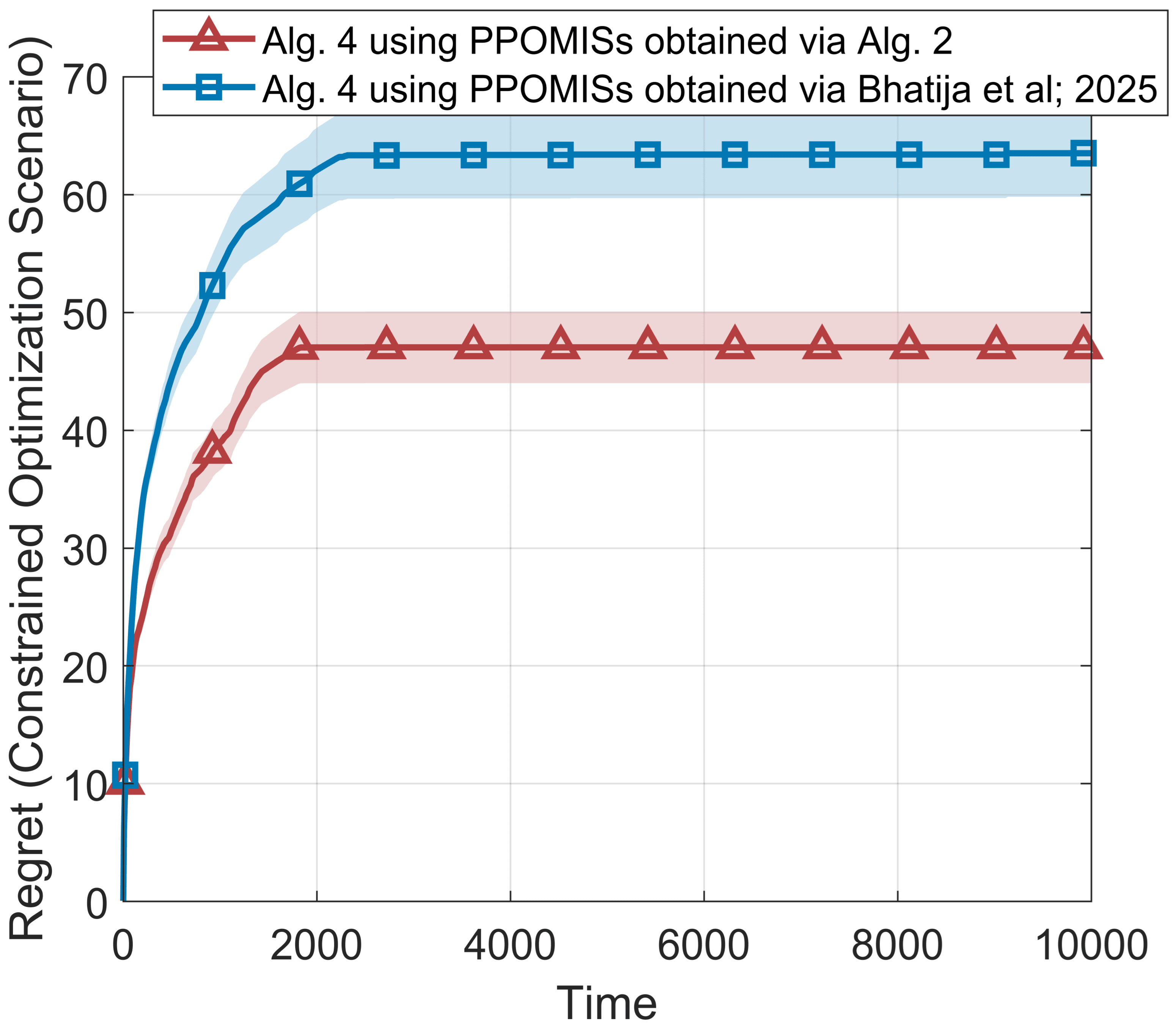}
  \subcaption{Constrained Optimization on $\mathcal{G}_7$}
\end{subfigure}
\caption{Simple linear scalarization and constrained optimization regret plots for graphs in Figure~\ref{grphs_exp}.}
\label{syn_res2}

\end{figure*}

We evaluate our algorithms on the causal graphs in Figure~\ref{grphs_exp}, comparing primarily with~\citet{pmlr-v267-bhatija25a}, whose PPOMIS characterization can include unnecessary intervention sets in multi-reward causal graphs. For $\mathcal{G}_5$ in Figure~\ref{grphs_exp}(a), their method returns $\{\emptyset,\{X_3\},\{X_2,X_3\}\}$, although $\emptyset$ is not a PPOMIS, whereas Algorithm~\ref{alg:final_ppomis} returns $\{\{X_3\},\{X_2,X_3\}\}$. Similarly, for $\mathcal{G}_6$ and $\mathcal{G}_7$, their method returns $\{\emptyset,\{X_4\},\{X_2,X_4\}\}$ and $\{\emptyset,\{X_3\},\{X_4\},\{X_3,X_4\},\{X_2,X_3,X_4\}\}$, respectively, while our algorithm returns $\{\{X_4\},\{X_2,X_4\}\}$ and $\{\{X_3,X_4\},\{X_2,X_3,X_4\}\}$. For each graph, we specify a consistent SCM and compute a common oracle Pareto set $F^\star$ and the corresponding regret gaps $\Delta_a^{\mathrm{reg}}$ using exact interventional means from the larger action space, independently of online learning trajectories. We report mean cumulative regret over $200$ independent runs, with shaded regions indicating one standard deviation.
Figure~\ref{syn_res1} shows that Algorithm~\ref{alg:pareto_ucb_cc} achieves lower Pareto regret using our reduced intervention space, with a larger separation in more complex graphs. Figure~\ref{syn_res2} reports results for linear scalarization and constrained optimization. Figure~\ref{syn_res2}(a) compares TS and UCB in the single-objective setting obtained through linear scalarization on $\mathcal{G}_5$, while Figures~\ref{syn_res2}(b) and~\ref{syn_res2}(c) consider constrained optimization on $\mathcal{G}_6$ and $\mathcal{G}_7$, respectively, with randomly chosen objective weights, constraint weights, and thresholds. In all these experiments, we compare the algorithms using the intervention spaces obtained from our PPOMIS characterization and that of~\citet{pmlr-v267-bhatija25a}. Our reduced intervention space consistently yields lower regret across these settings, demonstrating the benefit of pruning unnecessary interventions for both scalarized and constrained learning.

\begin{figure*}[!t]
\centering
\begin{subfigure}[b]{4.4cm}
  \includegraphics[height = 3.0cm,width=4.4cm]{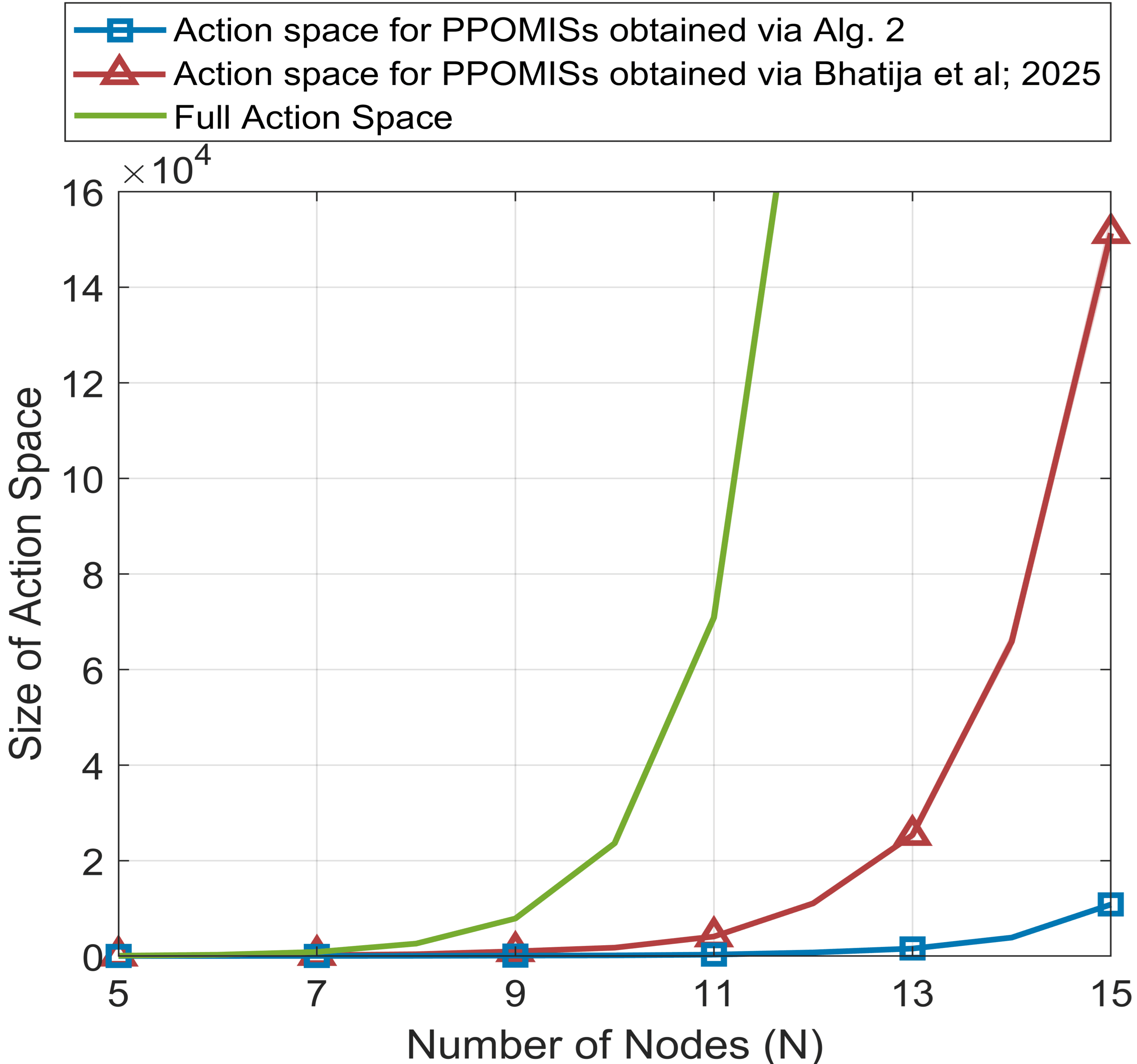}
  \subcaption{$\rho = 0.5,\;\rho_L = 0.2$}
\end{subfigure}
\enspace
\begin{subfigure}[b]{4.4cm}
  \includegraphics[height = 3.0cm,width=4.4cm]{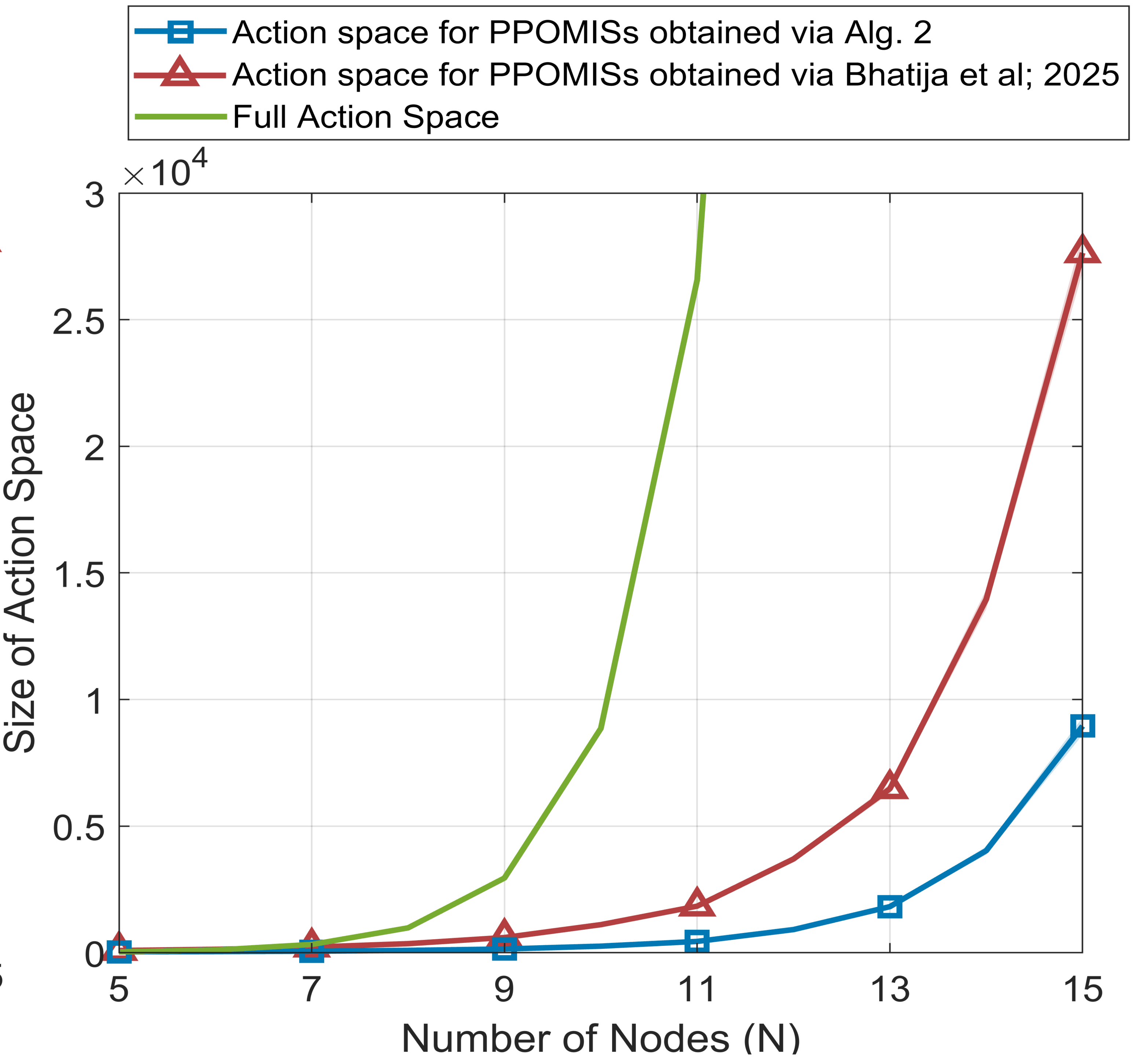}
  \subcaption{$\rho = 0.2,\;\rho_L = 0.5$}
\end{subfigure}
\enspace
\begin{subfigure}[b]{4.4cm}
  \includegraphics[height = 3.0cm,width=4.4cm]{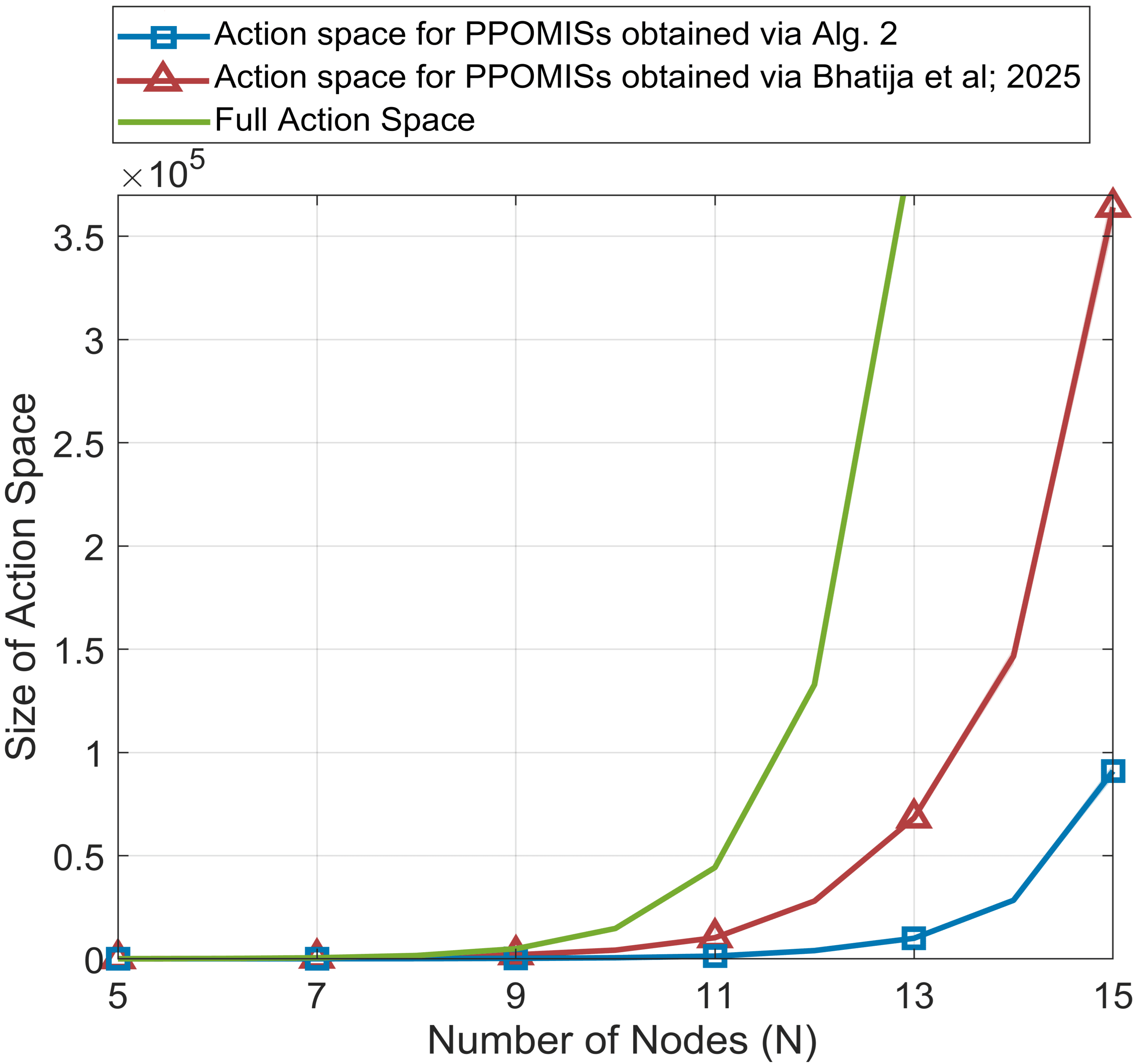}
  \subcaption{$\rho = 0.5,\;\rho_L = 0.5$}
\end{subfigure}
\caption{Comparison of the size of the action space for random causal graphs.}
\vspace{-1em}
\label{syn_res3}
\end{figure*}

To isolate the benefit of convex-combination elimination, we conduct an additional ablation using the binary SCM $X_1=1-U_1U_2$, with independent $U_1,U_2\sim\mathrm{Ber}(0.5)$, $Y_1=X_1\times\mathbbm{1}\{U_1=X_1\}$, and $Y_2=1-X_1$. The actions $\mathrm{do}()$, $\mathrm{do}(X_1=0)$, and $\mathrm{do}(X_1=1)$ have mean reward vectors $(0.25,0.25)$, $(0,1)$, and $(0.5,0)$, respectively. Although neither deterministic intervention dominates $\mathrm{do}()$, a mixture assigning probability $\lambda$ to $\mathrm{do}(X_1=0)$ and $1-\lambda$ to $\mathrm{do}(X_1=1)$ achieves $(0.5(1-\lambda),\lambda)$, which strictly dominates $\mathrm{do}()$ in all coordinates for $\lambda\in(1/4,1/2)$. The corresponding policy-level regret gap is $\Delta_{\mathrm{do}()}^{\mathrm{reg}}=\min_{\lambda\in[1/4,1/2]}\max\{1/4-\lambda/2,\lambda-1/4\}=1/12$, attained at $\lambda=1/3$. We compare Algorithm~\ref{alg:pareto_ucb_cc} with a pairwise-dominance ablation using identical confidence bounds and least-sampled-active-action selection, differing only in the elimination criterion. Across $50$ independent runs with $T=25{,}000$ and $\delta=0.1$, our algorithm eliminates $\mathrm{do}()$ in every run, with a mean elimination round of $12{,}941.52\pm1{,}582.26$, whereas the pairwise ablation never eliminates it. Final cumulative Pareto regret is $359.47\pm43.95$ for our algorithm versus $694.45\pm0.04$ for the ablation, where all reported uncertainties indicate one standard deviation. These results demonstrate that convex-combination elimination identifies policy-level suboptimal actions that pairwise dominance cannot eliminate.
Figure~\ref{fig:mixture_only_ablation} in Appendix~\ref{app:mixture_only_ablation} shows cumulative Pareto regret and the fraction of runs retaining $\mathrm{do}()$, demonstrating that convex-combination elimination removes this suboptimal action and achieves lower regret, whereas pairwise elimination fails to remove it.

We also compare the action-space sizes induced by Algorithm~\ref{alg:final_ppomis} and the characterization of~\citet{pmlr-v267-bhatija25a} on random causal graphs. We vary the directed-edge density $\rho$ and latent-confounder density $\rho_L$, which control observed edges and latent confounding, respectively. For each $(\rho,\rho_L)$, we vary the number of actionable nodes from $5$ to $15$, sample $200$ graphs per size, assume binary variables, and fix $3$ reward nodes. We additionally consider the naive baseline consisting of all interventions on subsets of non-reward ancestors of the reward nodes. Since all actionable variables are binary, each such node has three possible choices: no intervention, intervention to $0$, or intervention to $1$. Thus, the naive action-space size is $3^{m_{\mathcal G}}$, where $m_{\mathcal G}=|\mathsf{an}_{\mathcal G}(\mathbf{Y})\setminus\mathbf{Y}|$. Details of the random graph generation procedure are provided in Appendix~\ref{app:random_graph_generation}. Figure~\ref{syn_res3} shows that our PPOMIS characterization consistently yields smaller action spaces across the considered graph sizes and density settings.

\section{Conclusion}

We study multi-objective causal bandits at the policy level and provide a necessary-and-sufficient graphical characterization of possibly Pareto-optimal minimal intervention sets. Over the resulting reduced action space, we develop a UCB-based algorithm that eliminates interventions dominated by randomized policies and achieves logarithmic gap-dependent Pareto regret. We also address constrained optimization over randomized policies, proving sublinear bounds on regret and cumulative constraint violation. Our learning algorithms and guarantees extend to general finite-action multi-objective bandits. Future work includes integrating causal discovery with online learning for unknown or partially known graphs.

\section{Acknowledgements}
This research has been supported in part by NSF CAREER 2617987, IIS 2617859, Adobe Research and Intuit.
\clearpage

\clearpage

\bibliographystyle{plainnat}
\bibliography{main}

\appendix

\clearpage
\section{Technical appendices and supplementary material}

\subsection{Review of d-separation:}
Consider three disjoint sets of nodes $\mathbf{X}$, $\mathbf{Y}$, and $\mathbf{Z}$ in a causal graph
$\mathcal{G} = (\mathbf{V}, \mathbf{E})$.
We say that $\mathbf{X}$ and $\mathbf{Y}$ are \emph{$d$-separated} given $\mathbf{Z}$ in $\mathcal{G}$,
denoted by $(\mathbf{X} \indep \mathbf{Y} \mid \mathbf{Z})_{\mathcal{G}}$, if there exists no path
(directed or undirected) between any node in $\mathbf{X}$ and any node in $\mathbf{Y}$ that is
\emph{active} given $\mathbf{Z}$.
A path is active given $\mathbf{Z}$ if every collider on the path has either itself or one of its
descendants in $\mathbf{Z}$, and no non-collider on the path is contained in $\mathbf{Z}$.
A collider is a node at which two arrowheads meet, for example, node $B$ in the path
$A \rightarrow B \leftarrow C$.

\subsection{Pearl's Rules of do-Calculus:} 
Let $\mathcal{G}$ represent the causal DAG, and let $P$ denote the probability distribution induced by the corresponding causal model. For any disjoint subsets of variables $\textbf{X}, \textbf{Y}, \textbf{Z}$, and $\textbf{W}$, the following rules apply \cite{pearl2009causality}:

\textbf{Rule 1:} (Insertion/deletion of observations):
\begin{equation}
 P(\textbf{y}|do(\textbf{x}),\textbf{z},\textbf{w}) = P(\textbf{y}|do(\textbf{x}),\textbf{w}) \quad\text{if}\quad (\textbf{Y}  \indep \textbf{Z}  | \textbf{X},\textbf{W} )_{\mathcal{G}_{\overline{\textbf{X}}}}   
\end{equation} 

\textbf{Rule 2:} (Action/observation exchange):
\begin{equation}
P(\textbf{y}|do(\textbf{x}),do(\textbf{z}),\textbf{w}) = P(\textbf{y}|do(\textbf{x}),\textbf{z},\textbf{w}) \quad\text{if}\quad (\textbf{Y}  \indep \textbf{Z}  | \textbf{X},\textbf{W} )_{\mathcal{G}_{\overline{\textbf{X}}  \underline{\textbf{Z}}} } 
\end{equation}

\textbf{Rule 3:} (Insertion/deletion of actions):
\begin{equation} P(\textbf{y}|do(\textbf{x}),do(\textbf{z}),\textbf{w}) = P(\textbf{y}|do(\textbf{x}),\textbf{w}) \quad\text{if}\quad (\textbf{Y}  \indep \textbf{Z}  | \textbf{X},\textbf{W} )_{\mathcal{G}_{\overline{\textbf{X}},\overline{\textbf{Z}(\textbf{W})}} }
\end{equation}
where $\textbf{Z}(\textbf{W})$ is the set of nodes in $\textbf{Z}$ that are not ancestors of any of the nodes in $\textbf{W}$ in the graph $\mathcal{G}_{\overline{\textbf{X}}}$.  

\subsection{Proof of Proposition~\ref{prop:policy_level_pareto}}

\propPolicyPareto*

\begin{proof}
We prove both directions.

\emph{(\(\Rightarrow\))} 
Assume that \(\boldsymbol{\mu}(a)\in\mathcal{P}(\mathcal{C}_{\mathcal{A}})\). By definition of the Pareto frontier, there is no \(\mathbf{y}\in\mathcal{C}_{\mathcal{A}}\) such that
\[
\mathbf{y}\succeq \boldsymbol{\mu}(a)
\quad\text{and}\quad
\mathbf{y}\neq \boldsymbol{\mu}(a).
\]
Suppose, for contradiction, that there exists 
\(\mathbf{w}\in\mathcal{S}(\mathcal{A}\setminus\{a\})\) such that
\[
\mathbf{y}
:=
\sum_{a'\neq a}\mathbf{w}_{a'}\boldsymbol{\mu}(a')
\succeq \boldsymbol{\mu}(a),
\]
with strict inequality in at least one coordinate. Since
\(\mathbf{w}\in\mathcal{S}(\mathcal{A}\setminus\{a\})\), we have
\(\mathbf{y}\in\mathcal{C}_{\mathcal{A}}\). Moreover, the strict inequality implies
\(\mathbf{y}\neq \boldsymbol{\mu}(a)\), contradicting
\(\boldsymbol{\mu}(a)\in\mathcal{P}(\mathcal{C}_{\mathcal{A}})\). Hence, no such
\(\mathbf{w}\) exists.

\emph{(\(\Leftarrow\))} 
Assume that there is no 
\(\mathbf{w}\in\mathcal{S}(\mathcal{A}\setminus\{a\})\) such that
\[
\sum_{a'\neq a}\mathbf{w}_{a'}\boldsymbol{\mu}(a')
\succeq \boldsymbol{\mu}(a),
\]
with strict inequality in at least one coordinate. Suppose, for contradiction, that
\(\boldsymbol{\mu}(a)\notin\mathcal{P}(\mathcal{C}_{\mathcal{A}})\). Then, by definition of the Pareto frontier, there exists
\(\mathbf{y}\in\mathcal{C}_{\mathcal{A}}\) such that
\[
\mathbf{y}\succeq \boldsymbol{\mu}(a)
\quad\text{and}\quad
\mathbf{y}\neq \boldsymbol{\mu}(a).
\]
Since
\[
\mathcal{C}_{\mathcal{A}}
=
\operatorname{conv}\{\boldsymbol{\mu}(a'):a'\in\mathcal{A}\},
\]
there exists \(\bar{\mathbf{w}}\in\mathcal{S}(\mathcal{A})\) such that
\[
\mathbf{y}
=
\sum_{a'\in\mathcal{A}}\bar{\mathbf{w}}_{a'}\boldsymbol{\mu}(a').
\]

We now consider three cases. If \(\bar{\mathbf{w}}_{a}=0\), then
\[
\mathbf{y}
=
\sum_{a'\neq a}\bar{\mathbf{w}}_{a'}\boldsymbol{\mu}(a'),
\]
which directly contradicts the assumption.

If \(0<\bar{\mathbf{w}}_{a}<1\), define
\[
\widetilde{\mathbf{w}}_{a'}
:=
\frac{\bar{\mathbf{w}}_{a'}}{1-\bar{\mathbf{w}}_{a}},
\qquad a'\neq a.
\]
Then \(\widetilde{\mathbf{w}}\in\mathcal{S}(\mathcal{A}\setminus\{a\})\), and
\[
\mathbf{y}
=
\bar{\mathbf{w}}_{a}\boldsymbol{\mu}(a)
+
(1-\bar{\mathbf{w}}_{a})\mathbf{z},
\qquad
\mathbf{z}
:=
\sum_{a'\neq a}\widetilde{\mathbf{w}}_{a'}\boldsymbol{\mu}(a').
\]
Since \(\mathbf{y}\succeq\boldsymbol{\mu}(a)\), for every coordinate \(i\in[d]\),
\[
\bar{\mathbf{w}}_{a}\mu_i(a)
+
(1-\bar{\mathbf{w}}_{a})z_i
\ge
\mu_i(a).
\]
Rearranging and using \(1-\bar{\mathbf{w}}_{a}>0\) gives
\[
z_i\ge \mu_i(a),
\qquad \forall i\in[d].
\]
Thus, \(\mathbf{z}\succeq\boldsymbol{\mu}(a)\). Moreover, since
\(\mathbf{y}\neq\boldsymbol{\mu}(a)\) and
\(\mathbf{y}\succeq\boldsymbol{\mu}(a)\), there exists some coordinate
\(j\in[d]\) such that \(y_j>\mu_j(a)\). Applying the same rearrangement to coordinate \(j\) gives
\[
z_j>\mu_j(a).
\]
Therefore,
\[
\mathbf{z}
=
\sum_{a'\neq a}\widetilde{\mathbf{w}}_{a'}\boldsymbol{\mu}(a')
\succeq \boldsymbol{\mu}(a),
\]
with strict inequality in at least one coordinate, contradicting the assumption.

Finally, if \(\bar{\mathbf{w}}_{a}=1\), then
\[
\mathbf{y}=\boldsymbol{\mu}(a),
\]
which contradicts \(\mathbf{y}\neq\boldsymbol{\mu}(a)\). Hence,
\(\boldsymbol{\mu}(a)\in\mathcal{P}(\mathcal{C}_{\mathcal{A}})\).
\end{proof}

\subsection{Proof of Proposition~\ref{prop1}}
\vspace{0.5em}
\propPPOMIS*
\vspace{0.5em}

\textbf{Proof.}
We prove both directions.

\textbf{(If part).}
Assume that
$\mathbf{X}\subseteq \mathsf{an}(\mathbf{Y})_{\mathcal{G}_{\overline{\mathbf{X}}}}$.
We show that $\mathbf{X}$ is a minimal intervention set. Suppose, for contradiction, that $\mathbf{X}$ is not minimal. Then there exists a proper subset $\mathbf{X}'\subsetneq \mathbf{X}$ such that, for every SCM conforming to $\mathcal{G}$ and every $\mathbf{x}\in\mathcal{D}(\mathbf{X})$, there exists $\mathbf{x}'\in\mathcal{D}(\mathbf{X}')$ satisfying
$\mu_i(\mathbf{X}',\mathbf{x}')\ge \mu_i(\mathbf{X},\mathbf{x})$ for all $i\in[d]$.

Let $\mathbf{W}:=\mathbf{X}\setminus\mathbf{X}'$, which is nonempty. Since
$\mathbf{X}\subseteq \mathsf{an}(\mathbf{Y})_{\mathcal{G}_{\overline{\mathbf{X}}}}$,
there exists at least one directed path from some node $W\in\mathbf{W}$ to a reward node
$Y_j\in\mathbf{Y}$ in $\mathcal{G}_{\overline{\mathbf{X}}}$. This path does not pass through any node in $\mathbf{X}'$, since all incoming edges into nodes in $\mathbf{X}$ are removed in $\mathcal{G}_{\overline{\mathbf{X}}}$.

Consider an SCM in which all nodes are binary and all exogenous noise variables $U_V$ are mutually independent and uniformly distributed over $\{0,1\}$, i.e.,
$U_V\sim\mathrm{Ber}(1/2)$ for every node $V$. For all nodes $A\notin\mathbf{X}$, including the reward nodes, define their structural equations as the logical AND of their parents and corresponding noise variables, i.e.,
$A=\bigwedge_{Z\in\mathsf{pa}(A)} Z \wedge U_A$.
For all $X_\ell\in\mathbf{X}$, define
$X_\ell=f_\ell(\mathsf{pa}(X_\ell))\oplus U_{X_\ell}$,
where $f_\ell(\cdot)$ is any binary function, such as a logical AND. Since $U_{X_\ell}\sim\mathrm{Ber}(1/2)$ and is independent of the parents, $X_\ell$ is uniformly distributed over $\{0,1\}$ whenever it is not intervened upon.

Now choose an intervention value $\mathbf{x}\in\mathcal{D}(\mathbf{X})$ as follows. Set the selected node $W\in\mathbf{W}$ to $1$. For every other node in $\mathbf{X}$ that is a parent of some node on the selected directed path from $W$ to $Y_j$, set its value to $1$ as well. The remaining coordinates of $\mathbf{x}$ can be chosen arbitrarily. With this choice, the intervention $do(\mathbf{X}=\mathbf{x})$ fixes all intervened variables that could block the AND propagation along the selected path to value $1$.
Under $do(\mathbf{X}=\mathbf{x})$, the node $W$ is fixed to $1$. Moreover, along the selected path, each non-intervened node is equal to the logical AND of its parents and its exogenous noise. Therefore, conditional on the event that all exogenous noise variables on the selected path are equal to $1$, the value $1$ propagates from $W$ all the way to $Y_j$. This event has strictly positive probability. Hence
$\mu_j(\mathbf{X},\mathbf{x})>0$.

Now consider any intervention $do(\mathbf{X}'=\mathbf{x}')$ with $\mathbf{x}'\in\mathcal{D}(\mathbf{X}')$. Since $W\in\mathbf{X}\setminus\mathbf{X}'$, the intervention on $\mathbf{X}'$ does not set $W$. Therefore, under $do(\mathbf{X}'=\mathbf{x}')$, the variable $W$ remains uniformly distributed over $\{0,1\}$. In the constructed SCM, the selected path can transmit value $1$ to $Y_j$ only when $W=1$. Thus, compared with the full intervention $do(\mathbf{X}=\mathbf{x})$, the probability of reaching $Y_j=1$ under $do(\mathbf{X}'=\mathbf{x}')$ is reduced by the additional factor $\mathbb{P}(W=1)=1/2$. Consequently,
$\mu_j(\mathbf{X}',\mathbf{x}')<\mu_j(\mathbf{X},\mathbf{x})$
for every $\mathbf{x}'\in\mathcal{D}(\mathbf{X}')$.

This contradicts the assumption that $\mathbf{X}'$ uniformly weakly dominates $\mathbf{X}$ for every SCM conforming to $\mathcal{G}$ and every $\mathbf{x}\in\mathcal{D}(\mathbf{X})$. Therefore, no proper subset $\mathbf{X}'\subsetneq\mathbf{X}$ can uniformly weakly dominate $\mathbf{X}$, and $\mathbf{X}$ is a minimal intervention set.

\textbf{(Only if part).}
Assume that
$\mathbf{X}\not\subseteq \mathsf{an}(\mathbf{Y})_{\mathcal{G}_{\overline{\mathbf{X}}}}$.
Consider
$\mathbf{Z}:=\mathbf{X}\setminus \mathsf{an}(\mathbf{Y})_{\mathcal{G}_{\overline{\mathbf{X}}}}$,
which is nonempty, and let $\mathbf{X}':=\mathbf{X}\setminus\mathbf{Z}$. Then
$\mathbf{X}'\subsetneq\mathbf{X}$.

Since the variables in $\mathbf{Z}$ are not ancestors of any reward node in
$\mathcal{G}_{\overline{\mathbf{X}}}$, intervening on $\mathbf{Z}$ has no effect on any reward node once $\mathbf{X}'$ is fixed. By Pearl's do-calculus Rule 3, for every SCM conforming to $\mathcal{G}$, every $\mathbf{x}\in\mathcal{D}(\mathbf{X})$, and every $i\in[d]$, we have
$\mu_i(\mathbf{X},\mathbf{x})=\mu_i(\mathbf{X}',\mathbf{x}')$,
where $\mathbf{x}'$ is the restriction of $\mathbf{x}$ to $\mathbf{X}'$. Thus, for every $\mathbf{x}\in\mathcal{D}(\mathbf{X})$, we can choose its restriction
$\mathbf{x}'\in\mathcal{D}(\mathbf{X}')$ and obtain
$\mu_i(\mathbf{X}',\mathbf{x}')\ge \mu_i(\mathbf{X},\mathbf{x})$
for all $i\in[d]$. Hence the proper subset $\mathbf{X}'$ uniformly weakly dominates $\mathbf{X}$, so $\mathbf{X}$ is not a minimal intervention set. This completes the proof of Proposition~\ref{prop1}.

\subsection{Proof of Proposition \ref{prop2}:}
\vspace{0.5em}
\propbase*
\vspace{0.5em}
\textbf{Proof: }  Consider any set $\mathbf{W}$ such that $\mathbf{W} \neq \mathsf{pa}(\mathbf{Y})$. We have two possible cases. In the first case, suppose that the parents of all reward nodes are contained in $\mathbf{W}$, i.e., $\mathsf{pa}(\mathbf{Y}) \subset \mathbf{W}$. In this case, the set $\mathbf{W}$ violates the very definition of a minimal intervention set. When all parents of the reward nodes are intervened upon, no other nodes except those parents can be ancestors of the reward in the post-interventional graph, i.e., $\mathrm{an}(\mathbf{Y})_{\mathcal{G}_{\overline{\mathbf{W}}}} = \mathsf{pa}(\mathbf{Y})$ when $\mathsf{pa}(\mathbf{Y}) \subset \mathbf{W}$. Therefore, such a set $\mathbf{W}$ cannot be a minimal intervention set by definition and thus cannot be a Pareto-optimal minimal intervention set.

The second case implies that there exists a reward node $Y_i$ such that $\mathbf{pa}(Y_i) \not\subseteq \mathbf{W}$. Moreover, since the causal graph contains no latent confounders between any reward node and its corresponding ancestors, and assuming binary reward nodes, this implies that:

\begin{align}
    \mu_i(\mathbf{W},\widetilde{\mathbf{W}}) &= \mathbb{E}[Y_i \mid do(\mathbf{W}=\widetilde{\mathbf{W}})] \\ 
    &= \sum_{\mathbf{m} \in \mathcal{D}(\mathbf{pa}(Y_i))} P(Y_i = 1 \mid \mathbf{pa}(Y_i)=\mathbf{m}) \times P(\mathbf{pa}(Y_i)=\mathbf{m} \mid do(\mathbf{W}=\widetilde{\mathbf{W}})) \quad \forall i \in [d]
\end{align}

The stochastic intervention $do(\mathbf{W} = \widetilde{\mathbf{W}})$ induces a fixed distribution over the parents of the reward nodes, i.e., $P(\mathsf{pa}(\mathbf{Y}) \mid do(\mathbf{W} = \widetilde{\mathbf{W}}))$. Let us denote the parent set of all reward nodes $\mathsf{pa}(\mathbf{Y})$ as $\mathbf{pa}_Y$. Note that $\mathbf{pa}_Y$ is a minimal intervention set. Now, consider the stochastic intervention $do(\mathbf{pa}_Y = \widetilde{\mathbf{pa}_Y})$, where the distribution of $\widetilde{\mathbf{pa}_Y}$ is the same as $P(\mathsf{pa}(\mathbf{Y}) \mid do(\mathbf{W} = \widetilde{\mathbf{W}}))$. This results in $\mu_i(\mathbf{W}, \widetilde{\mathbf{W}}) = \mu_i(\mathsf{pa}(\mathbf{Y}), \widetilde{\mathbf{pa}_Y})$ for all $i \in [d]$. Thus, we can match the expected value of all reward nodes for any hard stochastic intervention on the set $\mathbf{W}$ using a corresponding hard stochastic intervention on the parent set of the reward nodes, i.e., $\mathbf{pa}_Y$. Note that the set of all parents of all reward nodes $\mathsf{pa}(\mathbf{Y})$ is already a minimal intervention set, and it can match the expected value of all reward nodes for any hard stochastic intervention on the set $\mathbf{W}$ such that $\mathsf{pa}(\mathbf{Y}) \not\subseteq \mathbf{W}$. This implies that any set $\mathbf{W}$ such that $\mathbf{pa}_Y \not\subseteq \mathbf{W}$ cannot be a possibly Pareto-optimal minimal intervention set.

We have shown that any set $\mathbf{W}$ such that $\mathbf{W} \neq \mathsf{pa}(\mathbf{Y})$ cannot be a possibly Pareto-optimal minimal intervention set. Now, we consider the parent set of all reward nodes, i.e., $\mathsf{pa}_{\mathbf{Y}}$, which is clearly a minimal intervention set. We further prove that it satisfies the definition of a possibly Pareto-optimal minimal intervention set. Consider the following SCM with all endogenous and exogenous variables being binary:

\begin{itemize}
    \item $f_{Y_i} = 1 - (U_{Y_i} \oplus \bigvee \mathsf{pa}_{Y_i})$ with $P(U_{Y_i} = 0) = 1 - \epsilon$, for all $i = 1, \ldots, a$ ($\epsilon > 0$ is a small real number)
    
    \item $f_X = (\bigoplus \mathbf{U}_X) \oplus (\bigoplus \mathsf{pa}_X)$ for every $X \in \mathsf{pa}_{\mathbf{Y}}$, and $P(U_i = 0) = 0.5$ for every $U_i \in U \setminus (\bigcup_{i=1}^{d} U_{Y_i})$.
\end{itemize}

Note that under the intervention $do(\mathsf{pa}_{\mathbf{Y}} = 0)$, we have $\mathbb{E}[Y_i \mid do(\mathsf{pa}_{\mathbf{Y}} = 0)] = P(U^{Y_i} = 0) \geq 1 - \epsilon$ for all $i \in [d]$. However, any intervention $\mathbf{W} \neq \mathsf{pa}_{\mathbf{Y}}$ yields expectations less than $0.5(1 + \epsilon)$ in at least one component. Note that $1 - \epsilon > 0.5(1 + \epsilon)$ for $\epsilon < \frac{1}{3}$. Thus, for this particular SCM, the intervention $do(\mathsf{pa}_{\mathbf{Y}} = 0)$ outperforms all other possible interventions on any set $\mathbf{W}$ such that $\mathsf{pa}_{\mathbf{Y}} \not\subseteq \mathbf{W}$ in at least one component $Y_i$. This implies that the parent set of all reward nodes, i.e., $\mathsf{pa}_{\mathbf{Y}}$, is the only set that satisfies the definition of a possibly Pareto-optimal minimal intervention set. Note that the condition $\nexists\, Y_i, Y_j \in \mathbf{Y}$ such that $Y_j \in \mathsf{pa}(Y_i)$
(i.e., $\mathsf{pa}(\mathbf{Y}) \cap \mathbf{Y} = \emptyset$) is required, since reward nodes are
non-manipulable by definition and therefore cannot be included in intervention sets. This completes the proof of Proposition \ref{prop2}.

\subsection{Projection of Causal Graph When Some Nodes Are Non-Manipulable}
\label{proj}

In this subsection, we describe the projection procedure for a causal graph onto a subset of its vertices. Consider a causal graph $\mathcal{G} = \langle \mathbf{V}, \mathbf{E} \rangle$ with vertex set $\mathbf{V}$. We project it onto $\mathbf{V} \setminus \mathbf{T}$, where $\mathbf{T}$ is a non-empty set of nodes that are non-manipulable. The optimality of an intervention involving all nodes outside $\mathbf{T}$, i.e., $\mathbf{V} \setminus \mathbf{T}$, is preserved during the projection step. The procedure follows steps similar to those in \cite{lee2019structural}. 

We initialize a graph $\mathcal{H} = \langle \mathbf{V} \setminus \mathbf{T}, \emptyset \rangle$ and then add edges as follows:

\begin{enumerate}
    \item Add a directed edge from $V_i$ to $V_j$ in $\mathcal{H}$ if the directed edge
    $V_i \rightarrow V_j$ already exists in $\mathcal{G}$, or if there exists
    a directed path from $V_i$ to $V_j$ in $\mathcal{G}$ such that all
    intermediate nodes on the path belong to $\mathbf{T}$.
    
    \item Add a bi-directed edge between $V_i$ and $V_j$ in $\mathcal{H}$ if a bi-directed edge
    $V_i \leftrightarrow V_j$ already exists in $\mathcal{G}$, or if there exist
    directed paths from a common exogenous noise variable to both $V_i$ and $V_j$
    in $\mathcal{G}$ such that all intermediate nodes on these paths belong to
    $\mathbf{T}$.
\end{enumerate}

\begin{algorithm}[h]
\small
\DontPrintSemicolon
\SetKwFunction{FMain}{ComputeMUCTandPPOMIS}
\SetKwProg{Fn}{Function}{:}{}
\Fn{\FMain{$\mathcal{G},\mathbf{Y}$}}{
      MUCT$ \gets \mathbf{Y}$\\
      \vspace{0.1em}
      \While{some $Y_i \in \mathbf{Y}$ is confounded with $V_i \in \mathsf{an}(Y_i)_{\mathcal{G}}$  in graph $\mathcal{G}$ $\;$ or $\;$  $\mathsf{pa}(\mathbf{Y})_{\mathcal{G}} \cap \mathbf{Y} \neq \emptyset$}{
      $\forall \; Y_i\in \mathbf{Y}$, set $\mathbf{T}_i \gets$ nodes in $\mathsf{an}(Y_i)_{\mathcal{G}}$ confounded with reward node $Y_i$ in graph $\mathcal{G}$. \\
      $\mathbf{T} \gets \bigcup_{i:Y_i \in \mathbf{Y} } \mathbf{T}_i$ . \\

      \If{$\mathbf{T} \neq \emptyset$}{
         Obtain the projection graph $\mathcal{H}$ on $\mathbf{V}_{\mathcal{G}} \setminus \mathbf{T}$ (Section \ref{proj}).  \\
      Updates: $\mathcal{G} \gets \mathcal{H},$ MUCT $\gets$ MUCT $\cup\; \mathbf{T}$.
      \\ Update: $\mathbf{Y} \gets \mathbf{Y} \cap \mathbf{V}_{\mathcal{G}}$ .
    }
  
    \vspace{0.5em}
    
      \If{$\mathsf{pa}(\mathbf{Y})_{\mathcal{G}} \cap \mathbf{Y} \neq \emptyset$}{
          $\mathbf{Z} \gets \mathsf{pa}(\mathbf{Y})_{\mathcal{G}} \cap \mathbf{Y} $\\ 
          Obtain the projection graph $\mathcal{H}$ on $\mathbf{V}_{\mathcal{G}} \setminus \mathbf{Z}$ (Section \ref{proj}).  \\
      Update: $\mathcal{G} \gets \mathcal{H}$
      \\ Update: $\mathbf{Y} \gets \mathbf{Y} \cap \mathbf{V}_{\mathcal{G}}$ .
    }
    \vspace{0.2em}
      \tcp*[f]{find the subset of reward nodes $\mathbf{Y}$ that are still present in resulting Graph $\mathcal{G}\;.\;\;\;\;$}\\
      $\;\;$Update: $\mathbf{Y} \gets \mathbf{Y} \cap \mathbf{V}_{\mathcal{G}}$ .\\  
    }
    \vspace{0.2em}

  Set $\text{PPOMIS} \gets \mathsf{pa}(\mathbf{Y})$ in the resulting graph $\mathcal{G}$ .\\ 
    \textbf{return}  MUCT, PPOMIS
}
\textbf{End Function}

\caption{Compute  MUCT and PPOMIS from Graph $\mathcal{G}$ \textbf{(Full Version)}}
\label{alg:pomis_muct_full}
\end{algorithm}

\subsection{Proof of Proposition \ref{prop3}}

\vspace{0.5em}
\propobs*
\vspace{0.5em}

\textbf{Proof: } If, in the causal graph $\mathcal{G}$, no reward node $Y_i \in \mathbf{Y}$ is confounded with its ancestors $\mathsf{an}(Y_i)$ and
$\mathsf{pa}(\mathbf{Y}) \cap \mathbf{Y} = \emptyset$, then the algorithm terminates immediately, as the
\texttt{while}-loop condition in \texttt{ComputeMUCTandPPOMIS} is false. In this case,
\(
\operatorname{PPOMIS}(\mathcal{G}, \mathbf{Y}) = \mathsf{pa}(\mathbf{Y}),
\)
as returned by Algorithm~\ref{alg:pomis_muct} or Algorithm~\ref{alg:pomis_muct_full}.
Moreover, by Proposition~\ref{prop2}, under these conditions the parent set
$\mathsf{pa}(\mathbf{Y})$ is only  Pareto-optimal minimal intervention set (PPOMIS).

We now analyze the scenario in which the \texttt{while}-loop condition is true. This can occur in two cases. The first case arises when the causal graph $\mathcal{G}$ contains at least one reward node $Y_i \in \mathbf{Y}$ that is confounded with at least one of its ancestors in $\mathsf{an}(Y_i)$. In this case, the condition of the while loop in the function call \texttt{ComputeMUCTandPPOMIS}($\mathcal{G},\mathbf{Y}$) will be true. Now consider a reward node $Y_i \in \mathbf{Y}$ such that there exists a subset of nodes in its ancestors, $\mathbf{T}_i \subseteq \mathsf{an}(Y_i)$, with the property that there is a latent confounder, i.e., a bi-directed edge between every node $t_j \in \mathbf{T}_i$ and the reward node $Y_i$, and moreover $\mathbf{T}_i \neq \emptyset$. There will be at least one such reward node $Y_i$ because the condition of the while-loop was true. We now proceed to specify a structural causal model $\mathcal{M}$ compatible with the causal graph. In this model, if we intervene on any of the nodes in the set $\mathbf{T}_i$, we will always be suboptimal with respect to the corresponding reward node $Y_i$. We use the notation $\mathsf{De}(\mathbf{T}_i)$ to denote the set $\mathbf{T}_i \cup \mathsf{de}(\mathbf{T}_i)$. We also use the notation
\(
\mathbf{P}_i := \bigl( \mathsf{an}(Y_i) \setminus \operatorname{De}(\mathbf{T}_i) \bigr) \cap \mathsf{pa}(Y_i),
\)
which denotes the subset of parent nodes of $Y_i$ such that neither these nodes nor their ancestors have a bi-directed edge to $Y_i$. We assume that all nodes in $\mathsf{De}(\mathbf{T}_i)$ consist of at least $|\mathbf{T}_i|+1$ bits, and all reward nodes are binary. Consider a node $t_j \in \mathbf{T}_i$ for any $j \in [|\mathbf{T}_i|]$, and let $U_{t_j}$ be the corresponding unobserved confounder that has edges to both $Y_i$ and $t_j$. We consider a model $\mathcal{M}$ where $U_{t_j}$ is a Bernoulli random variable with some fixed parameter, and the $j$-th bit of $t_j$ is set equal to $U_{t_j}$. Moreover, for every node in $\mathsf{De}(t_j)\setminus \mathbf{Y}$, the $j$-th bit is set equal to the $j$-th bit of its immediate parent. The reward $Y_i$ takes the value $0$ if the value of the unobserved confounder $U_{t_j}$ is not equal to the $j$-th bit of any of the nodes in $\mathsf{pa}(Y_i)\cap \mathsf{De}(t_j)$. We repeat the same construction for all nodes $t_j\in \mathbf{T}_i$ for all $j\in[|\mathbf{T}_i|]$. Thus, the reward $Y_i$ takes the value $0$ if the value of $U_{t_j}$ is not equal to the $j$-th bit of any of the nodes in $\mathsf{pa}(Y_i)\cap \mathsf{De}(t_j)$ for any $j\in[|\mathbf{T}_i|]$; otherwise, the reward node $Y_i$ is the AND-operation of its parent nodes in the set $\mathbf{P}_i$. This construction implies that if we intervene on any node $t_j \in \mathbf{T}_i$, then the $j$-th bit of $t_j$ is no longer equal to the unobserved confounder $U_{t_j}$, and therefore $Y_i$ takes the value $0$, because $U_{t_j}$ does not match the $j$-th bit of the nodes in $\mathsf{pa}(Y_i)\cap \mathsf{De}(t_j)$. This implies that we will be suboptimal on the reward node $Y_i$ whenever we intervene on any node in $\mathbf{T}_i$ for this possible  structural causal model ($\mathcal{M}$) construction. This construction extends to any reward node in the set $\mathbf{Y}$. That is, for every reward node $Y_i\in \mathbf{Y}$, intervening on any node in its corresponding set $\mathbf{T}_i$ will result in sub-optimality for the reward node $Y_i$. Overall, this implies that there exists a model $\mathcal{M}$ compatible with the underlying causal graph such that intervening on any node in the set $\mathbf{T}:=\bigcup_{i\in[d]} \mathbf{T}_i$ will necessarily result in sub-optimality on one or more reward nodes. Thus, there exists at least one PPOMIS in which none of the nodes in the set $\mathbf{T}$ is being intervened on.

Whenever the condition in the while loop is true, the algorithm constructs the set \(\mathbf{T} = \bigcup_{i \in [d]} \mathbf{T}_i\), which consists of all nodes among the ancestors of the reward nodes \(Y_i\) (i.e., \(\mathsf{an}(Y_i)\)) that are confounded with the corresponding reward nodes $Y_i \in \mathbf{Y}$. In the previous passage, we demonstrated the existence of a model $\mathcal{M}$ compatible with the causal graph \(\mathcal{G}\) such that intervening on any node in the set $\mathbf{T}$ results in sub-optimality on one or more reward nodes. Thus, there exists at least one PPOMIS for which none of the nodes in $\mathbf{T}$ is intervened on. To obtain that PPOMIS, we treat $\mathbf{T}$ as the set of non-intervenable nodes and compute the projection graph $\mathcal{H}$ on the rest of the nodes, starting with an empty graph on the nodes $\mathbf{V} \setminus \mathbf{T}$ (where $\mathbf{V} = \mathbf{V}_{\mathcal{G}}$), via the following steps.

\begin{enumerate}
    \item Add a directed edge from $V_i$ to $V_j$ in $\mathcal{H}$ if the directed edge
    $V_i \rightarrow V_j$ already exists in $\mathcal{G}$, or if there exists
    a directed path from $V_i$ to $V_j$ in $\mathcal{G}$ such that all
    intermediate nodes on the path belong to $\mathbf{T}$.
    
    \item Add a bi-directed edge between $V_i$ and $V_j$ in $\mathcal{H}$ if a bi-directed edge
    $V_i \leftrightarrow V_j$ already exists in $\mathcal{G}$, or if there exist
    directed paths from a common exogenous noise variable to both $V_i$ and $V_j$
    in $\mathcal{G}$ such that all intermediate nodes on these paths belong to
    $\mathbf{T}$.
\end{enumerate}

 We now show that given a model $\mathcal{M}^1 = \mathcal{M} = (\mathbf{V}, \mathbf{U}, F, P(\mathbf{U}))$ that conforms to $\mathcal{G}$, there exists a model $\mathcal{M}^2 = \mathcal{M}_{\mathbf{V}\setminus \mathbf{T}} = (\mathbf{V}\setminus \mathbf{T}, \mathbf{U}', F', P(\mathbf{U}'))$ that conforms to the projection graph \(\mathcal{H}\) such that $p^{\mathcal{M}_1}(\mathbf{y} \mid do(\mathbf{x})) = p^{\mathcal{M}_2}(\mathbf{y} \mid do(\mathbf{x}))$ for any $\mathbf{X}, \mathbf{Y} \subseteq \mathbf{V}\setminus \mathbf{T}$ and $\mathbf{Y} \neq \emptyset$. Also, for a model $\mathcal{M}^1 = \mathcal{M}_{[\mathbf{V} \setminus \mathbf{T}]} = \langle \mathbf{V} \setminus \mathbf{T}, \mathbf{U}', F', P(\mathbf{U}') \rangle$ conforming to $\mathcal{H}$, there exists a model $\mathcal{M}^2 = \mathcal{M} = \langle \mathbf{V}, \mathbf{U}, F, P(\mathbf{U}) \rangle$ that conforms to $\mathcal{G}$ such that $p^{\mathcal{M}_1}(\mathbf{y} \mid do(\mathbf{x})) = p^{\mathcal{M}_2}(\mathbf{y} \mid do(\mathbf{x}))$, for any $\mathbf{X}, \mathbf{Y} \subseteq \mathbf{V} \setminus \mathbf{T}$ and $\mathbf{Y} \neq \emptyset$. This guarantees that the optimality of an intervention involving all nodes outside $\mathbf{T}$, i.e., $\mathbf{V} \setminus \mathbf{T}$, is preserved during the projection step.

Starting from a model $\mathcal{M}^1=\mathcal{M}=(\mathbf{V},\mathbf{U},F,P(\mathbf{U}))$ conforming to $\mathcal{G}_{[\mathbf{V}]}$, we construct a model $\mathcal{M}_{\mathbf{V}\setminus\{t\}}=(\mathbf{V}\setminus\{t\},\mathbf{U}'',F'',P(\mathbf{U}''))$ conforming to the projected graph $\mathcal{G}_{[\mathbf{V}\setminus\{t\}]}$, for some $t\in\mathbf{T}$. The structural equations of all nodes $X\notin\mathsf{ch}(t)$ remain unchanged from $\mathcal{M}^1$ to $\mathcal{M}_{\mathbf{V}\setminus\{t\}}$. For nodes $X\in\mathsf{ch}(t)$, the structural equations in $\mathcal{M}_{\mathbf{V}\setminus\{t\}}$ are defined as $f''_X(\mathsf{pa}(X)_{\mathcal{G}_{[\mathbf{V}\setminus\{t\}]}},\mathbf{u}''_X)=f_X(\mathsf{pa}(X)_{\mathcal{G}_{[\mathbf{V}]}},f_t(\mathsf{pa}(t)_{\mathcal{G}_{[\mathbf{V}]}},\mathbf{u}_t),\mathbf{u}_X)$. Note that for $X\in\mathsf{ch}(t)$ we have
$\mathsf{pa}(X)_{\mathcal{G}_{[\mathbf{V}\setminus\{t\}]}}
=
\big(\mathsf{pa}(X)_{\mathcal{G}_{[\mathbf{V}]}} \setminus \{t\}\big)
\cup
\mathsf{pa}(t)_{\mathcal{G}_{[\mathbf{V}]}}$,
and $\mathbf{U}''_X = \mathbf{U}_X \cup \mathbf{U}_t$. This construction guarantees that both models $\mathcal{M}^1$ and $\mathcal{M}_{\mathbf{V}\setminus\{t\}}$ induce the same interventional distributions over all nodes other than $t$. By iteratively applying this procedure to all nodes in $\mathbf{T}$, we can obtain an equivalent model $\mathcal{M}^2=\mathcal{M}_{\mathbf{V}\setminus\mathbf{T}}=(\mathbf{V}\setminus\mathbf{T},\mathbf{U}',F',P(\mathbf{U}'))$ that conforms to the projected graph $\mathcal{H}$, such that $p^{\mathcal{M}_1}(\mathbf{y}\mid do(\mathbf{x}))=p^{\mathcal{M}_2}(\mathbf{y}\mid do(\mathbf{x}))$ for any $\mathbf{X},\mathbf{Y}\subseteq\mathbf{V}\setminus\mathbf{T}$ with $\mathbf{Y}\neq\emptyset$.

Starting from a model $\mathcal{M}_{[\mathbf{V} \setminus \{t\}]} = \langle \mathbf{V} \setminus \{t\}, \mathbf{U}'', F'', P(\mathbf{U}'') \rangle$ conforming to the projection graph $\mathcal{G}_{[\mathbf{V} \setminus \{t\}]}$, we show that one can construct a model $\mathcal{M}_{[\mathbf{V}]} = \langle \mathbf{V}, \mathbf{U}, F, P(\mathbf{U}) \rangle$ conforming to the graph $\mathcal{G}_{[\mathbf{V}]}$ such that they induce the same interventional distributions over all nodes other than $t$. For all nodes $X \notin \mathsf{ch}(t)_{\mathcal{G}_{[\mathbf{V}]}}$, we reuse the structural equations $f''_X$ from $\mathcal{M}_{[\mathbf{V} \setminus \{t\}]}$ in $\mathcal{M}_{[\mathbf{V}]}$. For any node $X \in \mathsf{ch}(t)_{\mathcal{G}_{[\mathbf{V}]}} \cup \{t\}$, we add any exogenous noise $U_i$ to $\mathbf{U}_X$ when $U_i \in \mathbf{U}''_X$ and $U_i$ has only one endogenous child, i.e., $X$. If a confounder exists between two nodes $V_i, V_j \in \mathsf{ch}(t)_{\mathcal{G}_{[\mathbf{V}]}}$ in $\mathcal{G}_{[\mathbf{V}]}$, then every exogenous noise variable 
$U \in \mathbf{U}_{V_i}'' \cap \mathbf{U}_{V_j}''$ is included in both $\mathbf{U}_{V_i}$ and $\mathbf{U}_{V_j}$ in the new model $\mathcal{M}_{[\mathbf{V}]}$. 
For every pair of nodes $V_i, V_j$ such that $V_i, V_j \in \mathsf{ch}(t)_{\mathcal{G}_{[\mathbf{V}]}}$, the projected graph 
$\mathcal{G}_{[\mathbf{V}\setminus\{t\}]}$ contains an unobserved confounder between $V_i$ and $V_j$. The exogenous noise variable corresponding to this 
confounder in $\mathcal{M}_{[\mathbf{V}\setminus\{t\}]}$ is added to  $U_{t}$ in the constructed model $\mathcal{M}_{[\mathbf{V}]}$.  Moreover, if $V_i$ is confounded with $t$ and $V_j$ is a child of $t$ in $\mathcal{G}_{[\mathbf{V}]}$, then marginalizing $t$ induces an unobserved confounder between 
$V_i$ and $V_j$ in the projected graph $\mathcal{G}_{[\mathbf{V}\setminus\{t\}]}$. Accordingly, the exogenous noise variable corresponding to this unobserved 
confounder in $\mathcal{M}_{[\mathbf{V}\setminus\{t\}]}$ is added to both $\mathbf{U}_{t}$ and $\mathbf{U}_{V_i}$ in the constructed model $\mathcal{M}_{[\mathbf{V}]}$, 
ensuring that all exogenous noise terms are correctly preserved during the transformation. The structural equation for any node $X \in \mathsf{ch}(t)_{\mathcal{G}_{[\mathbf{V}]}}$ in $\mathcal{M}_{[\mathbf{V}]}$ takes the form 
$f_X(\mathsf{pa}(X)_{\mathcal{G}_{[\mathbf{V}]}}, \mathbf{u}_X) = f_X''(\mathsf{pa}(X)_{\mathcal{G}_{[\mathbf{V}]}} \setminus \{t\}, f_t^{-1}(t), \mathbf{u}''_X)$, 
where $f_t$ is any one-to-one function whose argument can be passed to its children through $f_t^{-1}$, 
ensuring that both $\mathcal{M}_{[\mathbf{V} \setminus \{t\}]}$ and $\mathcal{M}_{[\mathbf{V}]}$ induce the same interventional distributions 
over all nodes other than $t$. Starting from the model 
$\mathcal{M}_1 = \mathcal{M}_{[\mathbf{V} \setminus \mathbf{T}]} 
= \langle \mathbf{V} \setminus \mathbf{T}, \mathbf{U}', F', P(\mathbf{U}') \rangle$,
which conforms to projected graph $\mathcal{H}$, one can iteratively apply the procedure described earlier by adding the nodes from set $\mathbf{T}$ one at a time. This process yields an equivalent model 
$\mathcal{M}_2 = \mathcal{M} = \langle \mathbf{V}, \mathbf{U}, F, P(\mathbf{U}) \rangle$ that conforms to the graph $\mathcal{G}$ and satisfies
\(
p^{\mathcal{M}_1}(\mathbf{y} \mid do(\mathbf{x})) 
= 
p^{\mathcal{M}_2}(\mathbf{y} \mid do(\mathbf{x}))
\) for all $\mathbf{X}, \mathbf{Y} \subseteq \mathbf{V} \setminus \mathbf{T}$ with $\mathbf{Y} \neq \emptyset$. Thus, the optimality of an intervention involving all nodes outside $\mathbf{T}$, 
that is, $\mathbf{V}_{\mathcal{G}} \setminus \mathbf{T}$, is preserved during the projection step. 
The function \texttt{ComputeMUCTandPPOMIS}$(\mathcal{G}, \mathbf{Y})$ in Algorithm~\ref{alg:pomis_muct_full} 
then adds the set $\mathbf{T}$ to MUCT, which is initialized as the set of reward nodes $\mathbf{Y}$, 
and updates the graph $\mathcal{G}$ to the corresponding projection graph $\mathcal{H}$.

The second case in which the \texttt{while}-loop condition remains true arises when there exist 
$Y_i, Y_j \in \mathbf{Y}$ such that $Y_j \in \mathsf{pa}(Y_i)$, which implies that 
$\mathsf{pa}(\mathbf{Y})_{\mathcal{G}} \cap \mathbf{Y} \neq \emptyset$. 
Since, by definition, all reward nodes $Y_i \in \mathbf{Y}$ are non-manipulable, 
this situation necessitates applying the projection operator once again by treating the nodes in the set 
$\mathbf{Z} := \mathsf{pa}(\mathbf{Y})_{\mathcal{G}} \cap \mathbf{Y}$ as non-manipulable and projecting the graph 
using the same procedure described earlier. 
To build intuition, consider a pair of reward nodes $Y_i, Y_j \in \mathbf{Y}$ such that 
$Y_j \in \mathsf{pa}(Y_i)$. One may conceptually merge these two reward nodes into a single node 
without affecting the interventional distributions; this is precisely the effect of the projection operation. 
Accordingly, when $\mathbf{Z} := \mathsf{pa}(\mathbf{Y}) _{\mathcal{G}}\cap \mathbf{Y} \neq \emptyset$, 
we project the graph $\mathcal{G}$ onto the vertex set 
$\mathbf{V}_{\mathcal{G}} \setminus \mathbf{Z}$ and update $\mathcal{G}$ to the resulting projection graph. 
As a result, the optimality of interventions involving all nodes outside $\mathbf{Z}$ is preserved 
throughout this projection step.

Since the goal of \texttt{ComputeMUCTandPPOMIS} is to construct a single PPOMIS, 
the \texttt{while} loop repeatedly checks whether there exists a reward node 
$Y_i \in \mathbf{Y}$ that remains confounded with its ancestors $\mathsf{an}(Y_i)$, 
or whether $\mathsf{pa}(\mathbf{Y})_{\mathcal{G}} \cap \mathbf{Y} \neq \emptyset$, 
in the graph obtained after each iteration. 
Whenever the \texttt{while}-loop condition holds, the algorithm recomputes 
$\mathbf{T} = \bigcup_{i \in [d]} \mathbf{T}_i$, updates MUCT by adding $\mathbf{T}$, 
and updates the graph $\mathcal{G}$ to the new projection graph $\mathcal{H}$ 
on the node set $\mathbf{V}_{\mathcal{G}} \setminus \mathbf{T}$. 
In the case where $\mathbf{Z} := \mathsf{pa}(\mathbf{Y})_{\mathcal{G}} \cap \mathbf{Y} \neq \emptyset$, 
the algorithm additionally projects $\mathcal{G}$ onto the vertex set 
$\mathbf{V}_{\mathcal{G}} \setminus \mathbf{Z}$.
This iterative process continues until the base case is reached, namely, when no reward node is confounded 
with its ancestors and the parent set $\mathsf{pa}(\mathbf{Y})_{\mathcal{G}}$ contains no reward nodes. 
At this point, the single PPOMIS can be read directly from the resulting graph as 
$\mathsf{pa}(\mathbf{Y})$, as established in Proposition~\ref{prop2}. 
This completes the proof of Proposition~\ref{prop3}.

\subsection{Proof of Lemma \ref{lem1}:}
\vspace{0.5em}
\lem*
\vspace{0.5em}
\textbf{Proof:} Consider $\mathbf{W} \subseteq \mathbf{V} \setminus \mathbf{Y}$. Any node $\mathbf{X}$ that is not an ancestor of any reward node, i.e., $X \notin \mathsf{an}(\mathbf{Y})$, plays no role in either MUCT or PPOMIS. Thus, 
$\text{MUCT}(\mathcal{G}_{\overline{\mathbf{W}}},\mathbf{Y}) = \text{MUCT}(\mathcal{G}_{\overline{\mathbf{W} \cap \mathsf{an}(\mathbf{Y})}},\mathbf{Y})$ 
and 
$\text{PPOMIS}(\mathcal{G}_{\overline{\mathbf{W}}},\mathbf{Y}) = \text{PPOMIS}(\mathcal{G}_{\overline{\mathbf{W} \cap \mathsf{an}(\mathbf{Y})}},\mathbf{Y})$. Let $\mathbf{W}' := \mathbf{W} \cap \mathsf{an}(\mathbf{Y})$. We can partition $\mathbf{W}'$ into two disjoint sets: 
\( \mathbf{W}' = (\mathbf{W}' \cap \mathbf{T}) \cup (\mathbf{W}' \setminus \mathbf{T}), \) 
where $\mathbf{T} = \text{MUCT}(\mathcal{G},\mathbf{Y})$. For any node $X \in \mathbf{W}' \setminus \mathbf{T}$, by the construction of $\mathbf{T} = \text{MUCT}(\mathcal{G},\mathbf{Y})$ (via Algorithm \ref{alg:pomis_muct} or Algorithm \ref{alg:pomis_muct_full} ), the node $X$ does not have any bi-directed edge or latent confounder with any reward node $Y_i \in \mathbf{Y}$ in any iteration. During all iterations of the while-loop in Algorithm \ref{alg:pomis_muct} or Algorithm \ref{alg:pomis_muct_full}, such a node $\mathbf{X}$ would have been added to $\mathbf{T}$ if it ever became confounded with any $Y_i$; however, since $X \notin \mathbf{T}$, this never occurs. This yields a contradiction if we assume otherwise. Moreover, since $\mathcal{G}_{\overline{\mathbf{W}' \cap \mathbf{T}}}$ is a sparser graph than $\mathcal{G}$, the node $X$ also cannot acquire a bi-directed edge or latent confounder with any $Y_i$ in any subsequent iteration of the while-loop when computing $\text{MUCT}(\mathcal{G}_{\overline{\mathbf{W}' \cap \mathbf{T}}},\mathbf{Y})$. Therefore, removing the incoming edges of nodes in $\mathbf{W}' \setminus \mathbf{T}$, in addition to removing those for nodes in $\mathbf{W}' \cap \mathbf{T}$, does not affect the resulting MUCT or the corresponding PPOMIS. Also note that $\mathbf{W}' \cap \mathbf{T} = \mathbf{W} \cap \mathsf{an}(\mathbf{Y}) \cap \mathbf{T} = \mathbf{W} \cap \mathbf{T}$ since $\mathbf{T} \subseteq \mathsf{an}(\mathbf{Y})$.

Hence, 
\[
\text{MUCT}(\mathcal{G}_{\overline{\mathbf{W}}},\mathbf{Y})
= \text{MUCT}(\mathcal{G}_{\overline{\mathbf{W} \cap \mathsf{an}(\mathbf{Y})}},\mathbf{Y})
= \text{MUCT}(\mathcal{G}_{\overline{\mathbf{W} \cap \mathbf{T}}},\mathbf{Y}),
\] 
and similarly, 
\[
\text{PPOMIS}(\mathcal{G}_{\overline{\mathbf{W}}},\mathbf{Y})
= \text{PPOMIS}(\mathcal{G}_{\overline{\mathbf{W} \cap \mathsf{an}(\mathbf{Y})}},\mathbf{Y})
= \text{PPOMIS}(\mathcal{G}_{\overline{\mathbf{W} \cap \mathbf{T}}},\mathbf{Y}).
\]  

This completes the proof of Lemma \ref{lem1}.

\subsection{Proof of Proposition \ref{prop4}}
\vspace{0.5em}
\propint*
\vspace{0.5em}

\begin{algorithm}[h]
\small
\DontPrintSemicolon
\SetKwFunction{FMain}{ComputeMUCTandPPOMIS}
\SetKwProg{Fn}{Function}{:}{}
\Fn{\FMain{$\mathcal{G}, \mathbf{W}, \mathbf{Y}$}}{
  $\mathcal{G}_0 \gets \mathcal{G}, \quad \mathbf{Y}_0 \gets \mathbf{Y}$\\
  \vspace{0.3em}
  \While{any reward node $Y_i \in \mathbf{Y}$ is confounded with one of its ancestors $V_i \in \mathsf{an}(Y_i)_{\mathcal{G}}$ in Graph $\mathcal{G}$ $\;$ or $\;$  $\mathsf{pa}(\mathbf{Y})_{\mathcal{G}} \cap \mathbf{Y} \neq \emptyset$}{\vspace{0.2em}
    $\forall \; Y_i\in \mathbf{Y}$, set $\mathbf{T}_i \gets$ nodes in $\mathsf{an}(Y_i)_{\mathcal{G}}$ confounded with reward node $Y_i$ in graph $\mathcal{G}$. \\
     $\mathbf{T} \gets \bigcup_{i:Y_i \in \mathbf{Y} } \mathbf{T}_i$ . \\

    \eIf{$\mathbf{T} \cap \mathbf{W} \neq \emptyset$}{
      $\mathcal{G} \gets \mathcal{G}_{\overline{\mathbf{T} \cap \mathbf{W}}}$\\
     \tcp*[f]{restart the while-loop with the updated graph}
    }{
            \If{$\mathbf{T} \neq \emptyset$}{
         Obtain the projection graph $\mathcal{H}$ on $\mathbf{V}_{\mathcal{G}} \setminus \mathbf{T}$ (Section \ref{proj}).  \\
      Update: $\mathcal{G} \gets \mathcal{H}$
      \\ Update: $\mathbf{Y} \gets \mathbf{Y} \cap \mathbf{V}_{\mathcal{G}}$ .
    }
  
    \vspace{0.5em}
    
      \If{$\mathsf{pa}(\mathbf{Y})_{\mathcal{G}} \cap \mathbf{Y} \neq \emptyset$}{
          $\mathbf{Z} \gets \mathsf{pa}(\mathbf{Y})_{\mathcal{G}} \cap \mathbf{Y} $\\ 
          Obtain the projection graph $\mathcal{H}$ on $\mathbf{V}_{\mathcal{G}} \setminus \mathbf{Z}$ (Section \ref{proj}).  \\
      Update: $\mathcal{G} \gets \mathcal{H}$
      \\ Update: $\mathbf{Y} \gets \mathbf{Y} \cap \mathbf{V}_{\mathcal{G}}$ .
    }
    \vspace{0.2em}
      \tcp*[f]{find the subset of reward nodes $\mathbf{Y}$ that are still present in resulting Graph $\mathcal{G}.$}\\
      $\;\;$Update: $\mathbf{Y} \gets \mathbf{Y} \cap \mathbf{V}_{\mathcal{G}}$ .\\ 
    }

  }

  \vspace{0.3em}
  
  Set $\text{PPOMIS} \gets \mathsf{pa}(\mathbf{Y})$ in the resulting graph $\mathcal{G}$   \\
  $\text{MUCT} \gets \text{MUCT}((\mathcal{G}_0)_{\overline{\mathbf{W}}}, \mathbf{Y}_0)$ (via Algorithm~\ref{alg:pomis_muct_full})\\
  \textbf{return} $\text{MUCT}, \text{PPOMIS}$
}
\caption{Compute PPOMIS and MUCT from a causal graph $\mathcal{G}$ and a set $\mathbf{W}$ \textbf{(Full Version)}}
\label{alg:pomis_muct2}
\end{algorithm}

\textbf{Proof:} To establish that for any subset $\mathbf{W} \subseteq \text{MUCT}(\mathcal{G}, \mathbf{Y}) \setminus \mathbf{Y}$, the set 
$\text{PPOMIS}(\mathcal{G}_{\overline{\mathbf{W}}}, \mathbf{Y})$ is a possibly Pareto-optimal minimal intervention set, we consider the alternative procedure 
\texttt{ComputeMUCTandPPOMIS}$(\mathcal{G}, \mathbf{W}, \mathbf{Y})$ (formally described in Algorithm~\ref{alg:pomis_muct2}), which outputs two sets:
$\text{PPOMIS}(\mathcal{G}, \mathbf{W}, \mathbf{Y})$ and $\text{MUCT}(\mathcal{G}, \mathbf{W}, \mathbf{Y})$.  
We show that these outputs are identical to 
$\text{PPOMIS}(\mathcal{G}_{\overline{\mathbf{W}}}, \mathbf{Y})$ and $\text{MUCT}(\mathcal{G}_{\overline{\mathbf{W}}}, \mathbf{Y})$, respectively, as returned by Algorithm~\ref{alg:pomis_muct_full}.  
We also show that $\text{PPOMIS}(\mathcal{G}, \mathbf{W}, \mathbf{Y})$ is indeed a possibly Pareto-optimal minimal intervention set. 

The function in Algorithm~\ref{alg:pomis_muct2}, i.e., \texttt{ComputeMUCTandPPOMIS}$(\mathcal{G}, \mathbf{W}, \mathbf{Y})$, is similar to 
\texttt{ComputeMUCTandPPOMIS}$(\mathcal{G}, \mathbf{Y})$ (Algorithm~\ref{alg:pomis_muct_full}), except for the additional input set $\mathbf{W}$. In the while-loop, for any iteration, if there is an intersection between the set $\mathbf{T}$ and $\mathbf{W}$, i.e., if $\mathbf{T} \cap \mathbf{W} \neq \emptyset$, then the incoming edges into the nodes in $\mathbf{T} \cap \mathbf{W}$ are removed in the causal graph, and the rest of the statements in the loop (i.e., the two projection steps) are skipped. However, if there is no intersection of $\mathbf{T}$ with $\mathbf{W}$, then all steps are identical to those in the function \texttt{ComputeMUCTandPPOMIS}$(\mathcal{G}, \mathbf{Y})$ (Algorithm~\ref{alg:pomis_muct_full}). To show that $\text{PPOMIS}(\mathcal{G},\mathbf{W}, \mathbf{Y})$ is indeed a possibly Pareto-optimal minimal intervention set, we follow an approach similar to the proof that $\text{PPOMIS}(\mathcal{G}, \mathbf{Y})$ is a possibly Pareto-optimal minimal intervention set (i.e., Proposition \ref{prop3}). If, in the causal graph $\mathcal{G}$, no node $Y_i \in \mathbf{Y}$ is confounded with its ancestors $\mathsf{an}(Y_i)$ and $\mathsf{pa}(\mathbf{Y}) \cap \mathbf{Y} = \emptyset$, then
\(
\text{PPOMIS}(\mathcal{G},\mathbf{W}, \mathbf{Y}) = \operatorname{pa}(\mathbf{Y}),
\)
as returned by the call \texttt{ComputeMUCTandPPOMIS}$(\mathcal{G}, \mathbf{W},\mathbf{Y})$ (Algorithm~\ref{alg:pomis_muct2}). Moreover, by Proposition \ref{prop2}, when no $Y_i$ is confounded with $\mathsf{an}(Y_i)$ and $\mathsf{pa}(\mathbf{Y}) \cap \mathbf{Y} = \emptyset$, the parent set $\operatorname{pa}(\mathbf{Y})$ is the unique Pareto-optimal minimal intervention set.

We now analyze the scenario in which the while-loop condition is true. This can occur in two cases. The first case arises when the causal graph $\mathcal{G}$ there exists at least one reward node $Y_i \in \mathbf{Y}$ that is confounded with some node in its ancestor set $\mathsf{an}(Y_i)$. In this situation the condition of the while-loop in the call 
\texttt{ComputeMUCTandPPOMIS}$(\mathcal{G}, \mathbf{W}, \mathbf{Y})$ (Algorithm~\ref{alg:pomis_muct2}) is satisfied.  
Consider such a reward node $Y_i \in \mathbf{Y}$ for which there exists a subset of ancestors $\mathbf{T}_i \subseteq \mathsf{an}(Y_i)$ such that each node $t_j \in \mathbf{T}_i$ shares a latent confounder (i.e., a bi-directed edge) with the reward node $Y_i$, and moreover $\mathbf{T}_i \neq \emptyset$. There must be at least one such reward node $Y_i$, because the condition of the while-loop was triggered. There are two sub-cases. The first sub-case is when $\mathbf{T} \cap \mathbf{W} \neq \emptyset$. In this case, let  
$\mathbf{T}' := \mathbf{T} \cap \mathbf{W}$ and $\mathbf{T}_i' := \mathbf{T}_i \cap \mathbf{W}$. We now proceed to specify a structural causal model (SCM) that is compatible with the causal graph. In this model, if we do not intervene on any of the nodes in the set $\mathbf{T}_i'$ or any of their descendants that lie on a directed path to $Y_i$, then we will always be suboptimal with respect to the reward node $Y_i$. We use the notation $\mathsf{De}(\mathbf{T}_i')$ to denote the set $\mathbf{T}_i' \cup \mathsf{de}(\mathbf{T}_i')$. We also define
\(
\mathbf{P}_i := \bigl( \mathsf{an}(Y_i) \setminus \mathsf{De}(\mathbf{T}_i') \bigr) \cap \operatorname{pa}(Y_i),
\)
which denotes the subset of parents of $Y_i$ such that neither these nodes nor their ancestors have
a bi-directed edge to $Y_i$. We assume that all nodes in $\mathsf{De}(\mathbf{T}_i')$ consist of at least $|\mathbf{T}_i'|+1$ bits, and all reward nodes are binary. Consider any node $t_j' \in \mathbf{T}_i'$ for $j \in [|\mathbf{T}_i'|]$, and let $U_{t_j'}$ be the corresponding unobserved confounder that has edges to both $Y_i$ and $t_j'$. We consider an SCM where $U_{t_j'}$ is a Bernoulli random variable with some fixed parameter, and the $j$-th bit of $t_j'$ is set equal to $1 - U_{t_j'}$. Moreover, for every node in $\mathsf{De}(t_j') \setminus \mathbf{Y}$, the $j$-th bit is defined to be equal to the $j$-th bit of its immediate parent. The reward $Y_i$ takes value $0$ if the value of the unobserved confounder $U_{t_j'}$ is not equal to the $j$-th bit of any node in $\operatorname{pa}(Y_i) \cap \mathsf{De}(t_j')$. We repeat this construction for all nodes $t_j' \in \mathbf{T}_i'$ for all $j \in [|\mathbf{T}_i'|]$. Thus, the reward $Y_i$ takes the value $0$ if, for any $j \in [|\mathbf{T}_i'|]$, the confounder $U_{t_j'}$ does not match the $j$-th bit of any node in $\operatorname{pa}(Y_i) \cap \mathsf{De}(t_j')$. Otherwise,the reward node $Y_i$ is defined to be the AND-operation of its parent nodes in the set $\mathbf{P}_i$. This construction implies that if we do not intervene on any node $t_j' \in \mathbf{T}_i'$ or its descendants along a directed path to $Y_i$, then $Y_i$ necessarily takes the value $0$. Hence, for this possible SCM construction, we will always be suboptimal for the reward node $Y_i$ whenever we do not intervene on any node in $\mathbf{T}_i'$ or its descendants that are ancestors of $Y_i$. This construction extends to every reward node in the set $\mathbf{Y}$. That is, for each reward node $Y_i \in \mathbf{Y}$, failing to intervene on any node in its corresponding set $\mathbf{T}_i'$ or its descendants along a directed path to $Y_i$ results in sub-optimality for $Y_i$. Overall, this shows that there exists an SCM compatible with the underlying causal graph such that not intervening on any node in the set $\mathbf{T}'$ or on their descendants along paths to the reward nodes necessarily results in sub-optimality on one or more reward nodes. Thus, we simply remove the incoming edges to the set $\mathbf{T}' := \mathbf{T} \cap \mathbf{W}$ and update the graph $\mathcal{G}$ to 
$\mathcal{G}_{\overline{\mathbf{T} \cap \mathbf{W}}}$. We then omit the projection steps and return to the beginning of the while-loop. The other sub-case is when $\mathbf{T} \cap \mathbf{W} = \emptyset$. In this case, we can replicate the projection step by computing the 
projection graph $\mathcal{H}$ on the remaining nodes outside the set $\mathbf{T}$, i.e., on $V_{\mathcal{G}} \setminus \mathbf{T}$. We have already established the soundness of this step in the proof of the Proposition \ref{prop3} which states that
$\text{PPOMIS}(\mathcal{G}, \mathbf{Y})$ (as returned by Algorithm~\ref{alg:pomis_muct_full}) is a possibly Pareto-optimal minimal intervention set—where we demonstrated the existence of an SCM compatible with the causal graph $\mathcal{G}$ such that intervening on any node in the set $\mathbf{T}$ results in 
sub-optimality on one or more reward nodes. Thus, there exists at least one PPOMIS for which none of the nodes in $\mathbf{T}$ is intervened on. To obtain that PPOMIS, we treat $\mathbf{T}$ as the set of non-intervenable nodes and compute the projection graph $\mathcal{H}$ accordingly. 

The second case in which the \texttt{while}-loop condition remains true arises when there exist
$Y_i, Y_j \in \mathbf{Y}$ such that $Y_j \in \mathsf{pa}(Y_i)$, which implies that
$\mathsf{pa}(\mathbf{Y})_{\mathcal{G}} \cap \mathbf{Y} \neq \emptyset$.
As explained in the proof of Proposition~\ref{prop3}, when
$\mathbf{Z} := \mathsf{pa}(\mathbf{Y})_{\mathcal{G}} \cap \mathbf{Y} \neq \emptyset$,
we project the graph $\mathcal{G}$ onto the vertex set
$\mathbf{V}_{\mathcal{G}} \setminus \mathbf{Z}$ and update $\mathcal{G}$ to the resulting projection graph.
As a result, the optimality of interventions involving all nodes outside $\mathbf{Z}$ is preserved
throughout this projection step. This overall process continues until the base case is reached, namely, when none of the reward nodes
is confounded with its ancestors and the parent set $\mathsf{pa}(\mathbf{Y})_{\mathcal{G}}$
contains no reward nodes. At that point, the single PPOMIS can be read directly from the resulting
graph as $\mathsf{pa}(\mathbf{Y})$, as shown in Proposition~\ref{prop2}.

Also note that in the function \texttt{ComputeMUCTandPPOMIS}$(\mathcal{G}, \mathbf{W}, \mathbf{Y})$ (Algorithm~\ref{alg:pomis_muct2}), we remove the incoming edges to the 
nodes in the set $\mathbf{W}$ whenever there is an intersection between the set $\mathbf{T}$ and $\mathbf{W}$ in the while-loop. This process is 
equivalent to simply removing the incoming edges to the set $\mathbf{W}$ at the beginning and then calling the function 
\texttt{ComputeMUCTandPPOMIS}$(\mathcal{G}_{\overline{\mathbf{W}}}, \mathbf{Y})$ (Algorithm~\ref{alg:pomis_muct} or Algorithm~\ref{alg:pomis_muct_full}). Thus, the sets returned by Algorithm~\ref{alg:pomis_muct2}, namely 
$\text{PPOMIS}(\mathcal{G}, \mathbf{W}, \mathbf{Y})$ and $\text{MUCT}(\mathcal{G}, \mathbf{W}, \mathbf{Y})$, are identical to the sets 
$\text{PPOMIS}(\mathcal{G}_{\overline{\mathbf{W}}}, \mathbf{Y})$ and $\text{MUCT}(\mathcal{G}_{\overline{\mathbf{W}}}, \mathbf{Y})$, respectively. Furthermore, since 
$\text{PPOMIS}(\mathcal{G}, \mathbf{W}, \mathbf{Y})$ is a possibly Pareto-optimal minimal intervention set, it follows that 
$\text{PPOMIS}(\mathcal{G}_{\overline{\mathbf{W}}}, \mathbf{Y})$ is also a possibly Pareto-optimal minimal intervention set. This completes the proof of Proposition \ref{prop4}.

\subsection{Proof of Theorem \ref{thm1}}

\thm*

\textbf{Proof:} \textbf{(If part)} We want to show that any possible set $\mathbf{W} \subseteq \mathbf{V} \setminus \mathbf{Y}$ is a PPOMIS if 
$\text{PPOMIS}(\mathcal{G}_{\overline{\mathbf{W}}}, \mathbf{Y}) = \mathbf{W}$. We have 
$\text{PPOMIS}(\mathcal{G}_{\overline{\mathbf{W}}}, \mathbf{Y}) = \text{PPOMIS}(\mathcal{G}_{\overline{\mathbf{W} \cap \mathbf{T}}}, \mathbf{Y})$, 
where $\mathbf{T} = \text{MUCT}(\mathcal{G}, \mathbf{Y})$ (Lemma \ref{lem1}). Let $\mathbf{W}' := \mathbf{W} \cap \mathbf{T}$, so that $\mathbf{W}' \subseteq \mathbf{T}$. By Proposition~\ref{prop4}, for any $\mathbf{W}' \subseteq \text{MUCT}(\mathcal{G}, \mathbf{Y}) \setminus \mathbf{Y}$, the set $\text{PPOMIS}(\mathcal{G}_{\overline{\mathbf{W}'}}, \mathbf{Y})$ is a possibly Pareto-optimal minimal intervention set. Note that 
$\mathbf{W} = \text{PPOMIS}(\mathcal{G}_{\overline{\mathbf{W}}}, \mathbf{Y}) = \text{PPOMIS}(\mathcal{G}_{\overline{\mathbf{W} \cap \mathbf{T}}}, \mathbf{Y}) = \text{PPOMIS}(\mathcal{G}_{\overline{\mathbf{W}'}}, \mathbf{Y})$, 
which confirms that $\mathbf{W}$ is indeed a possibly Pareto-optimal minimal intervention set.

\vspace{1em}

\textbf{(Only if part)}
We prove the contrapositive. Let
$\mathbf W\subseteq\mathbf V\setminus\mathbf Y$ satisfy
\[
\mathrm{PPOMIS}(\mathcal G_{\overline{\mathbf W}},\mathbf Y)
=\mathbf X\neq\mathbf W.
\]
If $\mathbf W\notin\mathbb M_{\mathcal G,\mathbf Y}$, it is not a
PPOMIS by Definition~\ref{defppomis}. Hence assume
$\mathbf W\in\mathbb M_{\mathcal G,\mathbf Y}$.
We show that $\mathbf X$ is another minimal intervention set and
that every randomized intervention on $\mathbf W$ can be matched
in all reward means by one randomized intervention on $\mathbf X$.

\medskip
\noindent\emph{Step 1: The final graph and minimality of $\mathbf X$.}
Write $\mathcal Q=\mathcal G_{\overline{\mathbf W}}$ and run
Algorithm~\ref{alg:pomis_muct_full} on $(\mathcal Q,\mathbf Y)$.
Throughout this proof, $\mathbf Y$ denotes the original reward
set. The revised Algorithm~\ref{alg:pomis_muct_full} updates its working reward set
immediately after each projection, so all subsequent graph
queries involve only reward vertices still present.

Each iteration removes at least one vertex, so the algorithm
terminates. Let $\mathbf R$ be all removed vertices, including
removed rewards, and set
\[
\mathbf S=\mathbf V\setminus\mathbf R,
\qquad
\mathbf Y_f=\mathbf Y\cap\mathbf S.
\]
Let $\mathcal H$ be the final graph. By Section~\ref{proj}, a projected
directed edge represents a directed path with removed internal
observed vertices; a projected bidirected edge represents paths
from a common exogenous cause with removed internal observed
vertices. Expanding intermediate projected edges yields such
paths in $\mathcal Q$ with internal vertices in $\mathbf R$;
splitting these paths at intermediate retained vertices gives
the converse. This applies to directed paths and to both paths
witnessing a bidirected edge. Hence $\mathcal H$ is the
projection of $\mathcal Q$ onto $\mathbf S$.
At termination,
\[
\mathbf X=\mathsf{pa}(\mathbf Y_f)_{\mathcal H},
\qquad
\mathbf X\cap\mathbf Y_f=\varnothing,
\]
and no $Y\in\mathbf Y_f$ has a bidirected edge to one of its
ancestors in $\mathcal H$.
Moreover, $\mathbf Y_f\neq\varnothing$: the finite acyclic graph
has an original reward with no other reward descendant. Such a
reward cannot enter either removal set,
since either removal requires a directed path to another reward.

Every vertex of $\mathbf W$ has no incoming directed edge or
incident bidirected edge in $\mathcal Q$. Projection preserves
these properties. Such a vertex cannot be removed for confounding,
and it is not a reward, so it cannot be removed as a reward parent.
Thus $\mathbf W\subseteq\mathbf S$.
Also, $\mathbf X\subseteq\mathbf S$ and
$\mathbf X\cap\mathbf Y_f=\varnothing$ imply
$\mathbf X\cap\mathbf Y=\varnothing$.
Each edge from $X\in\mathbf X$ to a retained reward expands to
a directed path in $\mathcal Q$ whose internal vertices lie in
$\mathbf R$, hence outside $\mathbf X$.
This path also exists in $\mathcal G_{\overline{\mathbf X}}$,
because deleting incoming edges to $\mathbf X$ removes none of
its edges. Consequently,
\[
\mathbf X\subseteq
\mathsf{an}(\mathbf Y)_{\mathcal G_{\overline{\mathbf X}}}.
\]
Proposition~\ref{prop1} gives $\mathbf X\in\mathbb M_{\mathcal G,\mathbf Y}$.

\medskip
\noindent\emph{Step 2: Recovering every original reward.}
We first show that every $R\in\mathbf R$ has a directed path
in $\mathcal Q$ to some $Y\in\mathbf Y_f$, with every vertex
except its endpoint in $\mathbf R$.
Use reverse topological induction in $\mathcal Q$: assume the
statement holds for all removed strict descendants of $R$,
including those removed in the same projection step.
All directed-path concatenations below remain paths, since
a repeated vertex would create a directed cycle.

If $R$ was removed as a reward parent, an edge $R\to Y_k$
existed at that step. Expand it into its path in $\mathcal Q$;
all internal vertices were already removed. If $Y_k$ survives,
this proves the claim. Otherwise, append the path supplied by
the induction hypothesis for the removed strict descendant $Y_k$.

If $R$ was removed for confounding, that step contained
$R\leftrightarrow Y_k$ and a directed path from $R$ to $Y_k$.
Expand the directed path and, if $Y_k$ is eventually removed,
append its inductive path to a retained reward $Y_\ell$.
If $Y_k$ survives, take $Y_\ell=Y_k$.
Suppose the resulting path has a first surviving vertex $V$
before $Y_\ell$.
The common exogenous cause witnessing $R\leftrightarrow Y_k$
has paths to $R$ and $Y_k$ through vertices already removed
at that step. Append the prefix $R\to\cdots\to V$ to its
first path and the suffix $Y_k\to\cdots\to Y_\ell$ to its
second path. Both extended paths have only removed internal
observed vertices. Hence $V\leftrightarrow Y_\ell$ appears
in $\mathcal H$. The directed suffix from $V$ to $Y_\ell$
also projects to a directed path in $\mathcal H$.
This contradicts termination, since $Y_\ell$ would be confounded
with an ancestor. Thus no such $V$ exists, proving the claim.

For each original reward $Y_i$, choose such a path $\rho_i$
to a retained reward $Y_{\ell(i)}$; if $Y_i$ survives, use
the length-zero path. Let $\mathcal H_i$ be the projection
of $\mathcal Q$ onto $\mathbf S\cup\{Y_i\}$.
We claim that
\begin{equation}
\mathsf{pa}(Y_i)_{\mathcal H_i}
\subseteq\mathbf X\cap\mathsf{an}(Y_i)_{\mathcal Q},
\label{eq:onlyif_auxiliary}
\end{equation}
and that $Y_i$ has no bidirected neighbor among its ancestors
in $\mathcal H_i$.
Indeed, a parent edge $V\to Y_i$ in $\mathcal H_i$ has
$V\in\mathbf S$. Expanding this edge in $\mathcal Q$ and
appending $\rho_i$ gives a directed path from $V$ to
$Y_{\ell(i)}$ with only removed internal vertices.
Thus $V\to Y_{\ell(i)}$ belongs to $\mathcal H$, giving
$V\in\mathbf X$; the first path also gives
$V\in\mathsf{an}(Y_i)_{\mathcal Q}$.
For the second claim, suppose $V\leftrightarrow Y_i$ in
$\mathcal H_i$ with $V$ an ancestor of $Y_i$. Again,
$V\in\mathbf S$. If $Y_i$ survives, this contradicts
termination directly. Otherwise, $\rho_i$ gives the edge
$Y_i\to Y_{\ell(i)}$ in $\mathcal H_i$. Projecting out
$Y_i$ therefore produces $V\leftrightarrow Y_{\ell(i)}$
in $\mathcal H$. The directed path from $V$ to $Y_i$,
followed by $\rho_i$, also projects to a directed path from
$V$ to $Y_{\ell(i)}$. This again contradicts termination.

\medskip
\noindent\emph{Do-calculus conventions.}
We use Pearl's Rules~2 and~3 of do-calculus~\citep{pearl2009causality} for disjoint sets
$\mathbf I$, $\mathbf J$, and $\{Y\}$.
In a graph $\mathcal K$, Rule~2 exchanges observing
$\mathbf J=\mathbf j$ with intervening on it, while retaining
$do(\mathbf I=\mathbf i)$, if
\[
(Y\perp\!\!\!\perp\mathbf J\mid\mathbf I)_
{\mathcal K_{\overline{\mathbf I},\underline{\mathbf J}}}.
\]
Rule~3 inserts or deletes $do(\mathbf J=\mathbf j)$, while
retaining $do(\mathbf I=\mathbf i)$, if
\[
(Y\perp\!\!\!\perp\mathbf J\mid\mathbf I)_
{\mathcal K_{\overline{\mathbf I},\overline{\mathbf J}}}.
\]
This form of Rule~3 applies because there are no ordinary
observations being conditioned on, so its ancestor restriction
leaves all of $\mathbf J$ in the deletion set.
An overline deletes incoming edges, including incident bidirected
edges; an underline deletes outgoing directed edges.
Mixed-graph separations are interpreted through their
latent-variable DAG representations. The rules imply the
corresponding equalities of reward expectations.

\medskip
\noindent\emph{Step 3: Applying Rule~2.}
Define the disjoint decompositions
\[
\mathbf W_0=\mathbf W\cap\mathbf X,
\qquad
\mathbf X'=\mathbf X\setminus\mathbf W,
\qquad
\mathbf Z=\mathbf W\setminus\mathbf X,
\]
so $\mathbf X=\mathbf W_0\cup\mathbf X'$ and
$\mathbf W=\mathbf W_0\cup\mathbf Z$.
For each $i\in[d]$, set
\[
\mathbf A_i=\mathbf X'\cap\mathsf{an}(Y_i)_{\mathcal Q},
\qquad
\mathbf B_i=\mathbf X'\setminus\mathbf A_i.
\]
Every member of $\mathbf A_i$ remains an ancestor of $Y_i$
in $\mathcal H_i$, because projection preserves directed
reachability between retained vertices.
By \eqref{eq:onlyif_auxiliary}, every parent of $Y_i$ in
$\mathcal H_i$ belongs to $\mathbf A_i\cup\mathbf W_0$.
We verify the Rule~2 condition
\begin{equation}
(Y_i\perp\!\!\!\perp\mathbf A_i\mid\mathbf W)_
{(\mathcal H_i)_{\overline{\mathbf W},
\underline{\mathbf A_i}}}.
\label{eq:onlyif_rule2_condition}
\end{equation}
The overline on $\mathbf W$ changes nothing, since its vertices
are already unconfounded roots.
These roots are neither colliders nor descendants of colliders.
Conditioning on them therefore cannot activate a collider, so an
active path from $Y_i$ to $\mathbf A_i$ would have to be
collider-free. Consider its first edge.
If it is $Y_i\leftarrow V$, then either $V\in\mathbf A_i$,
whose outgoing edge has been deleted, or $V\in\mathbf W_0$,
where conditioning blocks the path.
If it points out of $Y_i$, avoiding colliders forces the whole
path to be directed toward its endpoint in $\mathbf A_i$.
That is impossible: each such endpoint is an ancestor of $Y_i$,
so this would create a directed cycle.
Finally, if it is $Y_i\leftrightarrow V$, avoiding colliders
forces a directed continuation from $V$ to the endpoint
(or $V$ itself is the endpoint). Thus $V$ is an ancestor of
$Y_i$, contrary to Step~2.
These cases prove \eqref{eq:onlyif_rule2_condition}.

Fix an arbitrary SCM conforming to $\mathcal G$ and an assignment
$\mathbf w$ to $\mathbf W$. After $do(\mathbf W=\mathbf w)$,
project the model onto $\mathbf S\cup\{Y_i\}$ by substituting
the removed structural equations into their children's equations.
For each exogenous realization, this preserves all retained
variables, including under further interventions on retained nodes:
the intervened equations are replaced by the same constants in
both models, and substitution preserves the remaining equations'
values. Its graph is contained in $\mathcal H_i$.

We may therefore apply Rule~2 with
$\mathcal K=\mathcal H_i$, $\mathbf I=\mathbf W$, and
$\mathbf J=\mathbf A_i$. The condition was established in
\eqref{eq:onlyif_rule2_condition}, and preservation transfers
the resulting identity back to the original SCM:
\begin{equation}
\mathbb E[Y_i\mid do(\mathbf W=\mathbf w),\mathbf A_i=\mathbf a]
=
\mathbb E[Y_i\mid do(\mathbf W=\mathbf w),
do(\mathbf A_i=\mathbf a)].
\label{eq:onlyif_exchange}
\end{equation}
This identity is used only when the conditioning event has
positive probability.

\medskip
\noindent\emph{Step 4: Applying Rule~3 twice.}
First insert the actions on $\mathbf B_i$.
Use $\mathcal K=\mathcal G$,
$\mathbf I=\mathbf W\cup\mathbf A_i$, and
$\mathbf J=\mathbf B_i$.
The required graph is
$\mathcal G_{\overline{\mathbf W},\overline{\mathbf A_i},
\overline{\mathbf B_i}}$.
All intervention targets are unconfounded roots in this graph.
Also, $\mathbf B_i$ has no directed path to $Y_i$, because
its vertices are not ancestors of $Y_i$ in $\mathcal Q$ and
further edge deletion cannot create a path.
Conditioning on the roots $\mathbf W\cup\mathbf A_i$ cannot
activate a collider. Any collider-free path starting at a root
in $\mathbf B_i$ must follow directed edges away from it,
and no such path reaches $Y_i$. Hence
\[
(Y_i\perp\!\!\!\perp\mathbf B_i\mid
\mathbf W\cup\mathbf A_i)_
{\mathcal G_{\overline{\mathbf W},\overline{\mathbf A_i},
\overline{\mathbf B_i}}}.
\]
Rule~3 therefore gives, for every $\mathbf a,\mathbf b$,
\begin{equation}
\mathbb E[Y_i\mid do(\mathbf W=\mathbf w),do(\mathbf A_i=\mathbf a)]
=
\mathbb E[Y_i\mid do(\mathbf W=\mathbf w),
do(\mathbf A_i=\mathbf a),do(\mathbf B_i=\mathbf b)].
\label{eq:onlyif_insert}
\end{equation}

Next delete the actions on $\mathbf Z$ while retaining those
on $\mathbf X$.
Every directed path in $\mathcal Q$ from $\mathbf Z$ to $Y_i$
meets $\mathbf X$. To see this, append $\rho_i$ to such a path.
Its last surviving vertex before $Y_{\ell(i)}$ exists because
the starting vertex is in $\mathbf Z\subseteq\mathbf S$.
This last vertex is a parent of $Y_{\ell(i)}$ in $\mathcal H$,
hence belongs to $\mathbf X$.
It lies on the original path, since $\rho_i$ has no surviving
vertex before its endpoint.
Since $\mathbf Z\cap\mathbf X=\varnothing$ and
$Y_i\notin\mathbf X$, the path must enter $\mathbf X$
after its starting vertex. Deleting incoming edges to
$\mathbf X$ therefore blocks every such path.

Apply Rule~3 with $\mathcal K=\mathcal G$,
$\mathbf I=\mathbf X$, and $\mathbf J=\mathbf Z$.
Its required graph is
\[
\mathcal G_{\overline{\mathbf X},\overline{\mathbf Z}}
=\mathcal Q_{\overline{\mathbf X}}.
\]
Here $\mathbf X$ and $\mathbf Z$ are unconfounded roots.
Conditioning on $\mathbf X$ cannot activate colliders, and
any collider-free path from $\mathbf Z$ must be directed.
The preceding argument excludes such a path to $Y_i$, proving
\[
(Y_i\perp\!\!\!\perp\mathbf Z\mid\mathbf X)_
{\mathcal G_{\overline{\mathbf X},\overline{\mathbf Z}}}.
\]
Rule~3 deletes $do(\mathbf Z=\mathbf w_{\mathbf Z})$.
Since $\mathbf W\cup\mathbf X'=\mathbf X\cup\mathbf Z$,
this gives
\begin{equation}
\mathbb E[Y_i\mid do(\mathbf W=\mathbf w),do(\mathbf X'=\mathbf x')]
=\mu_i\bigl(\mathbf X,(\mathbf w_{\mathbf W_0},\mathbf x')\bigr).
\label{eq:onlyif_delete}
\end{equation}

\medskip
\noindent\emph{Step 5: One mixture for all rewards.}
For the fixed SCM and $\mathbf w$, define
\[
p_{\mathbf w}(\mathbf x')
=P(\mathbf X'=\mathbf x'\mid do(\mathbf W=\mathbf w)).
\]
For each $i$, identify $\mathbf x'$ with
$(\mathbf a,\mathbf b)$ according to
$\mathbf X'=\mathbf A_i\cup\mathbf B_i$, and write
\[
q_{\mathbf w,i}(\mathbf a)
=\sum_{\mathbf b\in\mathcal D(\mathbf B_i)}
p_{\mathbf w}(\mathbf a,\mathbf b)
=P(\mathbf A_i=\mathbf a\mid do(\mathbf W=\mathbf w)).
\]
The law of total expectation, followed by
\eqref{eq:onlyif_exchange}, \eqref{eq:onlyif_insert}, and
\eqref{eq:onlyif_delete}, gives
\begin{align}
\mu_i(\mathbf W,\mathbf w)
&=\sum_{\mathbf a}q_{\mathbf w,i}(\mathbf a)
\mathbb E[Y_i\mid do(\mathbf W=\mathbf w),\mathbf A_i=\mathbf a]
\nonumber\\
&=\sum_{\mathbf a}q_{\mathbf w,i}(\mathbf a)
\mathbb E[Y_i\mid do(\mathbf W=\mathbf w),do(\mathbf A_i=\mathbf a)]
\nonumber\\
&=\sum_{\mathbf a,\mathbf b}p_{\mathbf w}(\mathbf a,\mathbf b)
\mathbb E[Y_i\mid do(\mathbf W=\mathbf w),
do(\mathbf X'=(\mathbf a,\mathbf b))]
\nonumber\\
&=\sum_{\mathbf x'\in\mathcal D(\mathbf X')}
p_{\mathbf w}(\mathbf x')
\mu_i\bigl(\mathbf X,(\mathbf w_{\mathbf W_0},\mathbf x')\bigr).
\label{eq:onlyif_fixed}
\end{align}
The sums over $\mathbf a$ and $\mathbf b$ range over their
respective assignment domains. Zero-weight conditional terms
are omitted. The third equality is valid because
\eqref{eq:onlyif_insert} makes the intervened mean independent
of $\mathbf b$: expanding $q_{\mathbf w,i}(\mathbf a)$ into
$\sum_{\mathbf b}p_{\mathbf w}(\mathbf a,\mathbf b)$ therefore
does not change the sum.
Crucially, $p_{\mathbf w}$ is the same original joint distribution
for every reward coordinate, even though $\mathbf A_i$ and
$\mathbf B_i$ depend on $i$.
For an empty variable set, its assignment domain contains one
empty assignment, so these formulas also cover empty sets.

Now let $\widetilde{\mathbf W}$ have any distribution
$\pi_{\mathbf W}$ on $\mathcal D(\mathbf W)$. For
$\mathbf x\in\mathcal D(\mathbf X)$, define
\[
\pi_{\mathbf X}(\mathbf x)
=\sum_{\mathbf w\in\mathcal D(\mathbf W)}
\pi_{\mathbf W}(\mathbf w)
\mathbf 1\{\mathbf w_{\mathbf W_0}=\mathbf x_{\mathbf W_0}\}
p_{\mathbf w}(\mathbf x_{\mathbf X'}).
\]
This distribution is nonnegative and normalized, since
\[
\sum_{\mathbf x}\pi_{\mathbf X}(\mathbf x)
=\sum_{\mathbf w}\pi_{\mathbf W}(\mathbf w)
\sum_{\mathbf x'}p_{\mathbf w}(\mathbf x')=1.
\]
It can be sampled by first drawing $\mathbf w\sim\pi_{\mathbf W}$,
then drawing $\mathbf x'\sim p_{\mathbf w}$, and assigning
$(\mathbf w_{\mathbf W_0},\mathbf x')$ to $\mathbf X$,
using randomness independent of the SCM's exogenous variables.
Here $\mathbf w$ is an auxiliary draw; the executed intervention
has target set $\mathbf X$. Thus any dependence between the
assigned coordinates on
$\mathbf W_0$ and $\mathbf X'$ is preserved.
The distribution may depend on the fixed SCM, as permitted
by the existence claim.

Let $\widetilde{\mathbf X}^{*}$ have distribution
$\pi_{\mathbf X}$. Since \eqref{eq:onlyif_fixed} holds for all
$i$ with the same weights, it gives the vector identity
\begin{align*}
\boldsymbol\mu(\mathbf X,\widetilde{\mathbf X}^{*})
&=\sum_{\mathbf w}\pi_{\mathbf W}(\mathbf w)
\sum_{\mathbf x'}p_{\mathbf w}(\mathbf x')
\boldsymbol\mu\bigl(\mathbf X,
(\mathbf w_{\mathbf W_0},\mathbf x')\bigr)\\
&=\sum_{\mathbf w}\pi_{\mathbf W}(\mathbf w)
\boldsymbol\mu(\mathbf W,\mathbf w)\\
&=\boldsymbol\mu(\mathbf W,\widetilde{\mathbf W}).
\end{align*}
The SCM and the distribution $\pi_{\mathbf W}$ were arbitrary,
and $\mathbf X\neq\mathbf W$ with
$\mathbf X\in\mathbb M_{\mathcal G,\mathbf Y}$.
Hence every randomized intervention on $\mathbf W$ is matched
by one on another minimal intervention set.
Equality in every coordinate satisfies the non-strict
inequalities in Definition~\ref{defppomis}, so $\mathbf W$ is not a PPOMIS.
This proves the only-if direction of Theorem~\ref{thm1}.

\subsection{Details of Running Algorithm~\ref{alg:pomis_muct} on the Causal Graphs in Figure~\ref{exp_graphs_m} and Its Equivalence to \cite{lee2018structural} in the Single Reward-Node Setting}

\label{runnexm}

We recall the causal graphs from the Figure \ref{exp_graphs_m}:

\begin{figure}[h]
\begin{subfigure}[b]{3.4cm}
\centering
        \begin{tikzpicture}
      \node(1) {$X_3$};
      \node[red](2) [right of = 1,xshift = 0.15cm] {$Y_2$};
      \node(3) [left of = 1,xshift = -0.15cm] {$X_2$};
      \node(4) [above of = 3] {$X_1$};
      \node[red] (5) [below of=3] {$Y_1$};
 
        \draw (4) -> (3); 
  \draw (3) -> (5); 
  \draw (3) -> (1); 
  \draw (1) -> (2); 

   \draw[dashed, <->, bend right=45] (4) to (5); 

    \end{tikzpicture}
    \subcaption{$\mathcal{G}_1$}
\end{subfigure}
\hfill
\begin{subfigure}[b]{3.4cm}
    \centering
    \begin{tikzpicture}
  \node(1) {$X_3$};
      \node[red](2) [right of = 1,xshift = 0.15cm] {$Y_2$};
      \node(3) [left of = 1,xshift = -0.15cm] {$X_2$};
      \node(4) [above of = 3] {$X_1$};
      \node[red] (5) [below of=3] {$Y_1$};
 
        \draw (4) -> (3); 
  \draw (3) -> (5); 
  \draw (1) -> (3); 
  \draw (1) -> (2); 
\draw[dashed, <->, bend left=45] (3) to (2); 


    \end{tikzpicture}
    \subcaption{$\mathcal{G}_2$}
\end{subfigure} 
\hfill
\begin{subfigure}[b]{3.4cm}
\centering
   \begin{tikzpicture}
      \node(1) {$X_3$};
      \node[red](2) [right of = 1,xshift = 0.15cm] {$Y_2$};
      \node(3) [left of = 1,xshift = -0.15cm] {$X_2$};
      \node(4) [above of = 3] {$X_1$};
        \node(6) [right of = 4,xshift = 0.15cm] {$X_4$};
        \node[red](7) [right of = 6,xshift = 0.15cm] {$Y_3$};
      \node[red] (5) [below of=3] {$Y_1$};
 
        \draw (4) -> (3); 
  \draw (3) -> (5); 
  \draw (3) -> (1); 
  \draw (1) -> (2); 
 \draw (1) -> (2);
  \draw (6) -> (7);
    \draw (3) -> (6);
   \draw[dashed, <->, bend right=70] (3) to (5); 

    \end{tikzpicture}
    \subcaption{$\mathcal{G}_3$}
\end{subfigure}
\hfill
\begin{subfigure}[b]{3.4cm}
\centering
    \begin{tikzpicture}
      \node(1) {$X_3$};
      \node[red](2) [right of = 1,xshift = 0.15cm] {$Y_2$};
      \node(3) [left of = 1,xshift = -0.15cm] {$X_2$};
      \node(4) [above of = 3] {$X_1$};
        \node(6) [right of = 4,xshift = 0.15cm] {$X_4$};
        \node[red](7) [right of = 6,xshift = 0.15cm] {$Y_3$};
      \node[red] (5) [below of=3] {$Y_1$};
 
        \draw (4) -> (3); 
  \draw (3) -> (5); 
  \draw (3) -> (1); 
  \draw (1) -> (2); 
 \draw (1) -> (2);
  \draw (6) -> (7);
\draw (4) -> (6);
   \draw[dashed, <->, bend right=40] (4) to (5); 

    \end{tikzpicture}
    \subcaption{$\mathcal{G}_4$}
\end{subfigure}
\caption{Examples of causal graphs with multiple reward nodes}
\vspace{-1em}
\label{exp_graphs}
\end{figure}
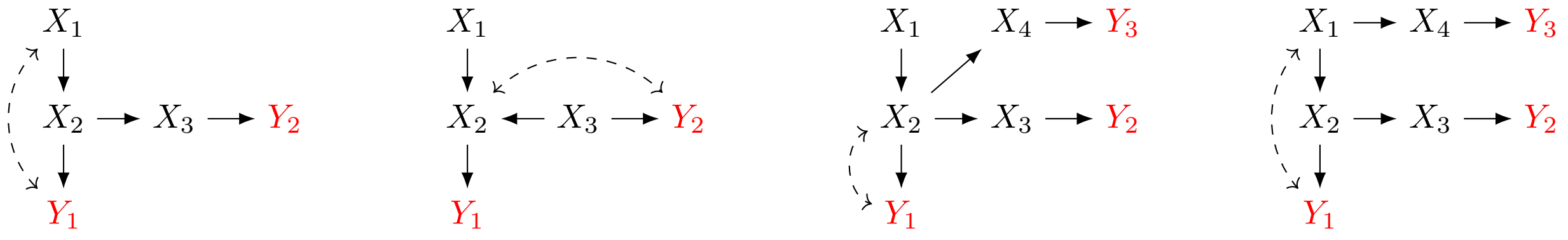

\begin{figure}[h]
\centering
\begin{subfigure}[t]{0.25\textwidth}
\centering
\begin{tikzpicture}[node distance=1cm]
  \node (3) {$X_2$};
  \node (1) [right of=3,xshift = 0.20cm] {$X_3$};
  \node[red] (2) [right of=1,xshift = 0.20cm] {$Y_2$};
  \node[red] (5) [below of=3,yshift = -0.80em] {$Y_1$};

  \draw (3) -> (5);
  \draw (3) -> (1);
  \draw (1) -> (2);
  \draw[dashed, <->, bend right=45] (3) to (5);
\end{tikzpicture}

{\small (a) Projection after 1st Iteration}
\end{subfigure}%
\hspace{1cm}
\begin{subfigure}[t]{0.25\textwidth}
\centering
\begin{tikzpicture}[node distance=1cm]
  \node (3) {\phantom{$X_2$}}; 
  \node (1) [right of=3] {$X_3$};
  \node[red] (2) [right of=1,xshift=0.20em] {$Y_2$};
  \node[red] (5) [below of=3,yshift = -1.0em] {$Y_1$};

  \draw (1) -> (2);
  \draw[dashed, <->, bend left=45] (1) to (5);
\end{tikzpicture}

{\small (b) Projection after 2nd Iteration}
\end{subfigure}

\caption{Projection graph(s) for $\mathcal{G}_1$ during execution of Algorithm \ref{alg:pomis_muct_full}.}
\label{exm33}
\vspace{-0.5em}
\end{figure}
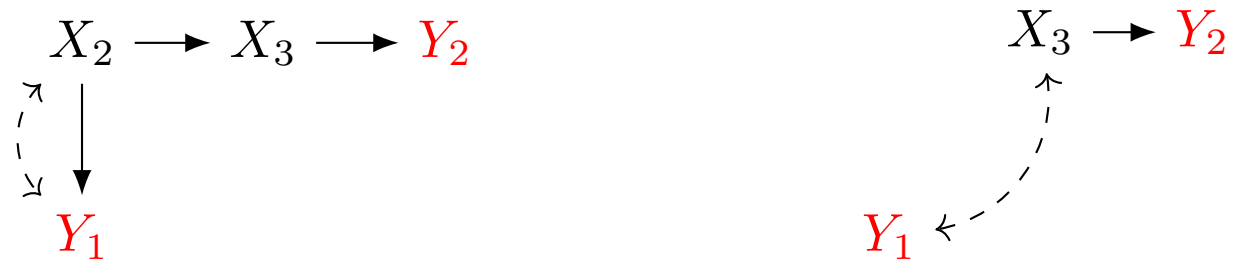

When we run Algorithm~\ref{alg:pomis_muct} on the graph $\mathcal{G}_1$, the while loop first identifies a bi-directed edge (or latent confounder) between the reward node $Y_1$ and $X_1$. Since $X_1$ is an ancestor of $Y_1$, the loop condition is satisfied. The set $\mathbf{T} = \{X_1\}$, and the projected graph is shown in Figure~\ref{exm33}(a). In this projected graph, there is a bi-directed edge (or latent confounder) between $Y_1$ and $X_2$, and $X_2$ is an ancestor of $Y_1$, so the while-loop condition holds again, with $\mathbf{T} = \{X_2\}$, and the subsequent projected graph is shown in Figure~\ref{exm33}(b). The base case is reached since none of the reward nodes have latent confounders with their corresponding ancestors, making the while-loop condition false. The resulting MUCT is $\{Y_1, Y_2, X_1, X_2\}$, and the corresponding PPOMIS is $\{X_3\}$. Repeating this procedure for all $\mathcal{G}_{1\overline{\mathbf{W}}}$ with $\mathbf{W} \subseteq \{X_1, X_2\}$, the full set of PPOMISs is $\{\{X_3\}, \{X_2, X_3\}\}$.

As for the graph $\mathcal{G}_2$, one can see that there is a latent confounder between $X_2$ and $Y_2$, but actually $X_2$ is not an ancestor of the reward node $Y_2$. Therefore, the base case is reached since none of the reward nodes have latent confounders with their corresponding ancestors. Applying Algorithm~\ref{alg:pomis_muct} results in a MUCT of $\{Y_1, Y_2\}$ and a PPOMIS of $\{X_2, X_3\}$. Since the MUCT is composed only of reward nodes, no further steps are needed, and the procedure terminates here.

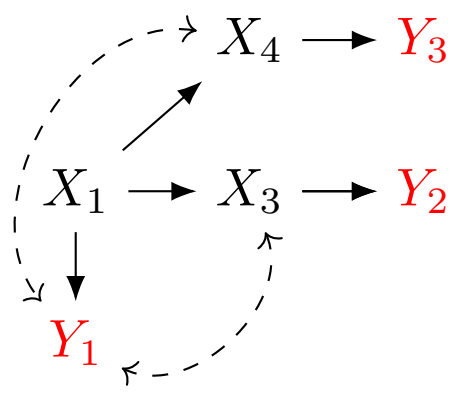
\begin{figure}[h]
\centering
\begin{subfigure}[t]{0.25\textwidth}
\centering
 \begin{tikzpicture}
      \node(1) {$X_3$};
      \node[red](2) [right of = 1,xshift = 0.15cm] {$Y_2$};
      \node(3) [left of = 1,xshift = -0.15cm] {$X_1$};
     
        \node(6) [right of = 4,xshift = 0.15cm] {$X_4$};
        \node[red](7) [right of = 6,xshift = 0.15cm] {$Y_3$};
      \node[red] (5) [below of=3] {$Y_1$};

  \draw (3) -> (5); 
  \draw (3) -> (1); 
  \draw (1) -> (2); 
 \draw (1) -> (2);
  \draw (6) -> (7);
    \draw (3) -> (6);
   \draw[dashed, <->, bend left=70] (1) to (5); 

    \draw[dashed, <->, bend right=70] (6) to (5); 

    \end{tikzpicture}

{\small (a) Projection after 1st Iteration}
\end{subfigure}%

\caption{Projection graph(s) for $\mathcal{G}_3$ during execution of Algorithm \ref{alg:pomis_muct_full}.}
\label{exm34}
\vspace{-0.5em}
\end{figure}

For the graph $\mathcal{G}_3$, there is a latent confounder between $X_2$ and $Y_1$. Hence, Algorithm~\ref{alg:pomis_muct} first identifies $\mathbf{T}=\{X_2\}$ and projects the graph onto $\mathbf{V}_{\mathcal{G}_3}\setminus\mathbf{T}$, as shown in Figure~\ref{exm34}(a). Although the projected graph contains induced latent confounders between $X_3$ and $Y_1$ and between $X_4$ and $Y_2$, neither $X_3$ nor $X_4$ is an ancestor of the reward node $Y_1$ in the projected graph. Therefore, the base case is reached, since no reward node is confounded with any of its ancestors. The resulting MUCT is $\{Y_1,Y_2,Y_3,X_2\}$, and the corresponding PPOMIS is $\{X_1,X_3,X_4\}$. Repeating this procedure for all $\mathcal{G}_{3,\overline{\mathbf{W}}}$ with $\mathbf{W}\subseteq \{X_2\}$ yields the full PPOMIS collection $\{\{X_1,X_3,X_4\},\{X_2,X_3,X_4\}\}$.

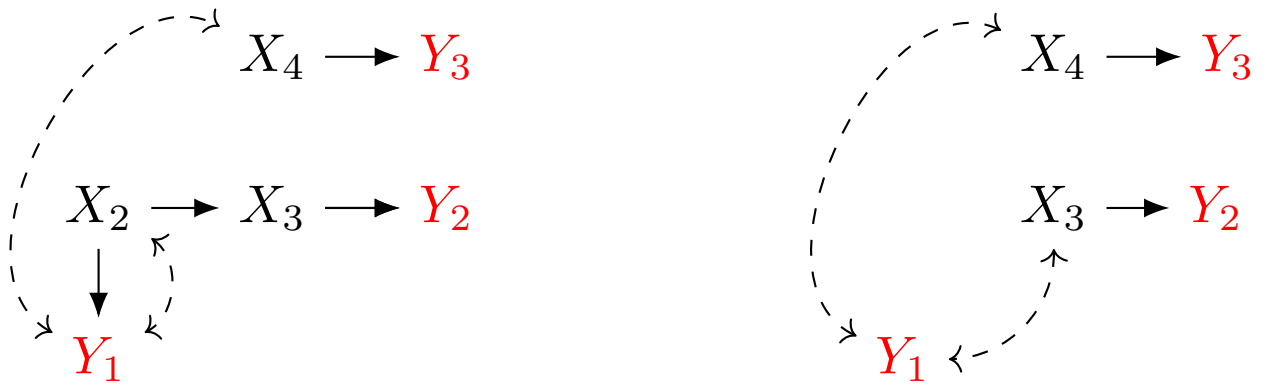
\begin{figure}[h]
\centering
\begin{subfigure}[t]{0.25\textwidth}
\centering
 \begin{tikzpicture}
      \node(1) {$X_3$};
      \node[red](2) [right of = 1,xshift = 0.15cm] {$Y_2$};
      \node(3) [left of = 1,xshift = -0.15cm] {$X_2$};
     
        \node(6) [right of = 4,xshift = 0.15cm] {$X_4$};
        \node[red](7) [right of = 6,xshift = 0.15cm] {$Y_3$};
      \node[red] (5) [below of=3] {$Y_1$};

  \draw (3) -> (5); 
  \draw (3) -> (1); 
  \draw (1) -> (2); 
 \draw (1) -> (2);
  \draw (6) -> (7);
   
   \draw[dashed, <->, bend left=60] (3) to (5); 

    \draw[dashed, <->, bend left=90] (5) to (6); 

    \end{tikzpicture}

{\small (a) Projection after 1st Iteration}
\end{subfigure}%
\hspace{1cm}
\begin{subfigure}[t]{0.25\textwidth}
\centering
\begin{tikzpicture}[node distance=1cm]
  \node (3) {\phantom{$X_2$}}; 
  \node (1) [right of=3,yshift = -1em] {$X_3$};
  \node[red] (2) [right of=1,xshift=0.20em] {$Y_2$};
  \node[red] (5) [below of=3,yshift = -1.0em] {$Y_1$};
  \node(6) [above of = 1] {$X_4$};
        \node[red](7) [right of = 6,xshift = 0.15cm] {$Y_3$};
  \draw (1) -> (2);
   \draw (6) -> (7);
  \draw[dashed, <->, bend left=45] (1) to (5);
   \draw[dashed, <->, bend left=90] (5) to (6); 
\end{tikzpicture}

{\small (b) Projection after 2nd Iteration}
\end{subfigure}

\caption{Projection graph(s) for $\mathcal{G}_4$ during execution of Algorithm \ref{alg:pomis_muct_full}.}
\label{exm35}
\vspace{-0.5em}
\end{figure}

For the graph $\mathcal{G}_4$, there is a latent confounder between $X_1$ and $Y_1$. Hence, Algorithm~\ref{alg:pomis_muct} first identifies $\mathbf{T}=\{X_1\}$ and projects the graph onto $\mathbf{V}_{\mathcal{G}_4}\setminus\mathbf{T}$, as shown in Figure~\ref{exm35}(a).  In this projected graph, there is a bi-directed edge (or latent confounder) between $Y_1$ and $X_2$, and $X_2$ is an ancestor of $Y_1$, so the while-loop condition holds again, with $\mathbf{T} = \{X_2\}$, and the subsequent projected graph is shown in Figure~\ref{exm35}(b). The base case is reached since none of the reward nodes have latent confounders with their corresponding ancestors, making the while-loop condition false. The resulting MUCT is $\{Y_1, Y_2, Y_3,X_1, X_2\}$, and the corresponding PPOMIS is $\{X_3,X_4\}$. Repeating this procedure for all $\mathcal{G}_{4\overline{\mathbf{W}}}$ with $\mathbf{W} \subseteq \{X_1, X_2\}$, the full set of PPOMISs is $\{\{X_3,X_4\}, \{X_2, X_3,X_4\}\}$.

The work of~\citep{lee2018structural} studies causal bandits with a single reward node under a known causal graph. Their search-space reduction is based on two objects: the \emph{Minimal Unobserved Confounder Territory (MUCT)} and the \emph{Interventional Border (IB)}. For a single reward node $Y$, the set $\mathrm{MUCT}(\mathcal{G},Y)$ is constructed by starting from $\{Y\}$ and iteratively expanding the current set by adding its c-component and its descendants until convergence. The interventional border is then defined as $\mathrm{IB}(\mathcal{G},Y)=\mathsf{pa}(\mathrm{MUCT}(\mathcal{G},Y))\setminus \mathrm{MUCT}(\mathcal{G},Y)$. Our Algorithm~\ref{alg:pomis_muct} generalizes this construction to multiple reward nodes. In the special case of a single reward node, it reduces to the procedure of~\citep{lee2018structural}: it returns the same $\mathrm{MUCT}(\mathcal{G},Y)$ and its output $\mathrm{PPOMIS}(\mathcal{G},Y)$ coincides with $\mathrm{IB}(\mathcal{G},Y)$.

To see this, note that Algorithm~\ref{alg:pomis_muct} checks whether $Y$ is latently confounded with any of its ancestors. Whenever this occurs, the ancestors of $Y$ that are bidirected-connected to $Y$ are added to the current territory. Let $\mathbf{T}$ denote the set of such nodes identified in one iteration. The algorithm then projects the graph onto $\mathbf{V}_{\mathcal{G}}\setminus\mathbf{T}$ using the projection procedure in Section~\ref{proj}. This projection transfers the relevant confounding structure to the remaining variables: children of nodes in $\mathbf{T}$ inherit bidirected connections with $Y$, and variables confounded with nodes in $\mathbf{T}$ inherit bidirected connections with the corresponding children. The graph is then updated to the projected graph, and the process repeats until $Y$ is no longer confounded with any of its ancestors. At that point, the algorithm returns the accumulated territory as $\mathrm{MUCT}(\mathcal{G},Y)$ and the parents of $Y$ in the final projected graph as $\mathrm{PPOMIS}(\mathcal{G},Y)$. This is exactly the single-reward MUCT/IB construction of~\citep{lee2018structural}.

\begin{figure}[h]
\centering

\begin{subfigure}[t]{0.23\textwidth}
\centering
\begin{tikzpicture}[scale=0.9]
  \node (3) {$X_2$};
  \node (4) [above of=3] {$X_1$};
  \node (6) [right of=4] {$X_4$};
  \node (7) [right of=3] {$X_3$};
  \node[red] (5) [below of=3] {$Y$};

  \draw[->] (3) -- (5);
  \draw[->] (6) -- (7);
  \draw[->] (4) -- (3);
  \draw[->] (7) -- (5);
  \draw[dashed, <->, bend right=45] (4) to (5);
  \draw[dashed, <->, bend left=60] (3) to (7);
\end{tikzpicture}
\caption{Graph $\mathcal{G}$}
\end{subfigure}%
\hfill
\begin{subfigure}[t]{0.23\textwidth}
\centering
\begin{tikzpicture}[scale=0.9]
  \node (3) {$X_2$};
  \node (6) [above right of=3,yshift = 1em,xshift = 0.9em] {$X_4$};
  \node (7) [right of=3] {$X_3$};
  \node[red] (5) [below of=3] {$Y$};

  \draw[->] (3) -- (5);
  \draw[->] (6) -- (7);
  \draw[->] (7) -- (5);
  \draw[dashed, <->, bend right=75] (3) to (5);
  \draw[dashed, <->, bend left=60] (3) to (7);
\end{tikzpicture}
\caption{After 1st iteration}
\end{subfigure}%
\hfill
\begin{subfigure}[t]{0.23\textwidth}
\centering
\begin{tikzpicture}[scale=0.9]
  \node (6) {$X_4$};
  \node (7) [right of=6] {$X_3$};
  \node[red] (5) [below of=7] {$Y$};

  \draw[->] (6) -- (7);
  \draw[->] (7) -- (5);
  \draw[dashed, <->, bend right=55] (7) to (5);
\end{tikzpicture}
\caption{After 2nd iteration}
\end{subfigure}%
\hfill
\begin{subfigure}[t]{0.23\textwidth}
\centering
\begin{tikzpicture}[scale=0.9]
  \node (7) {$X_4$};
  \node[red] (5) [below of=7] {$Y$};

  \draw[->] (7) -- (5);
\end{tikzpicture}
\caption{After 3rd iteration}
\end{subfigure}

\caption{Application of Algorithm~\ref{alg:pomis_muct_full} to a causal graph $\mathcal{G}$ with a single reward node $Y$.}
\label{exm36}
\end{figure}
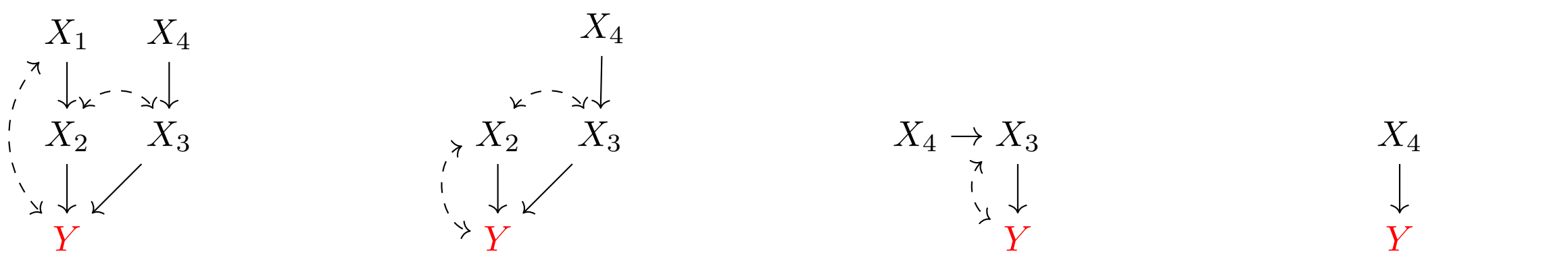

Using the procedure of~\citep{lee2018structural} for the graph $\mathcal{G}$ in Figure~\ref{exm36}(a), we obtain $\mathrm{MUCT}(\mathcal{G},Y)=\{Y,X_1,X_2,X_3\}$ and $\mathrm{IB}(\mathcal{G},Y)=\{X_4\}$. Figures~\ref{exm36}(b)--(d) show the successive projected graphs produced by Algorithm~\ref{alg:pomis_muct_full}. The final projected graph has parent set $\{X_4\}$ for $Y$, so $\mathrm{PPOMIS}(\mathcal{G},Y)=\{X_4\}=\mathrm{IB}(\mathcal{G},Y)$. The accumulated territory is $\{Y,X_1,X_2,X_3\}$, matching $\mathrm{MUCT}(\mathcal{G},Y)$. Hence, for causal graphs with a single reward node, Algorithm~\ref{alg:pomis_muct_full} is equivalent to the construction of~\citep{lee2018structural}.

\subsection{Proof of Lemma \ref{lem2}}
\vspace{0.5em}
\lemmm*
\vspace{0.5em}
 \textbf{Proof:} Consider two sets $\mathbf{W} \subseteq \text{MUCT}(\mathcal{G},\mathbf{Y}) \setminus \mathbf{Y}$ and $\mathbf{W}' \subseteq \left(\text{MUCT}(\mathcal{G},\mathbf{Y}) \setminus \mathbf{Y}\right) \setminus \mathbf{W}$, and let $\mathbf{T} = \text{MUCT}(\mathcal{G}_{\overline{\mathbf{W}}},\mathbf{Y})$. We can partition $\mathbf{W}'$ into two disjoint sets:  
\(
\mathbf{W}' = (\mathbf{W}' \cap \mathbf{T}) \cup (\mathbf{W}' \setminus \mathbf{T}),
\)  
where $\mathbf{T} = \text{MUCT}(\mathcal{G}_{\overline{\mathbf{W}}},\mathbf{Y})$. For any node $X \in \mathbf{W}' \setminus \mathbf{T}$, by the construction of $\mathbf{T} = \text{MUCT}(\mathcal{G}_{\overline{\mathbf{W}}},\mathbf{Y})$ (via Algorithm \ref{alg:pomis_muct} or Algorithm \ref{alg:pomis_muct_full} ), the node $X$ does not have any bi-directed edge or latent confounder with any reward node $\mathbf{Y}_i \in \mathbf{Y}$ in any iteration. During all iterations of the while-loop in Algorithm \ref{alg:pomis_muct} or Algorithm \ref{alg:pomis_muct_full}, such a node $X$ would have been added to $\mathbf{T}$ if it ever became confounded with any $\mathbf{Y}_i$; however, since $X \notin \mathbf{T}$, this never occurs. This yields a contradiction if we assume otherwise. Moreover, since $\mathcal{G}_{\overline{\mathbf{W} \cup (\mathbf{W}' \cap \mathbf{T})}}$ is a sparser graph than $\mathcal{G}_{\overline{\mathbf{W}}}$, the node $\mathbf{X}$ also cannot acquire a bi-directed edge or latent confounder with any $\mathbf{Y}_i$ in any subsequent iteration of the while-loop when computing $\mathcal{G}_{\overline{\mathbf{W} \cup (\mathbf{W}' \cap \mathbf{T})}}$. Therefore, removing the incoming edges of nodes in $\mathbf{W}' \setminus \mathbf{T}$, in addition to removing those for nodes in $\mathbf{W} \cup (\mathbf{W}' \cap \mathbf{T})$, does not affect the resulting $\text{MUCT}$ or the corresponding $\text{PPOMIS}$. Hence,  
\[
\text{MUCT}(\mathcal{G}_{\overline{\mathbf{W}\cup \mathbf{W}'}},\mathbf{Y})
= \text{MUCT}(\mathcal{G}_{\overline{\mathbf{W} \cup (\mathbf{W}' \cap \mathbf{T}) \cup (\mathbf{W}' \setminus \mathbf{T} )}},\mathbf{Y})
= \text{MUCT}(\mathcal{G}_{\overline{\mathbf{W} \cup (\mathbf{W}' \cap \mathbf{T})}},\mathbf{Y}),
\]
and similarly,  
\[
\text{PPOMIS}(\mathcal{G}_{\overline{\mathbf{W}\cup \mathbf{W}'}},\mathbf{Y})
= \text{PPOMIS}(\mathcal{G}_{\overline{\mathbf{W} \cup (\mathbf{W}' \cap \mathbf{T}) \cup (\mathbf{W}' \setminus \mathbf{T} )}},\mathbf{Y})
= \text{PPOMIS}(\mathcal{G}_{\overline{\mathbf{W} \cup (\mathbf{W}' \cap \mathbf{T})}},\mathbf{Y}).
\] 
This completes the proof of Lemma \ref{lem2}.

\subsection{Proof of Lemma \ref{lem3}}
\vspace{0.5em}
\lemmmmm*
\vspace{0.5em}
\textbf{Proof:} Consider any possible subset $\mathbf{W} \subseteq \text{MUCT}(\mathcal{G},\mathbf{Y}) \setminus \mathbf{Y}$. Note that the graphs $\mathcal{G}_{\overline{\mathbf{W} \cup \mathbf{W}'}}$ and $\mathcal{G}_{\overline{\mathbf{I} \cup \mathbf{W}'}}$ share the same edges in the corresponding induced subgraph on the nodes in the set $\mathbf{T} \cup \mathbf{I}$, where $\mathbf{T} = \text{MUCT}(\mathcal{G}_{\overline{\mathbf{W}}},\mathbf{Y})$, $\mathbf{I} = \text{PPOMIS}(\mathcal{G}_{\overline{\mathbf{W}}},\mathbf{Y})$, and $\mathbf{W}'$ is any subset of $\mathbf{T} \setminus \mathbf{Y}$. The only difference is that the set $\mathbf{I}$ has no parents in the graph $\mathcal{G}_{\overline{\mathbf{I} \cup \mathbf{W}'}}$, which is irrelevant to identifying the $\text{MUCT}$ and $\text{PPOMIS}$ in both $\mathcal{G}_{\overline{\mathbf{W} \cup \mathbf{W}'}}$ and $\mathcal{G}_{\overline{\mathbf{I} \cup \mathbf{W}'}}$. Therefore, both causal graphs share the same $\text{MUCT}$ and $\text{PPOMIS}$, i.e., we have $\text{MUCT}(\mathcal{G}_{\overline{\mathbf{W} \cup \mathbf{W}'}},\mathbf{Y}) = \text{MUCT}(\mathcal{G}_{\overline{\mathbf{I} \cup \mathbf{W}'}}[\mathbf{T} \cup \mathbf{I}],\mathbf{Y})$ and $\text{PPOMIS}(\mathcal{G}_{\overline{\mathbf{W} \cup \mathbf{W}'}},\mathbf{Y}) = \text{PPOMIS}(\mathcal{G}_{\overline{\mathbf{I} \cup \mathbf{W}'}}[\mathbf{T} \cup \mathbf{I}],\mathbf{Y})$. This completes the proof of Lemma \ref{lem3}.

\subsection{Proof of Theorem \ref{thm2}}
\vspace{0.5em}
\thmmm*
\vspace{0.5em}
\textbf{Proof:} We define the notation $\boldsymbol{\mathcal{P}}_{\mathcal{G},\mathbf{Y}} := \{\text{PPOMIS}(\mathcal{G}_{\overline{\mathbf{W}}},\mathbf{Y}) \mid \mathbf{W} \subseteq \mathbf{V} \setminus \mathbf{Y}\}$ as the collection of all the PPOMISs for the causal graph $\mathcal{G}$ with the set of reward nodes $\mathbf{Y}$. Also, let $\pi : \mathbf{W} \to \{1,\dots,n\}$ denote some arbitrary ordering of the elements of the set $\mathbf{W}$, where $n = |\mathbf{W}|$. We have $\{Z\} \prec \mathbf{B}$ if $\pi(Z) < \min_{b \in \mathbf{B}} \pi(b)$, where $Z \in \mathbf{W}$ and $\mathbf{B} \subseteq \mathbf{W}$. We also use the notation $\mathbf{T}_{\mathbf{W}} := \text{MUCT}(\mathcal{G}_{\overline{\mathbf{W}}},\mathbf{Y})$ and $\mathbf{I}_{\mathbf{W}} := \text{PPOMIS}(\mathcal{G}_{\overline{\mathbf{W}}},\mathbf{Y})$.

\begin{align*}
    \mathbf{\mathcal{P}}_{\mathcal{G},\mathbf{Y}} &= \{\text{PPOMIS}(\mathcal{G}_{\overline{\mathbf{W}}},\mathbf{Y})\}_{\mathbf{W} \subseteq \mathbf{V} \setminus \mathbf{Y}}\\
    &= \{\text{PPOMIS}(\mathcal{G}_{\overline{\mathbf{W}}},\mathbf{Y})\}_{\mathbf{W} \subseteq \mathbf{T} \setminus \mathbf{Y}}\\
    &= \{\text{PPOMIS}(\mathcal{G},\mathbf{Y})\}  \; \cup \; \{\text{PPOMIS}(\mathcal{G}_{\overline{\mathbf{W}}},\mathbf{Y})\}_{\mathbf{W} \subseteq \mathbf{T} \setminus \mathbf{Y},\; \mathbf{W} \neq \emptyset}\\
  &= \{\text{PPOMIS}(\mathcal{G},\mathbf{Y})\}  \; \cup \;
     \bigcup_{Z \in \mathbf{T} \setminus \mathbf{Y}}
     \{\text{PPOMIS}(\mathcal{G}_{\overline{Z,\mathbf{W}}},\mathbf{Y})\}_{\mathbf{W} \subseteq \mathbf{T} \setminus (\mathbf{Y}\cup\{Z\}), \; ( \; \{Z\} \prec \mathbf{W} \ \text{or}\ \mathbf{W}=\emptyset \;)}\\
       &= \{\text{PPOMIS}(\mathcal{G},\mathbf{Y})\}  \; \cup \;
     \bigcup_{Z \in \mathbf{T} \setminus \mathbf{Y}}
     \{\text{PPOMIS}(\mathcal{G}_{\overline{Z,\mathbf{W}}},\mathbf{Y})\}_{\mathbf{W} \subseteq \mathbf{T}_Z \setminus \mathbf{Y}, \; ( \; \{Z\} \prec \mathbf{W} \ \text{or}\ \mathbf{W}=\emptyset \;)}
\end{align*}

The first equality follows from the definition of $\mathbf{\mathcal{P}}_{\mathcal{G},\mathbf{Y}}$ as the collection of all PPOMISs. The second equality follows from Lemma \ref{lem1}. The third equality is straightforward.  The fourth equality follows from the fact that the power set (i.e., the set of all possible subsets) of $\mathbf{T}\setminus\mathbf{Y}$ can be partitioned according to the unique minimal element of each nonempty subset under the ordering $\pi$. Specifically, each nonempty subset belongs to exactly one partition indexed by a variable $Z \in \mathbf{T} \setminus \mathbf{Y}$, where $Z$ is the element with the smallest $\pi$-value in the subset, i.e., no other element $Z'$ in the subset satisfies $\pi(Z') \le \pi(Z)$. The last equality is a direct consequence of Lemma \ref{lem2}, i.e., for any node $Z \in \text{MUCT}(\mathcal{G}, \mathbf{Y})$, all variables outside the set $\mathbf{T}_Z = \text{MUCT}(\mathcal{G}_{\overline{Z}}, \mathbf{Y})$ can be safely excluded from any set $\mathbf{W}$ while computing $\text{PPOMIS}(\mathcal{G}_{\overline{Z,\mathbf{W}}},\mathbf{Y})$ without affecting the outcome. We now proceed as follows:

\begin{align}
    \mathbf{\mathcal{P}}_{\mathcal{G},\mathbf{Y}} 
       &= \{\text{PPOMIS}(\mathcal{G},\mathbf{Y})\}  \; \cup \bigcup_{Z \in \mathbf{T} \setminus \mathbf{Y}} \{\text{PPOMIS}(\mathcal{G}_{\overline{Z,\mathbf{W}}},\mathbf{Y})\}_{\mathbf{W} \subseteq \mathbf{T}_Z \setminus \mathbf{Y}, \; ( \; \{Z\} \prec \mathbf{W} \ \text{or}\ \mathbf{W}=\emptyset \;)} \nonumber \\
        &=\{\text{PPOMIS}(\mathcal{G},\mathbf{Y})\} \; \cup \; \bigcup_{Z \in \mathbf{T} \setminus \mathbf{Y}} \{\text{PPOMIS}(\mathcal{G}_{\overline{\mathbf{I}_Z,\mathbf{W}}}[\mathbf{T}_Z \; \cup \;\mathbf{I}_Z],\mathbf{Y})\}_{\mathbf{W} \subseteq \mathbf{T}_Z \setminus \mathbf{Y}, \; ( \; \{Z\} \prec \mathbf{W} \ \text{or}\ \mathbf{W}=\emptyset \;)}  
        \label{eq2}
\end{align}

Consider any $Z \in \text{MUCT}(\mathcal{G},\mathbf{Y}) \setminus \mathbf{Y}$, and  $\mathbf{T}_Z = \text{MUCT}(\mathcal{G}_{\overline{Z}},\mathbf{Y})$ and $\mathbf{I}_Z = \text{PPOMIS}(\mathcal{G}_{\overline{Z}},\mathbf{Y})$. For any $\mathbf{W} \subseteq \mathbf{T}_Z \setminus \mathbf{Y}$, we have $\text{MUCT}(\mathcal{G}_{\overline{Z \cup \mathbf{W}}},\mathbf{Y}) = \text{MUCT}(\mathcal{G}_{\overline{\mathbf{I}_Z \cup \mathbf{W}}}[\mathbf{T} \cup \mathbf{I}],\mathbf{Y})$ and $\text{PPOMIS}(\mathcal{G}_{\overline{Z \cup \mathbf{W}}},\mathbf{Y}) = \text{PPOMIS}(\mathcal{G}_{\overline{\mathbf{I}_Z \cup \mathbf{W}}}[\mathbf{T}_Z \cup \mathbf{I}_Z],\mathbf{Y})$. This follows directly from Lemma \ref{lem3} and serves as the justification for Equation \ref{eq2}. Also note that:

$$
    \mathbf{\mathcal{P}}_{\mathcal{G}_{\overline{Z}},\mathbf{Y}} = \{\text{PPOMIS}(\mathcal{G}_{\overline{Z,\mathbf{W}}},\mathbf{Y})\}_{\mathbf{W} \subseteq \mathbf{T}_Z \setminus \mathbf{Y}, \; ( \; \{Z\} \prec \mathbf{W} \ \text{or}\ \mathbf{W}=\emptyset \; )}
$$

With the help of Lemma \ref{lem3}, we have 
\(
\text{PPOMIS}(\mathcal{G}_{\overline{Z \cup \mathbf{W}}},\mathbf{Y}) = \text{PPOMIS}(\mathcal{G}_{\overline{\mathbf{I}_Z \cup \mathbf{W}}}[\mathbf{T}_Z \cup \mathbf{I}_Z],\mathbf{Y}).
\)
Thus, we obtain the following:

$$
    \mathbf{\mathcal{P}}_{\mathcal{G}_{\overline{Z}},\mathbf{Y}} =\{\text{PPOMIS}(\mathcal{G}_{\overline{\mathbf{I}_Z,\mathbf{W}}}[\mathbf{T}_Z \cup \mathbf{I}_Z],\mathbf{Y})\}_{\mathbf{W} \subseteq \mathbf{T}_Z \setminus \mathbf{Y}, \; ( \; \{Z\} \prec \mathbf{W} \ \text{or}\ \mathbf{W}=\emptyset \;)}       
$$

\begin{align}
    \mathbf{\mathcal{P}}_{\mathcal{G},\mathbf{Y}} 
      &=\{\text{PPOMIS}(\mathcal{G},\mathbf{Y})\}  \; \cup \; \bigcup_{Z\in \mathbf{T} \setminus \mathbf{Y}} \{\text{PPOMIS}(\mathcal{G}_{\overline{\mathbf{I}_Z,\mathbf{W}}}[\mathbf{T}_Z \cup \mathbf{I}_Z],\mathbf{Y})\}_{\mathbf{W} \subseteq \mathbf{T}_Z \setminus \mathbf{Y}, \; (\; \{Z\} \prec \mathbf{W} \ \text{or}\ \mathbf{W}=\emptyset \;)} \nonumber \\
       &=\{\text{PPOMIS}(\mathcal{G},\mathbf{Y})\}  \; \cup \; \bigcup_{Z\in \mathbf{T} \setminus \mathbf{Y}}  \mathbf{\mathcal{P}}_{\mathcal{G}_{\overline{Z}},\mathbf{Y}}
       \label{eq3}
\end{align}

The Algorithm \ref{alg:final_ppomis} implements this recursion exactly. It first computes $\mathbf{T} = \text{MUCT}(\mathcal{G},\mathbf{Y})$ and $\mathbf{I} = \text{PPOMIS}(\mathcal{G},\mathbf{Y})$ using Algorithm \ref{alg:pomis_muct} or Algorithm \ref{alg:pomis_muct_full} , and adds the set $\mathbf{I}$ to the collection of PPOMISs. After this, the function $\text{subPOMISs}(\mathcal{G},\mathbf{W},Y)$ is called to recursively compute all remaining PPOMISs. Specifically, starting with $\mathbf{T} = \text{MUCT}(\mathcal{G},\mathbf{Y})$, for every element $Z \in \mathbf{T} \setminus \mathbf{Y}$, it computes $\mathbf{T}_Z = \text{MUCT}(\mathcal{G}_{\overline{Z}},\mathbf{Y})$ and $\mathbf{I}_Z = \text{PPOMIS}(\mathcal{G}_{\overline{Z}},\mathbf{Y})$ using Algorithm \ref{alg:pomis_muct} or Algorithm \ref{alg:pomis_muct_full} . The next lines, $\mathbf{Z}_{\text{index}} = \mathbf{T}.\text{index}(Z)$ and $\mathbf{W} = \mathbf{T}_Z \cap \mathbf{T}\big[\;\mathbf{Z}_{\text{index}}+1:|\mathbf{T}|\;\big]$, narrow down the search over sets $\mathbf{W}$ such that $\mathbf{W} \subseteq \mathbf{T}_Z \setminus \mathbf{Y}$ and $\{Z\} \prec \mathbf{W}$. The set $\mathbf{I}_Z$ is added to the collection of PPOMISs, and the recursive call $\text{subPOMISs}(\mathcal{G}_{\overline{\mathbf{I}_Z,\mathbf{W}}}[\mathbf{T}_Z \cup \mathbf{I}_Z],\mathbf{W},Y)$ is made. This function call repeats the same process and is equivalent to computing $\mathbf{\mathcal{P}}_{\mathcal{G}_{\overline{Z}},\mathbf{Y}}$. The same process continues for every node $Z \in \mathbf{T} \setminus \mathbf{Y}$. Hence, the algorithm efficiently enumerates all PPOMISs by implementing the recursive formulation in Eq. \ref{eq3}. This completes the proof of Theorem \ref{thm2}.

\subsection{Proof of Theorem~\ref{thm3} and Corollary~\ref{crr1} (Justification of the Stopping Criterion for Algorithm~\ref{alg:pareto_ucb_cc})}

\label{pff_ap}
\vspace{0.5em}
\thmmmm*
\vspace{0.5em}
\textbf{Proof:} Let $X_1, \ldots, X_n$ be independent random variables such that $a_i \le X_i \le b_i$ almost surely. Consider the sum
\[
S_n = X_1 + \cdots + X_n.
\]
Hoeffding's inequality states that, for all $\epsilon > 0$,
\[
\mathbb{P}\big(|S_n - \mathbb{E}[S_n]| \ge \epsilon\big) \le 2 \exp\left( \frac{-2 \epsilon^2}{\sum_{i=1}^n (b_i - a_i)^2} \right),
\]
where $\mathbb{E}[S_n]$ is the expected value of $S_n$.

Suppose $\widehat{\mu}_a^i(t)$ is the estimated reward for objective $i \in [d]$ for action $a \in \mathcal{A}$, and let $N_a(t)$ be the number of times action $a$ has been played up to time step $t$. By Hoeffding's inequality, we have
\[
\mathbb{P}\Big(\big|\widehat{\mu}_a^i(t) - \mu_a^i\big| \ge \epsilon\Big) \le 2 \exp\big(-2 \epsilon^2 N_a(t)\big).
\]

Setting $2 \exp(-2 \epsilon^2 N_a(t)) = \delta'$, we solve for $\epsilon$ to get
\[
\epsilon = \sqrt{\frac{1}{2 N_a(t)} \log\left( \frac{2}{\delta'} \right)}.
\]

Thus,
\[
\mathbb{P}\Big(\big|\widehat{\mu}_a^i(t) - \mu_a^i\big| \ge \sqrt{\frac{1}{2 N_a(t)} \log\left( \frac{2}{\delta'} \right)} \Big) \le \delta',
\]
which is equivalent to
\[
\mathbb{P}\Big(\big|\widehat{\mu}_a^i(t) - \mu_a^i\big| \le \sqrt{\frac{1}{2 N_a(t)} \log\left( \frac{2}{\delta'} \right)} \Big) \ge 1 - \delta'.
\]

Let us define the event
\[
\mathcal{E}_a^i(t) := \Big\{ \big|\widehat{\mu}_a^i(t) - \mu_a^i\big| \le \sqrt{\frac{1}{2 N_a(t)} \log\left( \frac{2}{\delta'} \right)} \Big\}.
\]

Next, define the intersection of all such events:
\[
\mathcal{E} := \bigcap_{\substack{i \in [d] \\ a \in \mathcal{A} \\ t \in [T]}} \mathcal{E}_a^i(t).
\]

Since $\mathbb{P}(\mathcal{E}_a^i(t)) \ge 1 - \delta'$ for each $i, a, t$, using the union bound we have
\[
\mathbb{P}(\mathcal{E}) = 1 - \mathbb{P}\Big(\bigcup_{i\in[d],a\in \mathcal{A},t \in [T]} \overline{\mathcal{E}_a^i(t)}\Big) 
\ge 1 - \sum_{i\in[d],a\in \mathcal{A},t \in [T]} \mathbb{P}\big(\overline{\mathcal{E}_a^i(t)}\big) 
\ge 1 - k d T  \delta',
\]
where $k = |\mathcal{A}|$. If we choose $\delta' := \frac{\delta}{k d T}$, then by the union bound we have $\mathbb{P}(\mathcal{E}) \ge 1 - \delta$. The event $\mathcal{E}$ ensures that, for all time steps, the true expected reward for every objective and every action lies within the corresponding confidence interval; that is,

 \[
\mathcal{E} := \left\{
\begin{aligned}
& \forall a \in \mathcal{A} \\
& \forall t \in [T] \quad \quad
\big|\widehat{\mu}_a^i(t) - \mu_a^i\big| \le \sqrt{\frac{1}{2 N_a(t)} \log\left(\frac{2 k d T}{\delta}\right)} \\
& \forall i \in [d]
\end{aligned}
\right\}.
\]

 The probability of this good event satisfies $\mathbb{P}(\mathcal{E}) \ge 1 - \delta$. Also we use the notations $\mu^{+i}_a(t) := \widehat{\mu}_a^i(t)  + \sqrt{\frac{1}{2 N_a(t)} \log\left(\frac{2 k d T}{\delta}\right)}$  and $\mu^{-i}_a(t) := \widehat{\mu}_a^i(t)  - \sqrt{\frac{1}{2 N_a(t)} \log\left(\frac{2 k d T}{\delta}\right)}$.

Under the good event $\mathcal{E}$, the estimated set of Pareto-optimal arms $\widehat{\mathcal{F}}_t$ is a superset of the true set of Pareto-optimal arms, i.e., $\mathcal{F}^* \subseteq \widehat{\mathcal{F}}_t$ for all $t \in [T]$. Thus, the true optimal arms are never eliminated at any time $t \in [T]$ under the event $\mathcal{E}$. To prove this, suppose for contradiction that there exists an arm $a^* \in \mathcal{F}^*$ such that $a^* \notin \widehat{\mathcal{F}}_{t+1}$ for some $t < T$. We use the notation $\mathcal{S}(\mathbf{X}) := \{\mathbf{w} \ge 0 \mid \sum_{x \in \mathbf{X}} \mathbf{w}_x = 1\}$ to define probability simplex over set $\mathbf{X}$. 
Since $a^*\in\mathcal{F}^*$, by the definition of true Pareto-optimal actions, no convex combination of the remaining actions strictly improves all reward coordinates of $a^*$. Thus, we have:

\[
 \nexists\mathbf{w} \in \mathcal{S}(\mathcal{A} \setminus \{a^*\}) \;\; \text{s.t.} 
\sum_{a' \in \mathcal{A} \setminus \{a^*\}} \mathbf{w}_{a'} \mu_{a'}^i > \mu_{a^*}^i \quad \forall i \in [d].
\]

Since we assume that $a^* \notin \widehat{\mathcal{F}}_{t+1}$, this implies that there exists a weight vector
$\mathbf{w} \in \mathcal{S}(\widehat{\mathcal{F}}_{t} \setminus \{a^*\})$ such that
\[
\sum_{a' \in \widehat{\mathcal{F}}_{t} \setminus \{a^*\}} \mathbf{w}_{a'} \mu_{a'}^{-i}(t)
> \mu_{a^*}^{+i}(t), \quad \forall i \in [d].
\]

Since $\widehat{\mathcal{F}}_{t} \subseteq \mathcal{A}$, the weight vector  $\mathbf{w} \in \mathcal{S}(\widehat{\mathcal{F}}_{t} \setminus \{a^*\})$ can also be interpreted as an element of
$\mathcal{S}(\mathcal{A} \setminus \{a^*\})$ by extending $\mathbf{w}$ with zeros outside
$\widehat{\mathcal{F}}_{t}$. Hence $a^* \notin \widehat{\mathcal{F}}_{t+1}$ implies that
\[
\exists \mathbf{w} \in \mathcal{S}(\mathcal{A} \setminus \{a^*\}) \;\; \text{s.t.} \;\;
\sum_{a' \in \mathcal{A} \setminus \{a^*\}} \mathbf{w}_{a'} \mu_{a'}^{-i}(t)
> \mu_{a^*}^{+i}(t), \quad \forall i \in [d].
\]

Under the event $\mathcal{E}$ we have $\mu_{a'}^i \geq \mu_{a'}^{-i}(t)$  and $\mu_{a^*}^{+i}(t) \geq \mu_{a^*}^{i}$ Thus we have:

\[ \exists\mathbf{w} \in \mathcal{S}(\mathcal{A} \setminus \{a^*\}) \;\; \ \text{s.t.} 
\sum_{a' \in \mathcal{A} \setminus \{a^*\}} \mathbf{w}_{a'} \mu_{a'}^{i} \geq \sum_{a' \in \mathcal{A} \setminus \{a^*\}} \mathbf{w}_{a'} \mu_{a'}^{-i}(t) > \mu_{a^*}^{+i}(t) \geq \mu_{a^*}^{i}  \quad \forall i \in [d].
\]

Which implies that:
 
\[ \exists\mathbf{w} \in \mathcal{S}(\mathcal{A} \setminus \{a^*\}) \;\; \ \text{s.t.} 
\sum_{a' \in \mathcal{A} \setminus \{a^*\}} \mathbf{w}_{a'} \mu_{a'}^{i}  > \mu_{a^*}^{i}  \quad \forall i \in [d].
\]

This is a contradiction, since $a^* \in \mathcal{F}^*$. Therefore, if $a^* \notin \widehat{\mathcal{F}}_{t+1}$ for any $t <T$, it would imply $a^* \notin \mathcal{F}^*$. Hence, under the good event $\mathcal{E}$, the estimated set of Pareto-optimal arms $\widehat{\mathcal{F}}_t$ is a superset of the true set of Pareto-optimal arms, i.e., $\mathcal{F}^* \subseteq \widehat{\mathcal{F}}_t$ for all $t \in [T]$.

Fix an arm $a \in \mathcal{A}$ that is not Pareto-optimal, i.e., $a \notin \mathcal{F}^*$. 
Let $t \in [T]$ be any time step at which arm $a$ has not yet been eliminated and is therefore active, namely,
\(
a \in \widehat{\mathcal{F}}_{t}
 \text{ and } 
N_a(t) = \min_{b \in \widehat{\mathcal{F}}_{t}} N_b(t).
\)
That is, $t$ is a time step at which arm $a$ is active and has the minimum number of pulls among all active arms.
At such a time $t$, we have
\(
N_{a'}(t) \ge N_a(t) \text{ for all } a' \in \widehat{\mathcal{F}}_{t} \setminus \{a\}.
\)
We use this property to compare confidence radii and show that once $N_a(t)$ exceeds a suitable threshold, the elimination condition for arm $a$ must hold and $a \not\in \widehat{\mathcal{F}}_{t+1}$. The arm $a$ is eliminated when the following condition holds:

\[ \exists\mathbf{w} \in \mathcal{S}(\widehat{\mathcal{F}}_{t} \setminus \{a\}) \;\; \ \text{s.t.} 
\sum_{a' \in \widehat{\mathcal{F}}_{t} \setminus \{a\}} \mathbf{w}_{a'} \mu_{a'}^{-i}(t)  > \mu_{a}^{+i}(t)  \quad \forall i \in [d].
\]

Or equivalently,

\[
\exists \mathbf{w} \in \mathcal{S}(\widehat{\mathcal{F}}_{t} \setminus \{a\}) \;\; \text{s.t.} \quad
\sum_{a' \in \widehat{\mathcal{F}}_{t}\setminus \{a\}} \mathbf{w}_{a'} \mu_{a'}^{-i}(t) - \mu_a^{+i}(t) > 0 \quad \forall i \in [d].
\]

We define the following event $E_a$:

\[
E_a := \Bigg\{ \exists \mathbf{w} \in \mathcal{S}(\widehat{\mathcal{F}}_{t} \setminus \{a\}) \;\; \text{s.t.} \quad
\sum_{a' \in \widehat{\mathcal{F}}_{t} \setminus \{a\}} \mathbf{w}_{a'} \mu_{a'}^{-i}(t) - \mu_a^{+i}(t) > 0 \quad \forall i \in [d] \Bigg\}.
\]

Equivalently, using confidence intervals, we have:
\[
\begin{aligned}
E_a := \Bigg\{ \exists \mathbf{w} \in 
\mathcal{S}(\widehat{\mathcal{F}}_{t} \setminus \{a\}) 
\;\text{s.t.}\;&
\sum_{a' \in \widehat{\mathcal{F}}_{t} \setminus \{a\}} 
\mathbf{w}_{a'} 
\Bigg( \widehat{\mu}_{a'}^{i}(t) 
- \sqrt{\frac{1}{2N_{a'}(t)}
\log\frac{2kdT}{\delta}} \Bigg)  \\
&\quad - \widehat{\mu}_a^i(t) 
+ \sqrt{\frac{1}{2N_a(t)}
\log\frac{2kdT}{\delta}}
> 0,\quad \forall i \in [d]
\Bigg\}.
\end{aligned}
\]

Now, consider a time step $t$ at which the non Pareto-optimal arm $a \notin \mathcal{F}^*$ is played the least number of times, i.e., $N_{a'}(t) \ge N_a(t)$ for all $a' \in \widehat{\mathcal{F}}_{t} \setminus\{a\}$. We define another event $E_a'$ such that $E_a' \implies E_a$:

\[
E_a' := \Bigg\{ \exists \mathbf{w} \in \mathcal{S}(\widehat{\mathcal{F}}_{t} \setminus \{a\}) \;\text{s.t.}\;
\sum_{a' \in \widehat{\mathcal{F}}_{t}\setminus \{a\}} \mathbf{w}_{a'} \widehat{\mu}_{a'}^i(t) 
- \widehat{\mu}_a^i(t) > 2 \sqrt{\frac{1}{2 N_a(t)} \log\frac{2 k d T}{\delta}}, \;\forall i \in [d] 
\Bigg\}.
\]

Under the good event $\mathcal{E}$, we have
\[
\widehat{\mu}_{a'}^i(t) \ge \mu_{a'}^i -  \sqrt{\frac{1}{2 N_{a'}(t)} \log\frac{2 k d T}{\delta}}, \quad
\widehat{\mu}_{a}^i(t) \le \mu_{a}^i +  \sqrt{\frac{1}{2 N_{a}(t)} \log\frac{2 k d T}{\delta}}.
\]
Using this fact and $N_{a'}(t) \ge N_a(t)$ for all $a' \in \widehat{\mathcal{F}}_{t}\setminus\{a\}$, we can define another event $E_a''$ such that $E_a'' \implies E_a'$:

\[
E_a'' := \Bigg\{ \exists \mathbf{w} \in \mathcal{S}(\widehat{\mathcal{F}}_{t} \setminus \{a\}) \;\text{s.t.}\;
\sum_{a' \in \widehat{\mathcal{F}}_{t} \setminus \{a\}} \mathbf{w}_{a'} \mu_{a'}^i
- \mu_a^i > 4 \sqrt{\frac{1}{2 N_a(t)} \log\frac{2 k d T}{\delta}}, \;\forall i \in [d] 
\Bigg\}.
\]

In order to bound $N_a(t)$ for any $a \notin \mathcal{F}^*$, we define
\[
\Delta_a^{\mathrm{elim}} := \max_{\mathbf{w} \in \mathcal{S}(\mathcal{F}^*)} 
\min_{i \in [d]} \sum_{a' \in \mathcal{F}^*} \mathbf{w}_{a'} \mu_{a'}^i - \mu_a^i.
\]
For any non–Pareto-optimal arm $a \notin \mathcal{F}^*$, we have
$\Delta_a^{\mathrm{elim}} > 0$, since there exists a convex combination supported
on $\mathcal{F}^*$ that strictly dominates $a$ in all objectives. Note that Since $\mathcal{F}^* \subseteq \widehat{\mathcal{F}}_t$ on $\mathcal{E}$, any 
$\mathbf{w}\in\mathcal{S}(\mathcal{F}^*)$ is also admissible in 
$\mathcal{S}(\widehat{\mathcal{F}}_t\setminus\{a\})$ by assigning zero weight to 
arms in $\widehat{\mathcal{F}}_t\setminus\mathcal{F}^*$. Under the good event $\mathcal{E}$, once the term $4 \sqrt{\frac{1}{2 N_a(t)} \log \frac{2 k d T}{\delta}} \leq \Delta_a^{\mathrm{elim}}$, we can guarantee that the event $E_a''$ holds, based on the definition of $\Delta_a^{\mathrm{elim}}$. This, in turn, implies that $E_a'$ is true, and consequently $E_a$ is also true. When $E_a$ is true, the corresponding non Pareto-optimal arm $a \notin \mathcal{F}^*$ is eliminated. 
For the arm $a \notin \mathcal{F}^*$ to be eliminated we require:

\begin{equation*}
    4 \sqrt{\frac{1}{2 N_a(t)} \log \frac{2 k d T}{\delta}} \leq \Delta_a^{\mathrm{elim}}
\end{equation*}

\begin{equation*}
   N_a(t) \geq \frac{8 \log(2 k d T / \delta)}{(\Delta_a^{\mathrm{elim}})^2}
\end{equation*}

Once the inequality \(N_a(t) \geq \frac{8 \log(2 k d T / \delta)}{(\Delta_a^{\mathrm{elim}})^2}\) holds at any time step \(t\) at which a non Pareto-optimal arm \(a \notin \mathcal{F}^*\) is still active, i.e., \(a \in \widehat{\mathcal{F}}_t\), and has the minimum number of pulls among all active arms, the elimination condition is satisfied and \(a\) is removed from the next active set, i.e., \(a \notin \widehat{\mathcal{F}}_{t+1}\). Consequently, arm \(a\) cannot be selected after this elimination time. Algorithm~\ref{alg:pareto_ucb_cc} selects an arm with the minimum pull count in the active set  at each time step (breaking ties uniformly at random). We want to show that for any non Pareto-optimal arm $a \notin \mathcal{F}^*$ we have
$N_a(T) \le \left\lceil \frac{8 \log(2 k d T / \delta)}{(\Delta_a^{\mathrm{elim}})^2} \right\rceil$.
By contradiction, assume that an arm $a \notin \mathcal{F}^*$ is played more than
$\left\lceil \frac{8 \log(2 k d T / \delta)}{(\Delta_a^{\mathrm{elim}})^2} \right\rceil$ times.
Then there exists a time $t_a < T$ such that
$N_a(t_a) = \left\lceil \frac{8 \log(2 k d T / \delta)}{(\Delta_a^{\mathrm{elim}})^2} \right\rceil$
and a later time $t$ with $t_a < t \le T$ at which arm $a$ is selected.
Algorithm~\ref{alg:pareto_ucb_cc} always selects an arm with the minimum number of pulls among the active arms,
so for arm $a$ to be selected at time $t > t_a$, it must have the minimum pull count among all active arms at time $t$.
Since $N_a(t_a) = \left\lceil \frac{8 \log(2 k d T / \delta)}{(\Delta_a^{\mathrm{elim}})^2} \right\rceil$,
it follows that $N_a(t) \ge \frac{8 \log(2 k d T / \delta)}{(\Delta_a^{\mathrm{elim}})^2}$.
At the first such time $t > t_a$ when arm $a$ is the least-played (possibly tied) active arm, the elimination condition is satisfied,
and arm $a$ must be removed from the active set and therefore cannot be selected at time $t$,
which yields a contradiction.
Hence, for every non Pareto-optimal arm $a \notin \mathcal{F}^*$,
$N_a(T) \le \left\lceil \frac{8 \log(2 k d T / \delta)}{(\Delta_a^{\mathrm{elim}})^2} \right\rceil$,
or equivalently $N_a(T) = \mathcal{O}  \left(\frac{1}{(\Delta_a^{\mathrm{elim}})^2}
\log  \left(\frac{2 k d T}{\delta}\right)\right)$ under the good event $\mathcal{E}$

We define the cumulative Pareto regret as
\[
R_T = \sum_{a \in \mathcal{A}} \Delta_a^{\mathrm{reg}} \, N_a(T),
\]
where
\[
\Delta_a^{\mathrm{reg}} := \min_{\mathbf{w} \in \mathbb{D}_a} 
\max_{i \in [d]} \sum_{a' \in \mathcal{F}^* \setminus \{a\}} \mathbf{w}_{a'} \mu_{a'}^i - \mu_a^i,
\]
and
\[
\mathbb{D}_a := \Big\{ \mathbf{w} \in \mathcal{S}(\mathcal{F}^* \setminus \{a\}) \;\Big|
\sum_{a' \in \mathcal{F}^* \setminus \{a\}} \mathbf{w}_{a'} \mu_{a'}^i - \mu_a^i \geq 0, \;\forall i \in [d] \Big\}.
\]

Note that $\mathbb{D}_a \neq \emptyset$ for any non Pareto-optimal arm $a \notin \mathcal{F}^*$, and therefore $\Delta_a^{\mathrm{reg}} \geq 0$ for all such arms. However, for Pareto-optimal arms $a^* \in \mathcal{F}^*$, we have $\mathbb{D}_{a^*} = \emptyset$. Since the minimum over an empty set
is undefined, we set $\Delta_{a^*}^{\mathrm{reg}} = 0$ by convention for all
$a^* \in \mathcal{F}^*$. The cumulative Pareto regret can be expressed as
\[
R_T = \sum_{a \notin \mathcal{F}^*} \Delta_a^{\mathrm{reg}} \, N_a(T).
\]

Under the good event $\mathcal{E}$, we have 
\[
R_T \leq \sum_{a \notin \mathcal{F}^*} \Delta_a^{\mathrm{reg}} \; \mathcal{O}  \left(
\frac{1}{(\Delta_a^{\mathrm{elim}})^2}
\log  \left(\frac{2 k d T}{\delta}\right)
\right).
\]

In other words under event  $\mathcal{E}$ , we have
\[
R_T = \mathcal{O}  \left( \sum_{a \notin \mathcal{F}^*} \frac{\Delta_a^{\mathrm{reg}}}{(\Delta_a^{\mathrm{elim}})^2}  \log  \left(\frac{2 k d T}{\delta}\right)
 \right).
\]

This completes the proof of Theorem \ref{thm3}.

\vspace{0.5em}
\crr*
\vspace{0.5em}
\textbf{Proof:}
From the proof of Theorem~\ref{thm3}, on the good event $\mathcal{E}$ (which occurs with probability at least $1-\delta$),
any Pareto-optimal arm $a^* \in \mathcal{F}^*$ is never eliminated at any time.
Moreover, any non Pareto-optimal arm $a \notin \mathcal{F}^*$ is played at most
$N_a(T) \le \left\lceil \frac{8 \log(2 k d T / \delta)}{(\Delta_a^{\mathrm{elim}})^2} \right\rceil$ times.

We now show that any   non  Pareto-optimal arm $a \notin \mathcal{F}^*$ that has not yet been eliminated will necessarily be eliminated within a bounded number of subsequent steps once a time step $t_a$ is reached such that
\(
N_a(t_a) = \left\lceil \frac{8 \log(2 k d T / \delta)}{(\Delta_a^{\mathrm{elim}})^2} \right\rceil 
\). Once a time step $t_a$ is reached such that
$N_a(t_a) = \left\lceil \frac{8 \log(2 k d T / \delta)}{(\Delta_a^{\mathrm{elim}})^2} \right\rceil$
and arm $a$ is still active, we know that it cannot be selected any further,
since any non Pareto-optimal arm $a \notin \mathcal{F}^*$ is played at most
$N_a(T) \le \left\lceil \frac{8 \log(2 k d T / \delta)}{(\Delta_a^{\mathrm{elim}})^2} \right\rceil$ times. Since the Algorithm \ref{alg:pareto_ucb_cc} always selects a least-sampled active arm and the total number
of arms is \(k\), within at most \(k\) consecutive rounds every active arm attains the minimum count (possibly tied) at least once.
Consequently, within a finite number of subsequent steps, specifically at most
$k = |\mathcal{A}|$, arm $a$ will be the arm in the active set with the minimum number of pulls,
that is, at some time $t$ satisfying $t_a \le t \le t_a + k$.
At this time, we have
$N_a(t) \ge \frac{8 \log(2 k d T / \delta)}{(\Delta_a^{\mathrm{elim}})^2}$,
and since arm $a$ is the least-sampled (possibly tied) active arm, the elimination condition is triggered.
Therefore, any non Pareto-optimal arm $a \notin \mathcal{F}^*$ is eliminated no later than time step $t_a + k$.

\textbf{Stopping Criterion for Algorithm~\ref{alg:pareto_ucb_cc}:}
A sufficient stopping criterion for Algorithm~\ref{alg:pareto_ucb_cc} is to terminate once
$t > \tau + k$, where $\tau$ is the smallest time step such that
\[
\min_{a \in \widehat{\mathcal{F}}_\tau} N_a(\tau)
\;>\;
\frac{8\log(2kdT/\delta)}{\Delta^2},
\]
with $\Delta := \min_{a \notin \mathcal{F}^*} \Delta_a^{\mathrm{elim}}$.
Under this condition, with probability at least $1-\delta$, all remaining active arms in
$\widehat{\mathcal{F}}_t$ are truly Pareto-optimal. This follows because, under the good event $\mathcal{E}$, we have
$\mathcal{F}^* \subseteq \widehat{\mathcal{F}}_t$ for all $t$, and for each arm
$a \notin \mathcal{F}^*$ there exists a time $t_a$ such that
$N_a(t_a) = \left\lceil 8\log(2kdT/\delta)/(\Delta_a^{\mathrm{elim}})^2 \right\rceil$,
at which point arm $a$ is eliminated no later than time $t_a + k$.
Consequently, any arm $a \notin \mathcal{F}^*$ cannot belong to $\widehat{\mathcal{F}}_t$
for any $t > t_a + k$.
Therefore, once $t > \tau + k$, where $\tau$ is the smallest time step such that
\(
\min_{a \in \widehat{\mathcal{F}}_\tau} N_a(\tau)
\;>\;
\frac{8\log(2kdT/\delta)}{\Delta^2}
\), all non–Pareto-optimal arms have already been eliminated, and
all remaining active arms in $\widehat{\mathcal{F}}_t$ are Pareto-optimal. In practice, since $\Delta$ is unknown, it may be replaced by a conservative lower bound. We can now define the termination time as follows:
\[
T_0(\delta) :=  k +\min \Bigg\{ \tau \in \mathbb{N}  \;\Big|\;
N_a(\tau) > \frac{ 8\log(2 k d T / \delta)}{\Delta^2},
\;\forall a \in \widehat{\mathcal{F}}_{\tau} \Bigg\}.
\]

Under the event $\mathcal{E}$, all non  Pareto-optimal arms are eliminated by time $T_0(\delta)$,
while all Pareto-optimal arms remain active. Therefore, under event $\mathcal{E}$ we have
$\widehat{\mathcal{F}}_t = \mathcal{F}^*$ for all $t > T_0(\delta)$. Hence,
\(
\mathbb{P}\!\left(\widehat{\mathcal{F}}_t = \mathcal{F}^* \ \text{for all } t > T_0(\delta)\right) \ge 1-\delta 
\). This completes the proof of Corollary~\ref{crr1}, together with the justification of the stopping criterion for
Algorithm~\ref{alg:pareto_ucb_cc}.

\subsection{Anytime Version of Algorithm~\ref{alg:pareto_ucb_cc} and Its Analysis}
\label{pff_ap1}

\begin{theorem}[\textbf{Anytime regret and identification guarantees}]
\label{thm_app_1}
\hypertarget{keypoint}{}
For Algorithm~\ref{alg:pareto_ucb_cc_any}, with probability at least $1-\delta$, the following hold:
\vspace{-0.5em}
\begin{itemize}
    \item For all times $t \in \mathbb{N}$, any true Pareto-optimal arm is never eliminated, i.e.,
$\mathcal{F}^* \subseteq \widehat{\mathcal{F}}_t$.
 \item  Moreover, for any (arbitrary) evaluation time $T \in \mathbb{N}$, every non–Pareto-optimal arm
$a \notin \mathcal{F}^*$ satisfies
$N_a(T) = \mathcal{O}  \left(\frac{\log  \left(\pi^2 k d T^2 /3\delta\right)}{(\Delta_a^{\mathrm{elim}})^2}\right)$,
and the cumulative Pareto regret obeys
$R_T = \mathcal{O}  \left(\sum_{a \notin \mathcal{F}^*}
\frac{\Delta_a^{\mathrm{reg}}}{(\Delta_a^{\mathrm{elim}})^2}\log{\frac{\pi^2kdT^2}{3\delta}}\right)$.
Here, $T$ is used only as an evaluation time and is not known to the algorithm.
\end{itemize}
\vspace{-1em}
Furthermore, letting $k := |\mathcal{A}|$ and
$\Delta := \min_{a \notin \mathcal{F}^*} \Delta_a^{\mathrm{elim}}$, define the (horizon-free)
termination time $T_0(\delta) :=  \tau +k$, where $\tau$ is the smallest time step satisfying $\min_{a \in \widehat{\mathcal{F}}_\tau} N_a(\tau) > \frac{ 8\log(\pi^2 k d \;(\tau+k)^2 / 3\delta)}{\Delta^2}$. Then
$\mathbb{P}\big( \widehat{\mathcal{F}}_t = \mathcal{F}^*,\ \forall\, t > T_0(\delta) \big) \ge 1 - \delta$.
\end{theorem}

\begin{algorithm}[h]
\caption{Pareto-UCB with Convex Combinations \textbf{(Anytime Version)}}
\label{alg:pareto_ucb_cc_any}

\KwIn{Action set $\mathcal{A}$,confidence $\delta \in (0,1)$}
\textbf{Initialize:} $N_a = 0$,$\;\widehat{\mathcal{F}}_{|\mathcal{A}|+1}=\mathcal{A}$,$\;\widehat{\mu}_a^i = 0$ $\forall a\in\mathcal{A} , i\in[d]$\\
\vspace{0.1em}
Play every $a \in \mathcal{A}$ once and update estimates/counts

\For{$t = |\mathcal{A}|+1, |\mathcal{A}|+2, \ldots$}{
    
       \For{$a \in \widehat{\mathcal{F}}_t,\; i \in [d]$}{
        $\;\;\mu_a^{+i}(t) \gets \widehat{\mu}_a^i(t) 
        + \sqrt{\frac{\log(\pi^2kd\;t^2/3\delta)}{2N_a(t)}}$\\
        $\mu_a^{-i}(t) \gets \widehat{\mu}_a^i(t)
        - \sqrt{\frac{\log(\pi^2kd\;t^2/3\delta)}{2N_a(t)}}$\;
    }

    Eliminate any $a \in \widehat{\mathcal{F}}_t$ for which $\exists \{\mathbf{w}_{a'} \ge 0\}_{\substack{a' \in \widehat{\mathcal{F}}_t \\ a' \neq a}}$
 s.t. \vspace{-1.1em}
    \[
    \sum_{\substack{a' \in \widehat{\mathcal{F}}_t \\ a' \neq a}}
 \mathbf{w}_{a'} = 1,\;
        \sum_{\substack{a' \in \widehat{\mathcal{F}}_t \\ a' \neq a}} \mathbf{w}_{a'} \mu_{a'}^{-i}(t)
  > \mu_a^{+i}(t)\;\; \forall i \in [d]
    \vspace{-0.8em}\] 
    Let $\widehat{\mathcal{F}}_{t+1}$ be the set of remaining active arms.\\
    Play $A_t \gets \arg\min_{a \in \widehat{\mathcal{F}}_{t+1}} N_a(t)$ (Break ties uniformly at random)  and update estimates/counts.
}
\end{algorithm}
\vspace{0.5em}
\textbf{Proof:} Suppose $\widehat{\mu}_a^i(t)$ is the estimated reward for objective $i \in [d]$ for action $a \in \mathcal{A}$, and let $N_a(t)$ be the number of times action $a$ has been played up to time step $t$. By Hoeffding's inequality, we have
\[
\mathbb{P}\Big(\big|\widehat{\mu}_a^i(t) - \mu_a^i\big| \ge \epsilon\Big) \le 2 \exp\big(-2 \epsilon^2 N_a(t)\big).
\]

Setting $2 \exp(-2 \epsilon^2 N_a(t)) = \delta_t$, we solve for $\epsilon$ to get
\[
\epsilon = \sqrt{\frac{1}{2 N_a(t)} \log\left( \frac{2}{\delta_t} \right)}.
\]

Thus,
\[
\mathbb{P}\Big(\big|\widehat{\mu}_a^i(t) - \mu_a^i\big| \ge \sqrt{\frac{1}{2 N_a(t)} \log\left( \frac{2}{\delta_t} \right)} \Big) \le \delta_t,
\]
which is equivalent to
\[
\mathbb{P}\Big(\big|\widehat{\mu}_a^i(t) - \mu_a^i\big| \le \sqrt{\frac{1}{2 N_a(t)} \log\left( \frac{2}{\delta_t} \right)} \Big) \ge 1 - \delta_t.
\]

Let us define the event
\[
\mathcal{E}_a^i(t) := \Big\{ \big|\widehat{\mu}_a^i(t) - \mu_a^i\big| \le \sqrt{\frac{1}{2 N_a(t)} \log\left( \frac{2}{\delta_t} \right)} \Big\}.
\]

 Next, define the intersection of all such events:
\[
\mathcal{E}' := \bigcap_{\substack{i \in [d] \\ a \in \mathcal{A} \\ t \in [\mathbb{N}]}} \mathcal{E}_a^i(t).
\]

where $[\mathbb{N}] =\{1,2,....\}$ is set of natural numbers. Since $\mathbb{P}(\mathcal{E}_a^i(t)) \ge 1 - \delta_t$ for each $i, a, t$, using the union bound we have
\[
\mathbb{P}(\mathcal{E}') = 1 - \mathbb{P}\Big(\bigcup_{i\in[d],a\in \mathcal{A},t \in [\mathbb{N}]} \overline{\mathcal{E}_a^i(t)}\Big) 
\ge 1 - \sum_{i\in[d],a\in \mathcal{A},t \in [\mathbb{N}]} \mathbb{P}\big(\overline{\mathcal{E}_a^i(t)}\big) 
\ge 1 - k d   \sum_{t = 1}^{\infty}\delta_t
\]
where $k = |\mathcal{A}|$. If we choose $\delta_t = \frac{6 \delta}{\pi^2kdt^2}$, then by the union bound we have $\mathbb{P}(\mathcal{E}') \ge 1 - \delta$. This is because $\sum_{t = 1}^{\infty} \delta_t = \frac{6 \delta}{\pi^2kd} \sum_{t = 1}^{\infty} \frac{1}{t^2} = \frac{\delta}{kd}$ using the fact that $\sum_{t = 1}^{\infty} \frac{1}{t^2} =\frac{\pi^2}{6}$. The event $\mathcal{E}'$ ensures that, for all time steps, the true expected reward for every objective and every action lies within the corresponding confidence interval; that is,

\[
\mathcal{E}' := \left\{
\begin{aligned}
& \forall a \in \mathcal{A} \\
& \forall t \in [\mathbb{N}] \quad \quad
\big|\widehat{\mu}_a^i(t) - \mu_a^i\big| \le \sqrt{\frac{1}{2 N_a(t)} \log\left(\frac{ \pi^2 k d \;t^2}{3\delta}\right)} \\
& \forall i \in [d]
\end{aligned}
\right\}.
\]

 The probability of this good event satisfies $\mathbb{P}(\mathcal{E}') \ge 1 - \delta$. Also we use the notations $\mu^{+i}_a(t) := \widehat{\mu}_a^i(t)  + \sqrt{\frac{1}{2 N_a(t)} \log\left(\frac{\pi^2 k d \;t^2}{3\delta}\right)}$  and $\mu^{-i}_a(t) := \widehat{\mu}_a^i(t)  - \sqrt{\frac{1}{2 N_a(t)} \log\left(\frac{\pi^2 k d \;t^2}{3\delta}\right)}$.

Under the good event $\mathcal{E}'$, the estimated set of Pareto-optimal arms $\widehat{\mathcal{F}}_t$ is a superset of the true Pareto-optimal set, i.e., $\mathcal{F}^* \subseteq \widehat{\mathcal{F}}_t$ for all $t \in \mathbb{N}$. This follows directly from the proof of Theorem~\ref{thm3} in Section~\ref{pff_ap}, together with the definition of the good event $\mathcal{E}'$. The only difference between the anytime version, Algorithm~\ref{alg:pareto_ucb_cc_any}, and the horizon-dependent version, Algorithm~\ref{alg:pareto_ucb_cc}, is the form of the confidence intervals. Fix an arm $a \in \mathcal{A}$ that is not Pareto-optimal, i.e., $a \notin \mathcal{F}^*$. 
Let $t$ be any time step at which arm $a$ has not yet been eliminated and is therefore active, namely,
\(
a \in \widehat{\mathcal{F}}_{t}
 \text{ and } 
N_a(t) = \min_{b \in \widehat{\mathcal{F}}_{t}} N_b(t).
\)
That is, $t$ is a time step at which arm $a$ is active and has the minimum number of pulls among all active arms.
At such a time $t$, we have
\(
N_{a'}(t) \ge N_a(t) \text{ for all } a' \in \widehat{\mathcal{F}}_{t} \setminus \{a\}.
\)
The arm $a$ is eliminated when the following event happens:

\[
E_a := \Bigg\{ \exists \mathbf{w} \in \mathcal{S}(\widehat{\mathcal{F}}_{t} \setminus \{a\}) \;\; \text{s.t.} \quad
\sum_{a' \in \widehat{\mathcal{F}}_{t} \setminus \{a\}} \mathbf{w}_{a'} \mu_{a'}^{-i}(t) - \mu_a^{+i}(t) > 0 \quad \forall i \in [d] \Bigg\}.
\]

Under the good event $\mathcal{E}'$, all confidence bounds hold, and we assume that at time step $t$ the arm $a$ is a least-played active arm. We can then define another event $E_a'$ such that, whenever $E_a'$ holds, it implies that $E_a$ also holds.

\[
E_a' := \Bigg\{ \exists \mathbf{w} \in \mathcal{S}(\widehat{\mathcal{F}}_{t} \setminus \{a\}) \;\text{s.t.}\;
\sum_{a' \in \widehat{\mathcal{F}}_{t} \setminus \{a\}} \mathbf{w}_{a'} \mu_{a'}^i
- \mu_a^i > 4 \sqrt{\frac{1}{2 N_a(t)} \log\frac{\pi^2 k d \;t^2}{3\delta}}, \;\forall i \in [d] 
\Bigg\}.
\]

In order to bound $N_a(t)$, recall that for any $a \notin \mathcal{F}^*$, we define
\[
\Delta_a^{\mathrm{elim}}
\;:=\;
\max_{\mathbf{w} \in \mathcal{S}(\mathcal{F}^*)}
\;
\min_{i \in [d]}
\left(
\sum_{a' \in \mathcal{F}^*} \mathbf{w}_{a'} \mu_{a'}^i
\;-\;
\mu_a^i
\right).
\]

Under the good event $\mathcal{E}'$, once the confidence radius satisfies
\(
4 \sqrt{\frac{1}{2 N_a(t)} \log \frac{\pi^2 k d \, t^2}{3\delta}}
\;\le\;
\Delta_a^{\mathrm{elim}},
\)
we can guarantee that the event $E_a'$ holds, by the definition of $\Delta_a^{\mathrm{elim}}$.
This, in turn, implies that $E_a$ holds, since under the good event $\mathcal{E}'$ all confidence bounds
are valid and arm $a$ is a least-played active arm at time $t$.  When $E_a$ is true, the corresponding non Pareto-optimal arm $a \notin \mathcal{F}^*$ is eliminated. 
For the arm $a \notin \mathcal{F}^*$ to be eliminated we require:

\[
4 \sqrt{\frac{1}{2 N_a(t)} \log \frac{\pi^2 k d \, t^2}{3\delta}}
\;\le\;
\Delta_a^{\mathrm{elim}},
\]
which is equivalent to
\[
N_a(t)
\;\ge\;
\frac{8 \log  \left(\pi^2 k d \, t^2 / 3\delta\right)}
{(\Delta_a^{\mathrm{elim}})^2}.
\]

In order to show that the cumulative regret grows logarithmically with time,
we fix a arbitrary horizon $T \in [\mathbb{N}]$ and decompose the cumulative Pareto regret as
\[
R_T
\;=\;
\sum_{a \in \mathcal{A}}
\Delta_a^{\mathrm{reg}} \, N_a(T).
\]

Assuming that time $t$ ranges from $1$ to the horizon $T$, we replace the condition
$N_a(t) \ge \frac{8 \log(\pi^2 k d \, t^2 / 3\delta)}{(\Delta_a^{\mathrm{elim}})^2}$
by
$N_a(t) \ge \frac{8 \log(\pi^2 k d \, T^2 / 3\delta)}{(\Delta_a^{\mathrm{elim}})^2}$ since $t \leq T$.
We can use the same argument as in the proof of Theorem~\ref{thm3} in
Section~\ref{pff_ap} to show that, for every  non  Pareto-optimal arm
$a \notin \mathcal{F}^*$, we have
$N_a(T) \le \left\lceil \frac{8 \log(\pi^2 k d \, T^2 / 3\delta)}{(\Delta_a^{\mathrm{elim}})^2} \right\rceil$,
or equivalently
$N_a(T) = \mathcal{O}  \left(\frac{1}{(\Delta_a^{\mathrm{elim}})^2}
\log  \left(\frac{\pi^2 k d \, T^2}{3\delta}\right)\right)$.
By contradiction, suppose that an arm $a \notin \mathcal{F}^*$ is played more than
$\left\lceil \frac{8 \log(\pi^2 k d \, T^2 / 3\delta)}{(\Delta_a^{\mathrm{elim}})^2} \right\rceil$
times. Then there exists a time $t_a < T$ such that
$N_a(t_a) = \left\lceil \frac{8 \log(\pi^2 k d \, T^2 / 3\delta)}{(\Delta_a^{\mathrm{elim}})^2} \right\rceil$
and a later time $t$ with $t_a < t \le T$ at which arm $a$ is selected.
Algorithm~\ref{alg:pareto_ucb_cc_any} always selects an arm with the minimum number of pulls
among the active arms, so for arm $a$ to be selected at time $t > t_a$, it must
have the minimum pull count among all active arms at time $t$.
Since $N_a(t_a) = \left\lceil \frac{8 \log(\pi^2 k d \, T^2 / 3\delta)}{(\Delta_a^{\mathrm{elim}})^2} \right\rceil$,
it follows that
$N_a(t) \ge \frac{8 \log(\pi^2 k d \, T^2 / 3\delta)}{(\Delta_a^{\mathrm{elim}})^2}$.
At the first such time $t > t_a$ when arm $a$ is the least-played (possibly tied) active arm, the
elimination condition is satisfied, and arm $a$ must be removed from the active set,
which yields a contradiction.
Hence, for every  non  Pareto-optimal arm $a \notin \mathcal{F}^*$,
$N_a(T) \le \left\lceil \frac{8 \log(\pi^2 k d \, T^2 / 3\delta)}{(\Delta_a^{\mathrm{elim}})^2} \right\rceil$ under the good event $\mathcal{E}'$

Under the good event $\mathcal{E}'$, we have
\[
R_T \le \sum_{a \notin \mathcal{F}^*}
\Delta_a^{\mathrm{reg}} \,
\mathcal{O}  \left(
\frac{1}{(\Delta_a^{\mathrm{elim}})^2}
\log  \left(\frac{\pi^2 k d \, T^2}{3\delta}\right)
\right).
\]
In other words, under the event $\mathcal{E}'$,
\[
R_T
=
\mathcal{O}  \left(
\sum_{a \notin \mathcal{F}^*}
\frac{\Delta_a^{\mathrm{reg}}}{(\Delta_a^{\mathrm{elim}})^2}
\log  \left(\frac{\pi^2 k d \, T^2}{3\delta}\right)
\right).
\]

We now characterize a stopping time after which we can guarantee that Algorithm~\ref{alg:pareto_ucb_cc_any} has identified the true set of Pareto-optimal arms $\mathcal{F}^*$ with high probability. As shown above, on the good event $\mathcal{E}'$ (which occurs with probability at least $1-\delta$), any Pareto-optimal arm $a^* \in \mathcal{F}^*$ is never eliminated at any time. We now show that any   non  Pareto-optimal arm $a \notin \mathcal{F}^*$ that has not yet been eliminated will necessarily be eliminated within a bounded number of subsequent steps once a time step $t_a$ is reached such that $N_a(t_a) = \left\lceil \frac{8 \log\ \left(\pi^2 k d (t_a + k)^2 / 3\delta\right)}{(\Delta_a^{\mathrm{elim}})^2} \right\rceil$. Suppose that $t_a$ is the first time step at which this condition holds for a   non  Pareto-optimal arm $a \notin \mathcal{F}^*$ that has not yet been eliminated. Since Algorithm~\ref{alg:pareto_ucb_cc_any} always selects an arm with the minimum number of pulls among the active arms, it follows that within at most $k := |\mathcal{A}|$ subsequent steps, arm $a$ must become the least-sampled (possibly tied) active arm. Consequently, there exists a time $t$ satisfying $t_a \le t \le t_a + k$ such that arm $a$ is the least-sampled (possibly tied) arm among all active arms at time $t$. Moreover, since pull counts are non-decreasing over time, we have $N_a(t) \ge N_a(t_a)$, and by the definition of $t_a$ this implies $N_a(t) \ge \left\lceil \frac{8 \log  \left(\pi^2 k d t^2 / 3\delta\right)}{(\Delta_a^{\mathrm{elim}})^2} \right\rceil$, which guarantees that the elimination condition for arm $a$ is satisfied at time $t$. Therefore, arm $a$ is eliminated at or before time $t_a + k$.

\textbf{Stopping Criterion
for Algorithm \ref{alg:pareto_ucb_cc_any}:} A sufficient stopping criterion for Algorithm~\ref{alg:pareto_ucb_cc_any} is to terminate once
$t > \tau + k$, where $\tau$ is the smallest time step such that
\[
\min_{a \in \widehat{\mathcal{F}}_\tau} N_a(\tau)
\;>\;
\frac{8\log  \big(\pi^2 k d (\tau+k)^2/(3\delta)\big)}{\Delta^2},
\]
with $\Delta := \min_{a \notin \mathcal{F}^*} \Delta_a^{\mathrm{elim}}$.
In this case, with probability at least $1-\delta$, all remaining active arms in $\widehat{\mathcal{F}}_t$ are Pareto-optimal. This follows because under the good event $\mathcal{E}'$ we have
$\mathcal{F}^* \subseteq \widehat{\mathcal{F}}_t$ for all $t$,
and for each arm $a \notin \mathcal{F}^*$, there exists a time $t_a$ such that
\(
N_a(t_a)
=
\left\lceil
\frac{8 \log  \big(\pi^2 k d (t_a + k)^2/(3\delta)\big)}
{(\Delta_a^{\mathrm{elim}})^2}
\right\rceil,
\)
at which point arm $a$ is eliminated no later than time $t_a + k$.
Consequently, any arm $a \notin \mathcal{F}^*$ cannot belong to $\widehat{\mathcal{F}}_t$
for any $t > t_a + k$.
Therefore, once $t > \tau + k$, where $\tau$ is the smallest time step such that
\(
\min_{a \in \widehat{\mathcal{F}}_\tau} N_a(\tau)
\;>\;
\frac{8\log  \big(\pi^2 k d (\tau+k)^2/(3\delta)\big)}{\Delta^2}
\), all non–Pareto-optimal arms have already been eliminated, and
all remaining active arms in $\widehat{\mathcal{F}}_t$ are Pareto-optimal. In practice, since $\Delta$ is unknown, one may replace it by a conservative lower bound. We can now define the termination time as follows:
\[
T_0(\delta) :=  k +\min \Bigg\{ \tau \in \mathbb{N}  \;\Big|\;
N_a(\tau) > \frac{8\log  \big(\pi^2 k d (\tau+k)^2/(3\delta)\big)}{\Delta^2},
\;\forall a \in \widehat{\mathcal{F}}_{\tau} \Bigg\}.
\]

Under the event $\mathcal{E}'$, all non  Pareto-optimal arms are eliminated by time $T_0(\delta)$,
while all Pareto-optimal arms remain active. Therefore, under event $\mathcal{E}'$ we have
$\widehat{\mathcal{F}}_t = \mathcal{F}^*$ for all $t > T_0(\delta)$. Hence,
\(
\mathbb{P}\!\left(\widehat{\mathcal{F}}_t = \mathcal{F}^* \ \text{for all } t > T_0(\delta)\right) \ge 1-\delta 
\). This completes the proof of Theorem \hyperlink{keypoint}{A.1} together with the justification of the stopping criterion for Algorithm \ref{alg:pareto_ucb_cc_any}.

\subsection{Algorithms for Multi-Objective Causal Bandits with Linear Scalarization}
\label{secalg}

We consider the simpler setting in which the learner maximizes a fixed linear scalarization of the reward vector. Let $\mathrm{PPOMISs}(\mathcal{G},\mathbf{Y})=\{\mathbf{X}_1,\ldots,\mathbf{X}_N\}$ be the intervention sets returned by Algorithm~\ref{alg:final_ppomis}. The induced deterministic intervention action set is
\[
\mathcal{A}
=
\{\,do(\mathbf{X}=\mathbf{x}) \mid \mathbf{X}\in \mathrm{PPOMISs}(\mathcal{G},\mathbf{Y}),\ \mathbf{x}\in\mathcal{D}(\mathbf{X})\,\}.
\]
We relabel this finite set as $\mathcal{A}=\{a_1,\ldots,a_k\}$, where $k=|\mathcal{A}|$, and write $\mu_a^i=\mathbb{E}[Y_i\mid a]$ for the expected reward of action $a$ on objective $i\in[d]$.

For a known weight vector $\mathbf{w}\in\mathcal{S}(\mathbf{Y})$, define the scalarized reward as
\[
R(\mathbf{Y}) := \mathbf{w}^{\top}\mathbf{Y}.
\]
For any stochastic policy over interventions, the expected scalarized reward is a convex combination of the scalarized rewards of deterministic interventions. Hence, in the unconstrained linear-scalarization setting, an optimal policy can be chosen to be deterministic. Therefore, the optimal action is
\[
a^*
\in
\operatorname*{arg\,max}_{a\in\mathcal{A}}
\sum_{i=1}^{d} \mathbf{w}_i\mu_a^i .
\]
Thus, unlike the Pareto-frontier setting, randomization is not needed for optimality under a single fixed linear scalarization without constraints. We can therefore apply standard bandit algorithms to the scalar reward induced by $\mathbf{w}$. We present adaptations of Upper Confidence Bound and Thompson Sampling for this scalarized problem. FIn the UCB version, we compute one scalar UCB for each action using the known weight vector $\mathbf{w}$ and select the action with the largest UCB. For Thompson Sampling, we follow the Bernoulli-resampling reduction of~\citep{agrawal2012analysis} for stochastic rewards supported on $[0,1]$. Both algorithms achieve sublinear regret, as stated in Theorem~\ref{thm4}.

\begin{restatable}{theorem}{thmmmmm}
\label{thm4}
Under the linear-scalarization setting, for Algorithm~\ref{alg:ucb_sc}, if we set $\delta=d/T^2$, then
\[
\mathbb{E}[R_T]
\le
(d+1)\sum_{a\in\mathcal{A}}\Delta_a
+
4\sqrt{kT\log(2T^2)}.
\]
For Algorithm~\ref{alg:ts_sc} with a Beta posterior, we have
\[
\mathbb{E}[R_T]=\mathcal{O}(\sqrt{kT\log T}).
\]
\end{restatable}

\begin{algorithm}[h]
\caption{Weighted UCB for Linear Scalarization}
\label{alg:ucb_sc}

\KwIn{Action set $\mathcal{A}$, horizon $T$, confidence level $\delta\in(0,1)$, weight vector $\mathbf{w}\in\mathcal{S}(\mathbf{Y})$}
\textbf{Initialize:} $N_a\gets 0$ and $\widehat{\mu}_a^i\gets 0$ for all $a\in\mathcal{A}$ and $i\in[d]$\;
Play each $a\in\mathcal{A}$ once and update $N_a$ and $\widehat{\mu}_a^i$ for all $i\in[d]$\;

\For{$t=|\mathcal{A}|+1$ \KwTo $T$}{
    \For{$a\in\mathcal{A}$ and $i\in[d]$}{
        $\mu_a^{+i}(t)\gets \widehat{\mu}_a^i(t)+\sqrt{\frac{\log(2d/\delta)}{2N_a(t)}}$\;
    }

    \For{$a\in\mathcal{A}$}{
        $\mathrm{UCB}_a(t)\gets \sum_{i=1}^d \mathbf{w}_i\mu_a^{+i}(t)$\;
    }

    Play $A_t\gets \arg\max_{a\in\mathcal{A}}\mathrm{UCB}_a(t)$ and update $N_{A_t}$ and $\widehat{\mu}_{A_t}^i$ for all $i\in[d]$\;
}
\end{algorithm}

\begin{algorithm}[h]
\caption{Weighted Thompson Sampling for Linear Scalarization}
\label{alg:ts_sc}

\KwIn{Action set $\mathcal{A}$, horizon $T$, weight vector $\mathbf{w}\in\mathcal{S}(\mathbf{Y})$}
\textbf{Initialize:} $S_a\gets 0$ and $F_a\gets 0$ for all $a\in\mathcal{A}$\;

\For{$t=1$ \KwTo $T$}{
    Sample $\Theta_a(t)\sim\mathrm{Beta}(S_a+1,F_a+1)$ for all $a\in\mathcal{A}$\;
    Play $A_t\gets \arg\max_{a\in\mathcal{A}}\Theta_a(t)$\;
    Observe $Y_i$ for all $i\in[d]$ and compute $R_t=\sum_{i=1}^d \mathbf{w}_iY_i$\;
    Sample $R_t'\sim\mathrm{Ber}(R_t)$\;
    $S_{A_t}\gets S_{A_t}+R_t'$, \quad $F_{A_t}\gets F_{A_t}+1-R_t'$\;
}
\end{algorithm}
\subsection{Proof of Theorem \ref{thm4} }
\vspace{0.5em}
\thmmmmm*
\vspace{0.5em}
\textbf{Proof:} We start with Algorithm~\ref{alg:ucb_sc}. Using standard concentration inequalities, we know that for an action $a \in \mathcal{A}$ and an objective $i \in [d]$,
\[
\widehat{\mu}^i_a(t) - \sqrt{\frac{1}{2 N_a(t)} \log \frac{2}{\delta}} 
\leq \mu_a^i \leq 
\widehat{\mu}^i_a(t) + \sqrt{\frac{1}{2 N_a(t)} \log \frac{2}{\delta}}, 
\quad \text{with probability at least } 1-\delta.
\]

Using the union bound over all objectives $i \in [d]$ for action $a \in \mathcal{A}$, we have
\[
\widehat{\mu}^i_a(t)  - \sqrt{\frac{1}{2 N_a(t)} \log \frac{2}{\delta}} 
\leq \mu_a^i \leq 
\widehat{\mu}^i_a(t)  + \sqrt{\frac{1}{2 N_a(t)} \log \frac{2}{\delta}}, 
\quad \forall i \in [d], \;\; \text{with probability at least } 1-d\delta.
\]

Equivalently, for action $a \in \mathcal{A}$, we can write
\[
\widehat{\mu}^i_a (t) - \sqrt{\frac{1}{2 N_a(t)} \log \frac{2d}{\delta}} 
\leq \mu_a^i \leq 
\widehat{\mu}^i_a(t)  + \sqrt{\frac{1}{2 N_a(t)} \log \frac{2d}{\delta}}, 
\quad \forall i \in [d], \;\; \text{with probability at least } 1-\delta,
\]
where $N_a(t)$ denotes the number of times action $a$ has been selected up to time step $t$. The true expected reward for any action $a \in \mathcal{A}$ can be written as $\mu_a := \mathbb{E}[R(\mathbf{Y}) \mid a] = \sum_{i=1}^{d} \mathbf{w}_i \mu_a^i$, where $\mathcal{A} = \{a_1, \dots, a_k\}$ and $k$ is the total number of interventions across all PPOMISs. The optimal action is thus $a^* := \operatorname*{arg\,max}_{a \in \mathcal{A}} \mu_a$, and the sub-optimality gap for $a \neq a^*$ is defined as $\Delta_a := \mu_{a^*} - \mu_a$. The expected cumulative regret is therefore $\mathbb{E}[R_T] = \sum_{a \in \mathcal{A}} \Delta_a \,\mathbb{E}[N_a(T)]$. 

We also define the weighted empirical mean for action $a$ as $\widehat{\mu}_a(t) := \sum_{i=1}^{d} \mathbf{w}_i \widehat{\mu}_a^i(t)$, and the empirical mean after $\beta_a$ plays as $\widehat{\mu}_{a(\beta_a)} := \sum_{i=1}^{d} \mathbf{w}_i \widehat{\mu}_{a(\beta_a)}^i$, where $\widehat{\mu}_{a(\beta_a)}^i$ is the empirical mean of objective $i$ for action $a$ after $\beta_a$ selections. The upper confidence bound for each action is defined as $UCB_a(t) = \sum_{i =1}^d \mathbf{w}_i \mu_a^{+i}(t)$, where $\mu_a^{+i}(t) := \widehat{\mu}_a^i(t) + \sqrt{\frac{\log(2d/\delta)}{2 N_a(t)}}$. We then play the action $A_t = \operatorname*{arg\,max}_{a \in \mathcal{A}} UCB_a(t)$.

We define an event $\mathcal{E}_{a}$ as follows:

\[
\mathcal{E}_{a} := \Big\{ \mu_{a^*} < \min_{t \in [T]} UCB_{a^*}(t) \Big\} 
\;\; \cap \;\; 
\Big\{ \widehat{\mu}_{a(\beta_a)} + \sqrt{\frac{\log(2d/\delta)}{2 \beta_a}} < \mu_{a^*} \Big\}.
\]

When the event $\mathcal{E}_a$ occurs, we have $N_a(T) \leq \beta_a$. Suppose, for contradiction, that arm $a$ is selected more than $\beta_a$ times over $T$ rounds. Then there must exist a round $t \in [T]$ such that $N_a(t) = \beta_a$ and $A_t = a$. Under the event $\mathcal{E}_a$, we then have

\[
UCB_a(t) = \widehat{\mu}_{a(\beta_a)} + \sqrt{\frac{\log(2d/\delta)}{2 \beta_a}} \leq \mu_{a^*} \leq UCB_{a^*}(t).
\]

This implies that, at any such time step $t$, we must have $A_t \neq a$. Therefore, the event $\mathcal{E}_a$ guarantees that $N_a(T) \leq \beta_a$.

Now we try to bound the probability of $\mathcal{E}_a^c$. Note that

\[
\mathcal{E}_{a}^c := \Big\{ \mu_{a^*} \geq \min_{t \in [T]} UCB_{a^*}(t) \Big\} 
\;\; \cup \;\; 
\Big\{ \widehat{\mu}_{a(\beta_a)} + \sqrt{\frac{\log(2d/\delta)}{2 \beta_a}} \geq \mu_{a^*} \Big\}.
\]

First, consider the first event:

\begin{align}
\mathbb{P}\Big( \mu_{a^*} \geq \min_{t \in [T]} UCB_{a^*}(t) \Big) 
&\leq \mathbb{P}\Big( \bigcup_{t \in [T]} \{ \mu_{a^*} \geq UCB_{a^*}(t) \} \Big) \nonumber\\
&\leq \sum_{t \in [T]} \mathbb{P}\Big( \mu_{a^*} \geq UCB_{a^*}(t) \Big) \nonumber\\
&= \sum_{t \in [T]} \mathbb{P}\Bigg( 
       \sum_{i=1}^{d} \mathbf{w}_i \mu_{a^*}^i 
       \geq 
       \sum_{i=1}^{d} \mathbf{w}_i \Big( \widehat{\mu}_{a^*}^i(t) + \sqrt{\frac{\log(2d/\delta)}{2 N_{a^*}(t)}} \Big) 
     \Bigg)\nonumber \\
&= \sum_{t \in [T]} \mathbb{P}\Bigg( 
       \sum_{i=1}^{d} \mathbf{w}_i (\mu_{a^*}^i - \widehat{\mu}_{a^*}^i(t)) 
       \geq  \sqrt{\frac{\log(2d/\delta)}{2 N_{a^*}(t)}} 
     \Bigg)
\end{align}

Recall that for any action $a \in \mathcal{A}$, we have the concentration inequality:

\[
\widehat{\mu}^i_a(t)  - \sqrt{\frac{1}{2 N_a(t)} \log \frac{2d}{\delta}} 
\leq \mu_a^i \leq
\widehat{\mu}^i_a(t) + \sqrt{\frac{1}{2 N_a(t)} \log \frac{2d}{\delta}}, 
\quad \forall i \in [d], \;\; \text{with probability at least } 1-\delta.
\]

Thus, for all $i \in [d]$, we have 

\[
\mu_{a^*}^i - \widehat{\mu}_{a^*}^i(t) \leq \sqrt{\frac{\log(2d/\delta)}{2 N_{a^*}(t)}}, 
\quad \text{with probability at least } 1-\delta.
\]

Combining this with the previous step, we get

\begin{align}
\mathbb{P}\Big( \mu_{a^*} \geq \min_{t \in [T]} UCB_{a^*}(t) \Big)
&\leq \sum_{t \in [T]} \mathbb{P}\Bigg( 
       \sum_{i=1}^{d} \mathbf{w}_i (\mu_{a^*}^i - \widehat{\mu}_{a^*}^i(t))  
       \geq  \sqrt{\frac{\log(2d/\delta)}{2 N_{a^*}(t)}} 
     \Bigg) \nonumber\\
&\leq \sum_{t \in [T]} \delta \nonumber \\
&= T \delta.
\end{align}

The next step is to bound the probability of the second event.

\begin{align}
\mathbb{P}\Big(\widehat{\mu}_{a(\beta_a)} + \sqrt{\frac{\log(2d/\delta)}{2 \beta_a}} \geq \mu_{a^*}  \Big) &= \mathbb{P}\Big( \widehat{\mu}_{a(\beta_a)} + \sqrt{\frac{\log(2d/\delta)}{2 \beta_a}} \geq \mu_a + \Delta_a \Big) \nonumber\\
&= \mathbb{P}\Big( \widehat{\mu}_{a(\beta_a)} -\mu_a \geq  \Delta_a  - \sqrt{\frac{\log(2d/\delta)}{2 \beta_a}} \Big)
\end{align}

Suppose we set $\beta_a$ such that 
\[
\Delta_a - \sqrt{\frac{\log(2d/\delta)}{2 \beta_a}} \geq c \, \Delta_a,
\]  
where $c \in (0,1)$ is a constant. Then we have  

\begin{align}
\mathbb{P}\Big(\widehat{\mu}_{a(\beta_a)} + \sqrt{\frac{\log(2d/\delta)}{2 \beta_a}} \geq \mu_{a^*} \Big) 
&= \mathbb{P}\Big( \widehat{\mu}_{a(\beta_a)} - \mu_a \geq c \, \Delta_a \Big)  \nonumber\\
&\leq \exp\Big(-2 c^2 \Delta_a^2 \, \beta_a \Big),
\end{align}

where the final inequality follows from a one-sided Hoeffding concentration bound. Here, $\widehat{\mu}_{a(\beta_a)}$ is the average of the i.i.d. scalarized rewards obtained by playing action $a$ for $\beta_a$ rounds, $\mu_a$ is the true expected scalarized reward for action $a$, and each observed reward lies in $[0,1]$.

Consequently, we have

\[
\mathbb{P}(\mathcal{E}_{a}^c) \leq T \delta + \exp\Big(-2 c^2 \Delta_a^2 \, \beta_a \Big).
\]

This gives a bound on the expected number of times action $a$ is selected:

\[
\mathbb{E}[N_a(T)] \leq T \, \mathbb{P}(\mathcal{E}_{a}^c) + \beta_a 
\leq \beta_a + T \Big( T \delta + \exp(-2 c^2 \Delta_a^2 \, \beta_a) \Big)
= \beta_a + T^2 \delta + T \exp(-2 c^2 \Delta_a^2 \, \beta_a).
\]

Recall that we set $\beta_a$ to satisfy
\[
\Delta_a - \sqrt{\frac{\log(2d/\delta)}{2 \beta_a}} \geq c \, \Delta_a,
\] 
which is equivalent to
\[
(1-c) \Delta_a \geq \sqrt{\frac{\log(2d/\delta)}{2 \beta_a}} 
\quad \implies \quad
\beta_a \geq \frac{\log(2d/\delta)}{2 (1-c)^2 \Delta_a^2}.
\] 

Choosing 
\[
\beta_a = \frac{\log(2d/\delta)}{2 (1-c)^2 \Delta_a^2} 
\quad \text{and} \quad \delta = \frac{d}{T^2},
\] 
We obtain the final bound:

\begin{align}
\mathbb{E}[N_a(T)] 
&\leq \beta_a + d + T \exp\Bigg(- 2 c^2 \Delta_a^2 \cdot \frac{\log(2d/\delta)}{2 (1-c)^2 \Delta_a^2} \Bigg)  \nonumber\\
&= \beta_a + d + T \left( \frac{\delta}{2d} \right)^{\frac{c^2}{(1-c)^2}} \nonumber\\
&= \beta_a + d + 2^{-c^2/(1-c)^2} \, T^{1 - \frac{2c^2}{(1-c)^2}}.
\end{align}

If we set \(c = 0.5\), this simplifies to:

\begin{align*}
\mathbb{E}[N_a(T)] 
\leq  \beta_a + d + \frac{1}{2T}.
\end{align*}

For any $T \geq 1$, since $\frac{1}{2T} < 1$, we have
\begin{align*}
\mathbb{E}[N_a(T)] 
\leq  \beta_a + d + 1.
\end{align*}

Recall that
\[
\beta_a = \frac{\log(2d/\delta)}{2 (1-c)^2 \Delta_a^2}.
\]

Setting $c = 0.5$ and $\delta = \frac{d}{T^2}$, we obtain
\[
\beta_a = \frac{\log(2T^2)}{2 \cdot (0.5)^2 \, \Delta_a^2} = \frac{2 \log(2T^2)}{\Delta_a^2}.
\]

Thus, the final bound on the expected number of times action $a$ is selected is
\begin{align}
\mathbb{E}[N_a(T)] 
\leq \frac{2 \log(2T^2)}{\Delta_a^2} + d + 1.
\end{align}

Thus, the expected cumulative regret is
\begin{align}
\mathbb{E}[R_T] 
&= \sum_{a \in \mathcal{A}} \Delta_a \, \mathbb{E}[N_a(T)] \nonumber \\
&= \sum_{a: \Delta_a \geq \Delta} \Delta_a \, \mathbb{E}[N_a(T)] + \sum_{a: \Delta_a < \Delta} \Delta_a \, \mathbb{E}[N_a(T)]\nonumber \\
&\leq \sum_{a: \Delta_a \geq \Delta} \frac{2 \log(2T^2)}{\Delta_a} + (d+1) \sum_{a \in \mathcal{A}} \Delta_a + T \Delta \nonumber\\
&\leq k \frac{2 \log(2T^2)}{\Delta} + (d+1) \sum_{a \in \mathcal{A}} \Delta_a + T \Delta.
\end{align}

Setting $\Delta = \sqrt{\frac{2 k \log(2T^2)}{T}}$, we obtain
\begin{align}
\mathbb{E}[R_T] &\leq T \Delta + k \frac{2 \log(2T^2)}{\Delta} + (d+1) \sum_{a \in \mathcal{A}} \Delta_a \nonumber\\
&= 2 \sqrt{2 k T \log(2T^2)} + (d+1) \sum_{a \in \mathcal{A}} \Delta_a
\leq 4 \sqrt{k T \log(2T^2)} + (d+1) \sum_{a \in \mathcal{A}} \Delta_a
\end{align}

We now turn to \textbf{Algorithm~\ref{alg:ts_sc}}, which adapts Thompson Sampling (TS) from Bernoulli bandits to general
stochastic bandits with rewards supported on $[0,1]$, following Algorithm~2 of \cite{agrawal2012analysis}. This
construction is particularly useful in our setting, where rewards arise from the
\emph{scalarization of multi-objective outcomes}. Let $R_t \in [0,1]$ denote the observed scalarized reward for the arm
$A_t$ selected at time $t$. We generate an auxiliary Bernoulli random variable $R'_t \in \{0,1\}$ by performing a
Bernoulli trial with success probability $R_t$. The Beta--Bernoulli TS sufficient statistics of arm $A_t$ are then
updated as
\[
S_{A_t} \gets S_{A_t} + R'_t, \qquad
F_{A_t} \gets F_{A_t} + (1 - R'_t),
\]
where $S_a$ and $F_a$ denote the cumulative numbers of successes and failures for arm $a$.

\paragraph{Key observation.}
For any arm $a \in \mathcal{A}$, we have
\[
\mathbb{E}[R'_t \mid A_t = a]
=
\mathbb{E}\!\left[\mathbb{E}[R'_t \mid R_t] \mid A_t = a\right]
=
\mathbb{E}[R_t \mid A_t = a]
=
\mu_a,
\]
where $\mu_a$ denotes the mean scalarized reward of arm $a$. Under the standard stochastic bandit assumption that
rewards are i.i.d.\ conditional on the chosen arm, the induced feedback sequence
$\{R'_t : A_t = a\}$ is i.i.d.\ $\mathrm{Bernoulli}(\mu_a)$ for each arm $a$.
Consequently, the sufficient statistics $\{S_a, F_a\}$ evolve exactly as in a Bernoulli bandit with mean $\mu_a$.

Therefore, the regret guarantees for Beta--Bernoulli Thompson Sampling established by
Agrawal and Goyal~\cite{agrawal2012analysis,agrawal2017near} apply verbatim.
The Bernoulli resampling step introduces additional randomness but does not bias learning: the induced Bernoulli
rewards preserve the correct arm means in expectation. While this may increase the variance of the observations,
the posterior updates remain centered on the true scalarized means, ensuring the validity of the regret guarantees.

As a consequence, by applying the regret bound of \cite{agrawal2017near}, for
\textbf{Algorithm~\ref{alg:ts_sc}} with Beta priors, the expected cumulative regret satisfies
\[
\mathbb{E}[R_T] = \mathcal{O}\!\left(\sqrt{k T \log T}\right),
\]
where $k = |\mathcal{A}|$ denotes the number of arms. This completes the proof of
Theorem~\ref{thm4}.

\subsection{Proof of Theorem \ref{thm5} }
\vspace{0.5em}
\thmmmmmm*
\vspace{0.5em}
\textbf{Proof:} Suppose $\widehat{\mu}_a^i(t)$ is the empirical mean reward for objective $i \in [d]$ of action $a \in \mathcal{A}$, computed from observations collected before round $t$.
Let
\(
N_a(t) := \sum_{s=1}^{t-1} \mathbbm{1}\{A_s = a\}
\) denote the number of times action $a$ has been played up to (but not including) time $t$.
By Hoeffding's inequality, for any $\epsilon>0$,
\[
\mathbb{P}\Big(\big|\widehat{\mu}_a^i(t) - \mu_a^i\big| \ge \epsilon\Big)
\le 2 \exp\big(-2 \epsilon^2 N_a(t)\big).
\]

Setting $2 \exp(-2 \epsilon^2 N_a(t)) = \delta'$, we solve for $\epsilon$ to get
\[
\epsilon = \sqrt{\frac{1}{2 N_a(t)} \log\left( \frac{2}{\delta'} \right)}.
\]

Thus,
\[
\mathbb{P}\Big(\big|\widehat{\mu}_a^i(t) - \mu_a^i\big| \ge \sqrt{\frac{1}{2 N_a(t)} \log\left( \frac{2}{\delta'} \right)} \Big) \le \delta',
\]
which is equivalent to
\[
\mathbb{P}\Big(\big|\widehat{\mu}_a^i(t) - \mu_a^i\big| \le \sqrt{\frac{1}{2 N_a(t)} \log\left( \frac{2}{\delta'} \right)} \Big) \ge 1 - \delta'.
\]

Let us define the event
\[
\mathcal{E}_a^i(t) := \Big\{ \big|\widehat{\mu}_a^i(t) - \mu_a^i\big| \le \sqrt{\frac{1}{2 N_a(t)} \log\left( \frac{2}{\delta'} \right)} \Big\}.
\]

Next, define the intersection of all such events:
\[
\mathcal{E} := \bigcap_{\substack{i \in [d] \\ a \in \mathcal{A} \\ t \in [T]}} \mathcal{E}_a^i(t).
\]

 Since $\mathbb{P}(\mathcal{E}_a^i(t)) \ge 1 - \delta'$ for each $i, a, t$, using the union bound we have
\[
\mathbb{P}(\mathcal{E}) = 1 - \mathbb{P}\Big(\bigcup_{i\in[d],a\in \mathcal{A},t \in [T]} \overline{\mathcal{E}_a^i(t)}\Big) 
\ge 1 - \sum_{i\in[d],a\in \mathcal{A},t \in [T]} \mathbb{P}\big(\overline{\mathcal{E}_a^i(t)}\big) 
\ge 1 - k d T \delta',
\]
where $k = |\mathcal{A}|$. If we choose $\delta' := \frac{\delta}{k d T}$, then by the union bound we have $\mathbb{P}(\mathcal{E}) \ge 1 - \delta$. The event $\mathcal{E}$ ensures that, for all time steps, the true expected reward for every objective and every action lies within the corresponding confidence interval; that is,

\[
\mathcal{E} := \left\{
\begin{aligned}
& \forall a \in \mathcal{A} \\
& \forall t \in [T] \quad \quad
\big|\widehat{\mu}_a^i(t) - \mu_a^i\big| \le \sqrt{\frac{1}{2 N_a(t)} \log\left(\frac{2 k d T}{\delta}\right)} \\
& \forall i \in [d]
\end{aligned}
\right\}.
\]

 The probability of this good event satisfies $\mathbb{P}(\mathcal{E}) \ge 1 - \delta$. Also we use the notations $\mu^{+i}_a(t) := \widehat{\mu}_a^i(t)  + \sqrt{\frac{1}{2 N_a(t)} \log\left(\frac{2 k d T}{\delta}\right)}$  and $\mu^{-i}_a(t) := \widehat{\mu}_a^i(t)  - \sqrt{\frac{1}{2 N_a(t)} \log\left(\frac{2 k d T}{\delta}\right)}$.

The one step regret is given by:

\[
r_t := \max \bigg\{ 
\big(\mathbf{c}^\top \boldsymbol{\mu}(\mathbf{w}^\star) - \mathbf{c}^\top \boldsymbol{\mu}(\mathbf{w}^t)\big)_+, 
\mathbf{1}_{m}^\top \big[\mathbf{b} - \mathbf{A} \boldsymbol{\mu}(\mathbf{w}^t)\big]_+ 
\bigg\}
\]

\begin{equation*}
r_t  \leq\big(\mathbf{c}^\top \boldsymbol{\mu}(\mathbf{w}^\star) - \mathbf{c}^\top \boldsymbol{\mu}(\mathbf{w}^t)\big)_+ \;\;+\;\; \mathbf{1}_{m}^\top \big[\mathbf{b} - \mathbf{A} \boldsymbol{\mu}(\mathbf{w}^t)\big]_+ 
\end{equation*}

We focus on the first term, i.e.,
\[
\big(\mathbf{c}^{\top} \boldsymbol{\mu}(\mathbf{w}^\star) - \mathbf{c}^{\top} \boldsymbol{\mu}(\mathbf{w}^t)\big)_+.
\]  
Note that under the event \(\mathcal{E}\) we have
\[
\mathbf{c}^{\top} \sum_{a \in \mathcal{A}} \mathbf{w}^t_{a} \mathbf{U}_{a}(t) 
\ge 
\mathbf{c}^{\top} \sum_{a \in \mathcal{A}} \mathbf{w}^*_{a} \mathbf{U}_{a}(t).
\]  

The allocation \(\mathbf{w}^t\) is obtained by solving the following linear program:
\begin{equation}
\mathbf{w}^t = \operatorname*{arg\,max}_{\mathbf{w} \in \mathcal{S}(\mathcal{A})} 
\mathbf{c}^\top \sum_{a \in \mathcal{A}} \mathbf{w}_a \mathbf{U}_a(t)
\quad \text{s.t.} \quad
\mathbf{A} \sum_{a \in \mathcal{A}} \mathbf{w}_a \mathbf{U}_a(t) \ge \mathbf{b}  
\label{lp1}
\end{equation}

Similarly, the optimal allocation \(\mathbf{w}^*\) is the solution to:
\begin{equation}
\mathbf{w}^* = \operatorname*{arg\,max}_{\mathbf{w} \in \mathcal{S}(\mathcal{A})} 
\mathbf{c}^\top \sum_{a \in \mathcal{A}} \mathbf{w}_a \boldsymbol{\mu}_a
\quad \text{s.t.} \quad
\mathbf{A} \sum_{a \in \mathcal{A}} \mathbf{w}_a \boldsymbol{\mu}_a \ge \mathbf{b}  
\label{lp2}
\end{equation}

Any allocation \(\mathbf{w}\) feasible for LP~\eqref{lp2} is also feasible for LP~\eqref{lp1} under the event \(\mathcal{E}\), because entrywise we have \(\boldsymbol{\mu}_a \le \mathbf{U}_a(t)\) for all \(a \in \mathcal{A}\) and \(t \in [T]\). Also we have $\mathbf{A}\ge\mathbf{0}$. This implies
\[
\mathbf{b} \le 
\mathbf{A}\sum_{a \in \mathcal{A}} \mathbf{w}_a \boldsymbol{\mu}_a  
\le 
\mathbf{A} \sum_{a \in \mathcal{A}} \mathbf{w}_a \mathbf{U}_a(t),
\]  
so any weight vector \(\mathbf{w}\) feasible for LP~\eqref{lp2} is also feasible for LP~\eqref{lp1}.  

Since \(\mathbf{w}^*\) is contained in the feasible set of LP~\eqref{lp1}, and \(\mathbf{w}^t\) maximizes the objective 
\(\mathbf{c}^\top \sum_{a \in \mathcal{A}} \mathbf{w}_a \mathbf{U}_a(t)\) 
over this set, it follows that
\[
\mathbf{c}^{\top} \sum_{a \in \mathcal{A}} \mathbf{w}^t_{a} \mathbf{U}_{a}(t) 
\ge 
\mathbf{c}^{\top} \sum_{a \in \mathcal{A}} \mathbf{w}^*_{a} \mathbf{U}_{a}(t),
\]
under the event \(\mathcal{E}\). Recall that the first term 
\(
\big(\mathbf{c}^{\top} \boldsymbol{\mu}(\mathbf{w}^\star) - \mathbf{c}^{\top} \boldsymbol{\mu}(\mathbf{w}^t)\big)_+
\) can be bounded under the event \(\mathcal{E}\) as follows:
\begin{align}
\big(\mathbf{c}^{\top} \boldsymbol{\mu}(\mathbf{w}^\star) - \mathbf{c}^{\top} \boldsymbol{\mu}(\mathbf{w}^t)\big)_+
&\le 
\Bigg(
\mathbf{c}^{\top} \boldsymbol{\mu}(\mathbf{w}^\star) - \mathbf{c}^{\top} \boldsymbol{\mu}(\mathbf{w}^t)
+ 
\mathbf{c}^{\top} \sum_{a \in \mathcal{A}} \mathbf{w}^t_{a} \mathbf{U}_{a}(t) 
-
\mathbf{c}^{\top} \sum_{a \in \mathcal{A}} \mathbf{w}^*_{a} \mathbf{U}_{a}(t)
\Bigg)_+ \nonumber\\
&= \Bigg( \mathbf{c}^{\top} \boldsymbol{\mu}(\mathbf{w}^\star) - \mathbf{c}^{\top} \sum_{a \in \mathcal{A}} \mathbf{w}^*_{a} \mathbf{U}_{a}(t) \;\;+\;\; \mathbf{c}^{\top} \sum_{a \in \mathcal{A}} \mathbf{w}^t_{a} \mathbf{U}_{a}(t) - \mathbf{c}^{\top} \boldsymbol{\mu}(\mathbf{w}^t) \Bigg)_+ \nonumber\\
&= 
\Bigg(\;\;
\mathbf{c}^{\top} \sum_{a \in \mathcal{A}} \mathbf{w}^*_a \bigg{(}\boldsymbol{\mu}_a - \mathbf{U}_a(t)\bigg{)}
+
\mathbf{c}^{\top} \sum_{a \in \mathcal{A}} \mathbf{w}^t_a \bigg{(}\mathbf{U}_a(t) - \boldsymbol{\mu}_a\bigg{)}\;\;
\Bigg)_+ \nonumber \\
&\le 
\Bigg(\;
\mathbf{c}^{\top} \sum_{a \in \mathcal{A}} \mathbf{w}^t_a \bigg{(}\mathbf{U}_a(t) - \boldsymbol{\mu}_a\bigg{)}\;
\Bigg)_+
\label{eql1}
\end{align}
where the last inequality follows because under \(\mathcal{E}\), entrywise we have $\mathbf{c} \geq 0$ and \(\boldsymbol{\mu}_a \le \mathbf{U}_a(t)\) for all \(a \in \mathcal{A}\) and \(t \in [T]\), so 
\(\mathbf{c}^{\top} \sum_{a \in \mathcal{A}} \mathbf{w}^*_a (\boldsymbol{\mu}_a - \mathbf{U}_a(t)) \le 0\). We can rewrite the Equation \ref{eql1} as follows:

\begin{align}
\big(\mathbf{c}^{\top} \boldsymbol{\mu}(\mathbf{w}^\star) - \mathbf{c}^{\top} \boldsymbol{\mu}(\mathbf{w}^t)\big)_+
&\le 
\Bigg(
\mathbf{c}^{\top} \sum_{a \in \mathcal{A}} \mathbf{w}^t_a \big( \mathbf{U}_a(t) - \boldsymbol{\mu}_a \big)
\Bigg)_+ \nonumber \\
&= 
\Bigg(
\sum_{a \in \mathcal{A}} \mathbf{w}^t_a \, \mathbf{c}^{\top} \big( \mathbf{U}_a(t) - \boldsymbol{\mu}_a \big)
\Bigg)_+
\label{eql3}
\end{align}

Under the event $\mathcal{E}$, for all $a \in \mathcal{A}$ and $t \in [T]$, we have
\[
\mathbf{U}_a(t) - \boldsymbol{\mu}_a
\le 
2\mathbf{1}_d \sqrt{\frac{1}{2 N_a(t)} \log  \left(\frac{2 k d T}{\delta}\right)},
\]

where $\mathbf{1}_d$ denotes the $d$-dimensional column vector of all ones. Consequently, under the event $\mathcal{E}$, for all $a \in \mathcal{A}$,

\begin{equation}
  \mathbf{c}^{\top} \big( \mathbf{U}_a(t) - \boldsymbol{\mu}_a \big)
\le 
2\mathbf{c}^{\top} \mathbf{1}_d \sqrt{\frac{1}{2 N_a(t)} \log  \left(\frac{2 k d T}{\delta}\right)}
=
2\|\mathbf{c}\|_1 \sqrt{\frac{1}{2 N_a(t)} \log  \left(\frac{2 k d T}{\delta}\right)}  
\label{eql4}
\end{equation}

Combining~\eqref{eql3} and~\eqref{eql4}, we obtain the following bound on the first term:

\begin{align}
\big(\mathbf{c}^{\top} \boldsymbol{\mu}(\mathbf{w}^\star) - \mathbf{c}^{\top} \boldsymbol{\mu}(\mathbf{w}^t)\big)_+
&\leq \Bigg(
\sum_{a \in \mathcal{A}} \mathbf{w}^t_a \, \mathbf{c}^{\top} \big( \mathbf{U}_a(t) - \boldsymbol{\mu}_a \big)
\Bigg)_+ \nonumber\\
&\leq  \Bigg(
\sum_{a \in \mathcal{A}} \mathbf{w}^t_a \; 2\|\mathbf{c}\|_1 \sqrt{\frac{1}{2 N_a(t)} \log  \left(\frac{2 k d T}{\delta}\right)} 
\Bigg)_+ \nonumber\\
&= 2\|\mathbf{c}\|_1 \sqrt{\frac{1}{2} \log  \left(\frac{2 k d T}{\delta}\right)} \;\;
\sum_{a \in \mathcal{A}} \mathbf{w}^t_a \;  \sqrt{\frac{1}{ N_a(t)} } 
\label{eql6}
\end{align}

We define the filtration at time step $t$ as the $\sigma$-algebra containing the entire past history of actions and observed rewards, namely,
\[
\mathcal{F}_{t-1}
:=
\sigma\Big(
A_1, \dots, A_{t-1},
\mathbf{Y}_1(A_1), \dots, \mathbf{Y}_{t-1}(A_{t-1})
\Big),
\]
that is, $\mathcal{F}_{t-1}$ contains everything observed before time $t$. 
Here, $A_i$ denotes the action selected at time step $i$, and $\mathbf{Y}_i(A_i)$ denotes the corresponding observed reward vector over all objectives. 
The weight vector $\mathbf{w}^t$ is $\mathcal{F}_{t-1}$ measurable, and the UCB vectors $\mathbf{U}_a(t)$ are also $\mathcal{F}_{t-1}$ measurable for all $a \in \mathcal{A}$. 
According to the sampling rule of Algorithm~\ref{alg:constraint}, we have
\[
\mathbb{P}(A_t = a \mid \mathcal{F}_{t-1}) = \mathbf{w}^t_a .
\]

We now define
\[
g_{A_t}(t) := \sqrt{\frac{1}{N_{A_t}(t)}},
\]
where $A_t$ denotes the action selected at time step $t$ and is a random variable distributed according to
\[
\mathbb{P}(A_t = a \mid \mathcal{F}_{t-1}) = \mathbf{w}^t_a .
\]

Then,
\begin{equation}
\mathbb{E}  \left[g_{A_t}(t) \mid \mathcal{F}_{t-1}\right]
=
\sum_{a \in \mathcal{A}} \mathbb{P}(A_t = a \mid \mathcal{F}_{t-1}) g_a(t)
=
\sum_{a \in \mathcal{A}} \mathbf{w}^t_a \sqrt{\frac{1}{N_a(t)}} 
\label{eql5}
\end{equation}

From \eqref{eql6} and \eqref{eql5}, we obtain
\begin{align}
\big(\mathbf{c}^{\top} \boldsymbol{\mu}(\mathbf{w}^\star) - \mathbf{c}^{\top} \boldsymbol{\mu}(\mathbf{w}^t)\big)_+
&\le
2\|\mathbf{c}\|_1
\sqrt{\frac{1}{2} \log  \left(\frac{2 k d T}{\delta}\right)}
\;\;\sum_{a \in \mathcal{A}} \mathbf{w}^t_a \sqrt{\frac{1}{N_a(t)}} \nonumber\\
&=
2\|\mathbf{c}\|_1
\sqrt{\frac{1}{2} \log  \left(\frac{2 k d T}{\delta}\right)}
\;\; \mathbb{E}  \left[g_{A_t}(t) \mid \mathcal{F}_{t-1}\right] \nonumber\\
&=
2\|\mathbf{c}\|_1
\sqrt{\frac{1}{2} \log  \left(\frac{2 k d T}{\delta}\right)}
\;\;\mathbb{E}  \left[\sqrt{\frac{1}{N_{A_t}(t)}} \,\middle|\, \mathcal{F}_{t-1}\right]
\label{eql10}
\end{align}

Now we focus on the second term of the one-step regret:
\[
\mathbf{1}_{m}^\top \big[\mathbf{b} - \mathbf{A} \boldsymbol{\mu}(\mathbf{w}^t)\big]_+.
\]

Under the event $\mathcal{E}$, for all $a \in \mathcal{A}$ and $t \in [T]$, the true mean satisfies
\[
\boldsymbol{\mu}_a \ge \mathbf{U}_a(t) - 2 \mathbf{1}_d \sqrt{\frac{1}{2 N_a(t)} \log  \left(\frac{2 k d T}{\delta}\right)},
\]
where $\mathbf{1}_d$ is the $d$-dimensional vector of ones.

Consequently, for the linear combination over the stochastic policy $\mathbf{w}^t$, we have
\begin{align}
\mathbf{A} \boldsymbol{\mu}(\mathbf{w}^t) 
&= \mathbf{A} \sum_{a \in \mathcal{A}} \mathbf{w}_a^t \boldsymbol{\mu}_a \nonumber\\
&\ge \mathbf{A} \sum_{a \in \mathcal{A}} \mathbf{w}_a^t \Bigg( \mathbf{U}_a(t) - 2 \mathbf{1}_d \sqrt{\frac{1}{2 N_a(t)} \log  \left(\frac{2 k d T}{\delta}\right)} \Bigg) \nonumber \\
&= \mathbf{A} \sum_{a \in \mathcal{A}} \mathbf{w}_a^t \mathbf{U}_a(t) - 2 \mathbf{A} \sum_{a \in \mathcal{A}} \mathbf{w}_a^t \mathbf{1}_d \sqrt{\frac{1}{2 N_a(t)} \log  \left(\frac{2 k d T}{\delta}\right)}.
\end{align}

Therefore, the second term can be bounded as:
\begin{align}
\mathbf{1}_{m}^\top \big[\mathbf{b} - \mathbf{A} \boldsymbol{\mu}(\mathbf{w}^t)\big]_+ 
&\le \mathbf{1}_{m}^\top \big[ \mathbf{b} - \mathbf{A} \sum_{a \in \mathcal{A}} \mathbf{w}_a^t \mathbf{U}_a(t) 
+ 2 \mathbf{A} \sum_{a \in \mathcal{A}} \mathbf{w}_a^t \mathbf{1}_d \sqrt{\frac{1}{2 N_a(t)} \log  \left(\frac{2 k d T}{\delta}\right)} \Big]_+.
\end{align}

Since the allocation $\mathbf{w}^t$ is obtained by solving the linear program
\begin{equation*}
\mathbf{w}^t = \operatorname*{arg\,max}_{\mathbf{w} \in \mathcal{S}(\mathcal{A})} 
\mathbf{c}^\top \sum_{a \in \mathcal{A}} \mathbf{w}_a \mathbf{U}_a(t)
\quad \text{s.t.} \quad
\mathbf{A} \sum_{a \in \mathcal{A}} \mathbf{w}_a \mathbf{U}_a(t) \ge \mathbf{b},
\end{equation*}
we have
\[
\mathbf{b} - \mathbf{A} \sum_{a \in \mathcal{A}} \mathbf{w}_a^t \mathbf{U}_a(t) \le 0.
\]

Hence, the second term simplifies to
\begin{align}
\mathbf{1}_{m}^\top \big[\mathbf{b} - \mathbf{A} \boldsymbol{\mu}(\mathbf{w}^t)\big]_+ 
&\le 2 \, \mathbf{1}_m^\top \mathbf{A} \sum_{a \in \mathcal{A}} \mathbf{w}_a^t \mathbf{1}_d \sqrt{\frac{1}{2 N_a(t)} \log  \left(\frac{2 k d T}{\delta}\right)} \nonumber\\
&=  2 \, \mathbf{1}_m^\top \mathbf{A} \mathbf{1}_d \sqrt{\frac{1}{2} \log  \left(\frac{2 k d T}{\delta}\right)} \;\; \sum_{a \in \mathcal{A}} \mathbf{w}^t_a \;  \sqrt{\frac{1}{ N_a(t)} } 
\end{align}

Now, using \eqref{eql5}, we obtain the following bound on the second term:

\begin{equation}
    \mathbf{1}_{m}^\top \big[\mathbf{b} - \mathbf{A} \boldsymbol{\mu}(\mathbf{w}^t)\big]_+ 
\le 2 \, \mathbf{1}_m^\top \mathbf{A} \mathbf{1}_d \sqrt{\frac{1}{2} \log  \left(\frac{2 k d T}{\delta}\right)} \;\;\mathbb{E}  \left[\sqrt{\frac{1}{N_{A_t}(t)}} \,\middle|\, \mathcal{F}_{t-1}\right]\
\label{eql11}
\end{equation}

We now summarize the regret analysis so far:
\begin{equation*}
\label{eq:regret-constraints}
r_t := \max \Big\{\;\big(\mathbf{c}^{\top} \boldsymbol{\mu}(\mathbf{w}^*) - \mathbf{c}^{\top} \boldsymbol{\mu}(\mathbf{w}^t) \big)_+, \; \mathbf{1}_{m}^\top \big[b - \mathbf{A} \boldsymbol{\mu}(\mathbf{w}^t)\big]_+ \Big\}.
\end{equation*}

Using the inequality \(\max\{a,b\} \le a+b\) for non-negative \(a,b\), we have
\begin{equation*}
r_t \le \big(\mathbf{c}^{\top} \boldsymbol{\mu}(\mathbf{w}^*) - \mathbf{c}^{\top} \boldsymbol{\mu}(\mathbf{w}^t) \big)_+ + \mathbf{1}_{m}^\top \big[\mathbf{b} - \mathbf{A} \boldsymbol{\mu}(\mathbf{w}^t)\big]_+.
\end{equation*}

It follows directly from Equations \ref{eql10} and \ref{eql11} that under the event $\mathcal{E}$, we have

\begin{align}
\big(\mathbf{c}^{\top} \boldsymbol{\mu}(\mathbf{w}^*) - \mathbf{c}^{\top} \boldsymbol{\mu}(\mathbf{w}^t) \big)_+ 
&\le 2 \|\mathbf{c}\|_1 \sqrt{\frac{1}{2} \log\left(\frac{2 k d T}{\delta}\right)} 
\; \mathbb{E} \Big[\frac{1}{\sqrt{N_{A_t}(t)}} \,\Big|\, \mathcal{F}_{t-1} \Big], \\
\mathbf{1}_{m}^\top [\mathbf{b} - \mathbf{A} \boldsymbol{\mu}(\mathbf{w}^t)]_+ 
&\le 2 (\mathbf{1}_{m}^\top \mathbf{A} \mathbf{1}_{d}) \sqrt{\frac{1}{2} \log\left(\frac{2 k d T}{\delta}\right)} 
\; \mathbb{E} \Big[\frac{1}{\sqrt{N_{A_t}(t)}} \,\Big|\, \mathcal{F}_{t-1} \Big].
\end{align}

Using the indicator of the high-probability event $\mathcal E$ and summing the per-round bounds over $t=1,\dots,T$, we obtain the following bound for cumulative expected regret:

\begin{align}
\mathbb{E}\!\left[\mathbbm{1}\{\mathcal{E}\}R_T\right]
= \mathbb{E}\!\left[\mathbbm{1}\{\mathcal{E}\}\sum_{t=1}^T r_t\right]
&\le \mathbb{E}\!\left[\sum_{t=1}^T \mathbbm{1}\{\mathcal{E}\}
\left( \mathbf{c}^{\top} \boldsymbol{\mu}(\mathbf{w}^{\star})
- \mathbf{c}^{\top} \boldsymbol{\mu}(\mathbf{w}^t) \right)_+
+ \sum_{t=1}^T \mathbbm{1}\{\mathcal{E}\}\,
\mathbf{1}_m^\top [\mathbf{b} - \mathbf{A}\boldsymbol{\mu}(\mathbf{w}^t)]_+ \right] \nonumber\\
&\le \mathbb{E}\Bigg[
\sum_{t=1}^T \mathbbm{1}\{\mathcal{E}\}\, 2\|\mathbf{c}\|_1
\sqrt{\frac{1}{2}\log\!\left(\frac{2kdT}{\delta}\right)}
\;\mathbb{E}\!\left[\frac{1}{\sqrt{N_{A_t}(t)}}\Bigm|\mathcal{F}_{t-1}\right] \nonumber\\
&\qquad\quad
+ \sum_{t=1}^T \mathbbm{1}\{\mathcal{E}\}\, 2(\mathbf{1}_m^\top \mathbf{A}\mathbf{1}_d)
\sqrt{\frac{1}{2}\log\!\left(\frac{2kdT}{\delta}\right)}
\;\mathbb{E}\!\left[\frac{1}{\sqrt{N_{A_t}(t)}}\Bigm|\mathcal{F}_{t-1}\right]
\Bigg] \nonumber\\
&\le \sum_{t=1}^T 2\|\mathbf{c}\|_1
\sqrt{\frac{1}{2}\log\!\left(\frac{2kdT}{\delta}\right)}
\;\mathbb{E}\!\left[\mathbb{E}\!\left[\frac{1}{\sqrt{N_{A_t}(t)}}\Bigm|\mathcal{F}_{t-1}\right]\right] \nonumber\\
&\qquad\quad
+ \sum_{t=1}^T 2(\mathbf{1}_m^\top \mathbf{A}\mathbf{1}_d)
\sqrt{\frac{1}{2}\log\!\left(\frac{2kdT}{\delta}\right)}
\;\mathbb{E}\!\left[\mathbb{E}\!\left[\frac{1}{\sqrt{N_{A_t}(t)}}\Bigm|\mathcal{F}_{t-1}\right]\right] \nonumber\\
&= \sum_{t=1}^T 2\|\mathbf{c}\|_1
\sqrt{\frac{1}{2}\log\!\left(\frac{2kdT}{\delta}\right)}
\;\mathbb{E}\!\left[\frac{1}{\sqrt{N_{A_t}(t)}}\right] \nonumber\\
&\qquad\quad
+ \sum_{t=1}^T 2(\mathbf{1}_m^\top \mathbf{A}\mathbf{1}_d)
\sqrt{\frac{1}{2}\log\!\left(\frac{2kdT}{\delta}\right)}
\;\mathbb{E}\!\left[\frac{1}{\sqrt{N_{A_t}(t)}}\right] \nonumber\\
&= 2\|\mathbf{c}\|_1
\sqrt{\frac{1}{2}\log\!\left(\frac{2kdT}{\delta}\right)}
\;\mathbb{E}\!\left[\sum_{t=1}^T \frac{1}{\sqrt{N_{A_t}(t)}}\right] \nonumber\\
&\qquad\quad
+ 2(\mathbf{1}_m^\top \mathbf{A}\mathbf{1}_d)
\sqrt{\frac{1}{2}\log\!\left(\frac{2kdT}{\delta}\right)}
\;\mathbb{E}\!\left[\sum_{t=1}^T \frac{1}{\sqrt{N_{A_t}(t)}}\right].
\end{align}

The second inequality applies the per-round bounds, which hold on $\mathcal{E}$,
and multiplies both sides by $\mathbbm{1}\{\mathcal{E}\}$.
Next, since the resulting right-hand side is nonnegative,
we use $\mathbbm{1}\{\mathcal{E}\}\le 1$ to drop the indicator and obtain an upper bound.
Finally, we apply the tower property
$\mathbb{E}[\mathbb{E}[X\mid \mathcal{F}_{t-1}]]=\mathbb{E}[X]$
to replace the expectation of the conditional expectation
by an unconditional expectation.

Rewriting the sum in terms of individual actions $a \in \mathcal{A}$ and their selection counts $N_a(T)$, we obtain
\[
\sum_{t=1}^T \frac{1}{\sqrt{N_{A_t}(t)}} \;\leq\; \sum_{a \in \mathcal{A}} \sum_{j=1}^{N_a(T)} \frac{1}{\sqrt{j}}\,.
\]

and as a result, we have

\begin{align}
\mathbb{E}[\mathbbm{1}\{\mathcal{E}\}R_T] 
&\le 2 \|\mathbf{c}\|_1 \sqrt{\frac{1}{2} \log\left(\frac{2 k d T}{\delta}\right)} 
\;\mathbb{E} \Big[ \sum_{a \in \mathcal{A}} \sum_{j=1}^{N_a(T)} \frac{1}{\sqrt{j}} \Big]  \nonumber \\
&\quad + 2 (\mathbf{1}_{m}^\top\mathbf{A} \mathbf{1}_d) \sqrt{\frac{1}{2} \log\left(\frac{2 k d T}{\delta}\right)} 
\;\mathbb{E} \Big[ \sum_{a \in \mathcal{A}} \sum_{j=1}^{N_a(T)} \frac{1}{\sqrt{j}} \Big].
\end{align}

Using the standard bound $\sum_{j=1}^{N_a(T)} \frac{1}{\sqrt{j}} \le \int_{0}^{N_a(T)} \frac{1}{\sqrt{t}} \, dt \le 2 \sqrt{N_a(T)}$, we get

\begin{align}
\mathbb{E}[\mathbbm{1}\{\mathcal{E}\}R_T] 
&\le 4 \|\mathbf{c}\|_1 \sqrt{\frac{1}{2} \log\left(\frac{2 k d T}{\delta}\right)} 
\; \mathbb{E} \Big[ \sum_{a \in \mathcal{A}}  \sqrt{N_a(T)} \Big] \nonumber \\
&\quad + 4 (\mathbf{1}_{m}^\top\mathbf{A} \mathbf{1}_d) \sqrt{\frac{1}{2} \log\left(\frac{2 k d T}{\delta}\right)} 
\;  \mathbb{E} \Big[ \sum_{a \in \mathcal{A}}  \sqrt{N_a(T)} \Big].
\end{align}

We have
\[
\sum_{a \in \mathcal{A}} \sqrt{N_a(T)}
= \sum_{a \in \mathcal{A}} \sqrt{1}\,\sqrt{N_a(T)}.
\]
Applying the Cauchy--Schwarz inequality yields
\[
\sum_{a \in \mathcal{A}} \sqrt{1}\,\sqrt{N_a(T)}
\le \sqrt{\sum_{a \in \mathcal{A}} 1}\;
     \sqrt{\sum_{a \in \mathcal{A}} N_a(T)}
= \sqrt{k}\,\sqrt{\sum_{a \in \mathcal{A}} N_a(T)}.
\]
Since the total number of pulls satisfies
\[
\sum_{a \in \mathcal{A}} N_a(T) = T,
\]
it follows that
\[
\sum_{a \in \mathcal{A}} \sqrt{N_a(T)} \le \sqrt{kT}.
\]
Taking expectations on both sides, we obtain
\[
\mathbb{E}  \left[\sum_{a \in \mathcal{A}} \sqrt{N_a(T)}\right]
\le \sqrt{kT}.
\]

Thus, we obtain the following bound:
\begin{align}
\mathbb{E}[\mathbbm{1}\{\mathcal{E}\}R_T] 
&\le 4 \|\mathbf{c}\|_1 
\sqrt{\frac{1}{2} \log  \left(\frac{2 k d T}{\delta}\right)} \, \sqrt{kT}
+ 4 (\mathbf{1}_{m}^\top \mathbf{A} \mathbf{1}_d)
\sqrt{\frac{1}{2} \log  \left(\frac{2 k d T}{\delta}\right)} \, \sqrt{kT} \nonumber \\
&= \left( 2 \sqrt{2}\, \|\mathbf{c}\|_1
+ 2 \sqrt{2}\, (\mathbf{1}_{m}^\top \mathbf{A} \mathbf{1}_d) \right)
\sqrt{kT \log  \left(\frac{2 k d T}{\delta}\right)} \nonumber \\
&\le 4 \left( \|\mathbf{c}\|_1
+ (\mathbf{1}_{m}^\top \mathbf{A} \mathbf{1}_d) \right)
\sqrt{kT \log  \left(\frac{2 k d T}{\delta}\right)} .
\end{align}

Recall that
\[
r_t := \max\Big\{
(\mathbf{c}^\top \boldsymbol{\mu}(\mathbf{w}^\star)
- \mathbf{c}^\top \boldsymbol{\mu}(\mathbf{w}^t))_+,\;
\mathbf{1}_m^\top [\mathbf{b}-\mathbf{A}\boldsymbol{\mu}(\mathbf{w}^t)]_+
\Big\}.
\]
Since $\boldsymbol{\mu}(\mathbf{w})\in[0,1]^d$, $\mathbf{c}\ge\mathbf{0}$, $\mathbf{A}\ge\mathbf{0}$,
and $\mathbf{b}\ge\mathbf{0}$, we have
\[
(\mathbf{c}^\top \boldsymbol{\mu}(\mathbf{w}^\star)
- \mathbf{c}^\top \boldsymbol{\mu}(\mathbf{w}^t))_+
\le \|\mathbf{c}\|_1,
\qquad
\mathbf{1}_m^\top [\mathbf{b}-\mathbf{A}\boldsymbol{\mu}(\mathbf{w}^t)]_+
\le  \|\mathbf{b}\|_1.
\]
Therefore,
\[
r_t \le \max  \left\{\|\mathbf{c}\|_1,\ \|\mathbf{b}\|_1\right\}
=: \alpha,
\]
and consequently \(R_T=\sum_{t=1}^T r_t \le \alpha T\) almost surely.

Finally, accounting for the failure event $\mathcal{E}^c$, we have
$\mathbb{P}(\mathcal{E}^c) \leq \delta$, and we set $\delta := 1/T$. Then

\begin{align}
\mathbb{E}[R_T]
&= \mathbb{E}[\mathbbm{1}\{\mathcal{E}\}R_T] 
+ \mathbb{E}[\mathbbm{1}\{\mathcal{E}^c\}R_T]  \nonumber \\
&\le \mathbb{E}[\mathbbm{1}\{\mathcal{E}\}R_T] +  \mathbb{E}[\mathbbm{1}\{\mathcal{E}^c\}]  (\alpha T) \nonumber \\
&= \mathbb{E}[\mathbbm{1}\{\mathcal{E}\}R_T] +  \mathbb{P}(\mathcal{E}^c) (\alpha T) \nonumber \\
&\le \mathbb{E}[\mathbbm{1}\{\mathcal{E}\}R_T] + \delta\, \alpha T \nonumber \\
&\le \mathbb{E}[\mathbbm{1}\{\mathcal{E}\}R_T] + \alpha \nonumber \\
&\le 4 \left( \|\mathbf{c}\|_1 + \mathbf{1}_{m}^\top \mathbf{A} \mathbf{1}_d \right)
\sqrt{kT \log  \left(2 k d T^2 \right)} + \alpha .
\end{align}

This completes the proof of Theorem \ref{thm5}.

\subsection{Structural Causal Models used in the Experiments}

We specify the structural causal models (SCMs) consistent with the causal graphs shown in Figure~\ref{grphs_exp}, which are used in our experiments. 

\textbf{For $\mathcal{G}_5$ from Figure~\ref{grphs_exp}(a): } $U_1 = Bern(0.5)$ and $U_2 = Bern(0.5)$
\begin{equation}
  X_1  = U_1U_2, \;\;
   X_2 = X_1,\;\;
   Y_1 = X_2 \oplus U_1, \;\;
    X_3 = U_2,\;\; \text{and} \;\; 
    Y_2 = X_3   
\end{equation}

\textbf{For $\mathcal{G}_6$ from Figure~\ref{grphs_exp}(b): } $U_1 = Bern(0.5)$ and $U_2 = Bern(0.8)$
\begin{equation}
  X_1  = U_1, \;\;
  X_2 = 1-X_1,\;\;
   Y_1 = X_2U_1, \;\;
    X_3 = X_1,\;\;
    X_4 = 1-X_3,\;\;  \text{and} \;\;
  Y_2 = X_4U_2   
\end{equation}
  
\vspace{1em}  
\textbf{For $\mathcal{G}_7$ from Figure~\ref{grphs_exp}(c): } $U_1 = Bern(0.55)$, $U_2 = Bern(0.55)$ and $U_3 = Bern(0.8)$
\begin{equation}
  X_1  = U_1U_2, \;
  X_2 = 1-X_1,\;
   Y_1 = X_2U_1, \;
    X_3 =X_1U_2,\;
    Y_2 = X_3(1-X_4),\;
  X_4 = X_2\;   \& \;
  Y_3 = X_4U_3
\end{equation}

In addition to specifying the SCMs associated with the causal graphs in Figure~\ref{grphs_exp}, for the simple linear scalarization scenario we use the mean of the average reward nodes as the final reward signal. For the constrained optimization scenario, the objective is defined as the simple average of all reward nodes, while two linear constraints are imposed with randomly generated weights and thresholds.

\subsection{Targeted Convex-Combination Elimination Ablation}
\label{app:mixture_only_ablation}

\begin{figure*}[h]
\centering
\begin{subfigure}[b]{5cm}
    \centering
    \includegraphics[height=3.5cm,width=5cm]{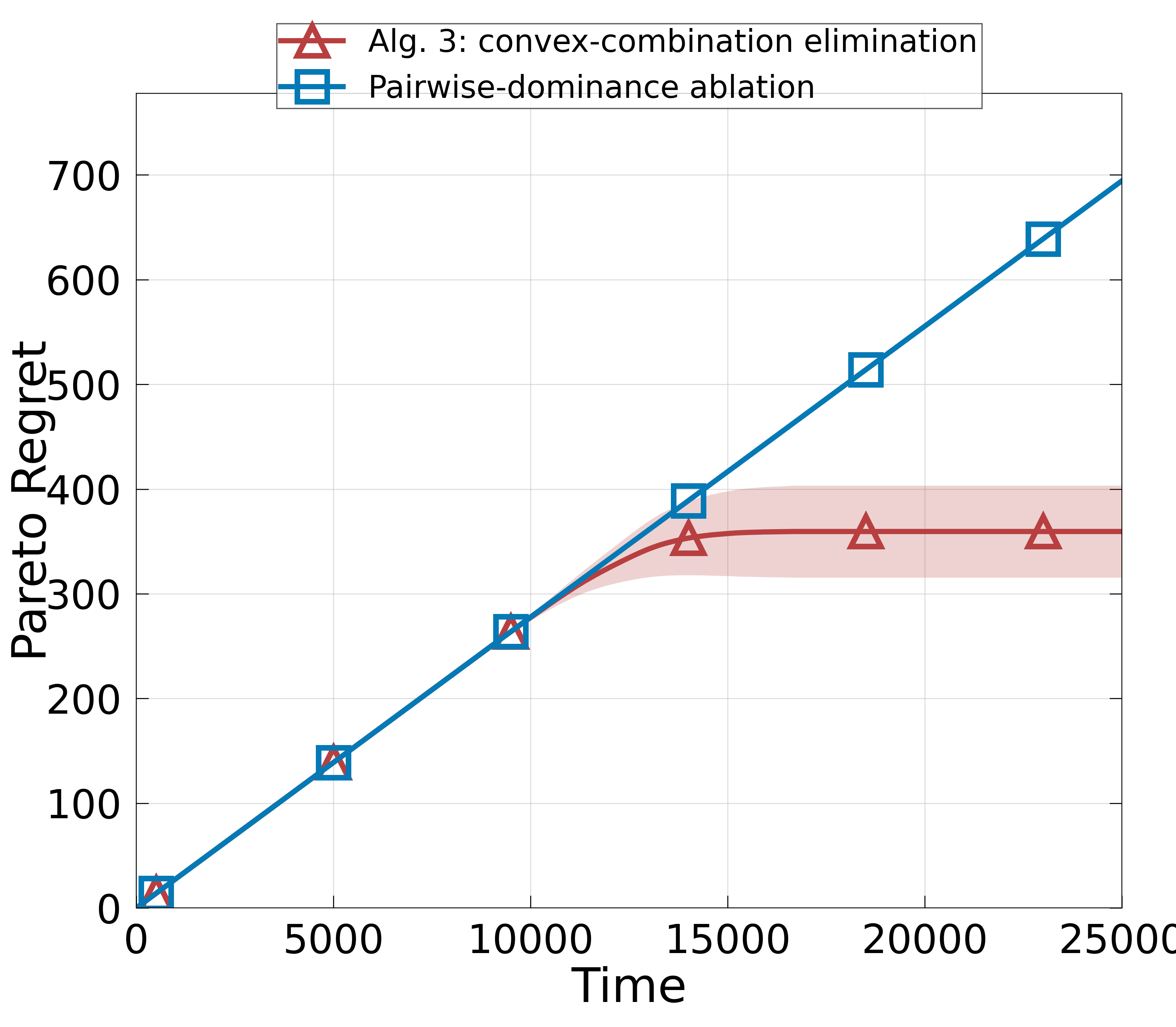}
    \subcaption{Cumulative Pareto regret}
\end{subfigure}
\enspace
\begin{subfigure}[b]{5cm}
    \centering
    \includegraphics[height=3.5cm,width=5cm]{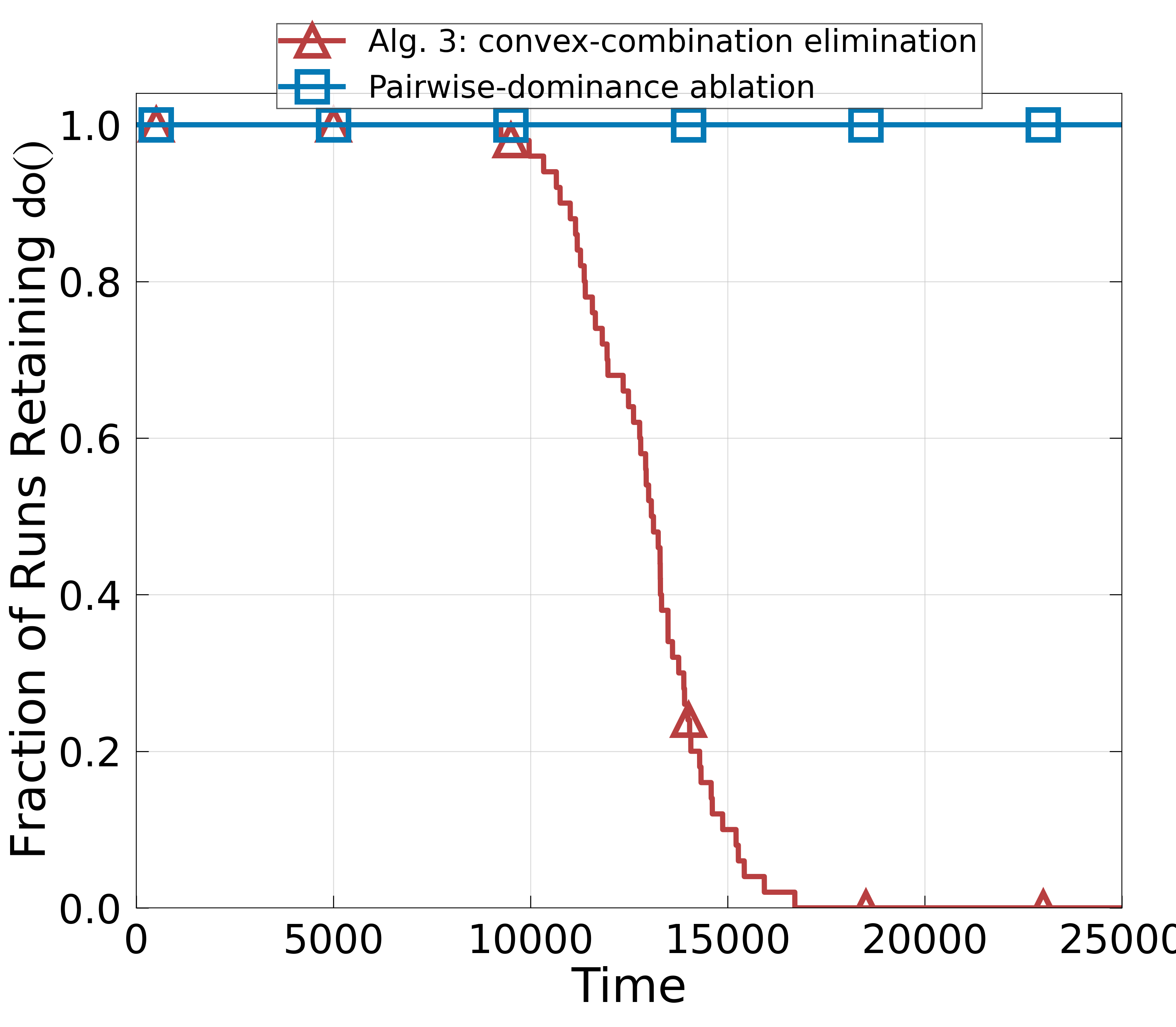}
    \subcaption{Fraction of runs retaining $\mathrm{do}()$}
\end{subfigure}
\caption{
Comparison of convex-combination and pairwise-dominance elimination using the SCM described in Experiments section.
Results are averaged over $50$ independent runs with $T=25{,}000$ and $\delta=0.1$. Shaded regions in (a) indicate one standard deviation.}
\label{fig:mixture_only_ablation}
\end{figure*}

Figure~\ref{fig:mixture_only_ablation} presents the targeted ablation described in Section~7, comparing Algorithm~\ref{alg:pareto_ucb_cc} with a pairwise-dominance variant on the SCM in Figure~1(b). Both methods use the same three-action space, confidence bounds, and least-sampled-active-action selection rule, differing only in the elimination criterion. Figure~\ref{fig:mixture_only_ablation}(a) shows that convex-combination elimination achieves lower cumulative Pareto regret, while Figure~\ref{fig:mixture_only_ablation}(b) shows that it eliminates the observational action $\mathrm{do}()$ in every run, whereas pairwise elimination retains it throughout.

\subsection{Random Graph Generation Procedure}
\label{app:random_graph_generation}

For each graph size, we generate random causal graphs as follows. We first sample a random ordering $\sigma$ over the vertices. Then, for each node in this ordering, we determine its in-degree as $X_n=\max\{1,\operatorname{Bin}(n-1,\rho)\}$, where $\rho$ is the directed-edge density parameter and $n-1$ is the number of preceding nodes in the ordering. We then select $X_n$ parents uniformly at random from the preceding nodes in the ordering. This ensures that all directed edges respect the ordering $\sigma$, and hence that the resulting directed graph is acyclic. Next, we chordalize the graph using the elimination algorithm of~\citet{koller2009probabilistic}, with an elimination ordering given by the reverse of $\sigma$. Finally, we introduce a latent confounder between each pair of nodes independently with probability $\rho_L$. This overall procedure follows the random graph generation setup used in~\citet{squires2020active}.

\subsection{Experimental Compute Resources and Runtime} \label{compute_res}

We ran our experiments on a server equipped with an AMD Ryzen Threadripper PRO 5995WX CPU with 64 cores and 128 threads, a base clock speed of 2.7 GHz, a maximum boost clock speed of 4.5 GHz, and 128 GB of RAM. The total runtime for generating the experimental plots in Figures~\ref{syn_res1} and~\ref{syn_res2} was approximately 30 minutes. The total runtime for generating the plots in Figure~\ref{syn_res3} was also approximately 30 minutes.

\subsection{Broader Impacts of Our Work}
\label{app:broader_impacts}

This paper aims to advance the field of machine learning by developing methods for multi-objective causal bandit problems. Since causal bandit frameworks can model real-world decision-making scenarios, our work may have potential societal implications. In particular, biased, misspecified, or incomplete causal models could lead to misleading conclusions, suboptimal interventions, or inappropriate policy recommendations when deployed in practice. Therefore, practical applications of our methodology require careful validation of the assumed causal structure, awareness of possible biases in the data, and explicit consideration of ethical constraints on allowable interventions. Extra care is especially needed in high-stakes domains, where interventions may affect individuals or communities. Our work is primarily theoretical and algorithmic, and we do not conduct experiments involving human subjects or deploy the proposed methods in real-world decision-making systems.


\end{document}